\pdfoutput=1
\documentclass[]{fairmeta}
\usepackage[round,authoryear]{natbib}
\usepackage{hyperref}
\hypersetup{pdftitle={HyperAgents}, pdfauthor={Jenny Zhang, Bingchen Zhao, Wannan Yang, Jakob Foerster, Jeff Clune, Minqi Jiang, Sam Devlin, Tatiana Shavrina}}
\usepackage[nameinlink,noabbrev]{cleveref}
\usepackage{colortbl}
\usepackage{nicematrix}
\usepackage{amssymb}
\usepackage{amsmath}
\usepackage{adjustbox}
\usepackage{soul}
\usepackage[inline]{enumitem}
\usepackage{booktabs}
\usepackage{tabularx}
\usepackage{color}
\usepackage{xcolor}
\usepackage{bbding}
\usepackage{listings}
\usepackage{multicol}
\usepackage{xspace}
\usepackage{lmodern}
\usepackage{tablefootnote}
\usepackage{float}
\usepackage{titletoc}
\usepackage[ruled,vlined]{algorithm2e}

\crefname{algocf}{Algorithm}{Algorithms}
\Crefname{algocf}{Algorithm}{Algorithms}

\makeatletter
\AtBeginDocument{

}
\newcommand\authorNewLine[2][]{
  \addtolist[#1]{#2}{\authorlist}{\authorformat}{\\ }%
}
\newcommand\affiliationNewLine[2][]{%
  \addtolist[#1]{#2}{\affiliationlist}{\affiliationformat}{\\ }%
}
\makeatother

\lstdefinestyle{pythonstyle}{
    language=Python,
    basicstyle=\ttfamily\footnotesize,
    keywordstyle=\color{blue},
    stringstyle=\color{red!60!black},
    commentstyle=\color{gray}\itshape,
    showstringspaces=false,
    breaklines=true,
    frame=single,
    tabsize=2
}
\lstdefinelanguage{Diff}{
  sensitive=false,
  morekeywords={diff,index,---,+++},               % keywords
  morecomment=[f][\color{red!80!black}]{-},         % lines beginning with '-'
  morecomment=[f][\color{green!60!black}]{+},       % lines beginning with '+'
}
\lstdefinestyle{diffstyle}{
  language=Diff,
  basicstyle=\ttfamily\scriptsize,
  columns=fullflexible,
  frame=single,
  rulecolor=\color{gray!50},
  backgroundcolor=\color{black!2},
  breaklines=true,
  breakatwhitespace=true,
  postbreak=\mbox{\textcolor{gray}{$\hookrightarrow$}\space},
  xleftmargin=1em,
  xrightmargin=1em,
  aboveskip=0.5em,                        % less vertical padding
  belowskip=0.5em,
  commentstyle=\itshape,                  % context lines
  keywordstyle=\bfseries\color{blue!70!black}, % diff/index/@@...
}
\title{{\ha}}

\author[1,2\dagger]{Jenny Zhang}
\author[3\dagger]{Bingchen Zhao}
\author[4\dagger]{Wannan Yang}
\author[6]{Jakob Foerster}
\authorNewLine[1,2,5]{Jeff Clune}
\author[\ddagger]{Minqi Jiang}
\author[7]{Sam Devlin}
\author[6]{Tatiana Shavrina}

\affiliation[1]{University of British Columbia}
\affiliation[2]{Vector Institute}
\affiliation[3]{University of Edinburgh}
\affiliationNewLine[4]{New York University}
\affiliation[5]{Canada CIFAR AI Chair}
\affiliation[6]{FAIR at Meta}
\affiliation[7]{Meta Superintelligence Labs}

\contribution[\dagger]{Work done during internship at Meta}
\contribution[\ddagger]{Work done at Meta}
\abstract{
Self-improving AI systems aim to reduce reliance on human engineering by learning to improve their own learning and problem-solving processes. Existing approaches to recursive self-improvement typically rely on fixed, handcrafted meta-level mechanisms, which fundamentally limit how fast such systems can improve. The Darwin G\"odel Machine (DGM) \citep{zhang2025darwin} demonstrates that open-ended self-improvement is achievable in coding. Starting from a single coding agent, the DGM repeatedly generates and evaluates self-modified variants, forming a growing archive of stepping stones for future improvement. Because both evaluation and self-modification are coding tasks, gains in coding ability can translate into gains in self-improvement ability. However, this alignment does not generally hold beyond coding domains. We introduce \textbf{hyperagents}, self-referential agents that integrate a task agent (which solves the target task) and a meta agent (which modifies itself and the task agent) into a single editable program. Crucially, the meta-level modification procedure is itself editable, enabling metacognitive self-modification, improving not only task-solving behavior, but also the mechanism that generates future improvements. We instantiate this framework by extending DGM to create DGM-Hyperagents (DGM-H). By allowing the improvement procedure to evolve, the DGM-H eliminates the assumption of domain-specific alignment between task performance and self-modification skill, and can potentially support self-accelerating progress on any computable task. Across diverse domains (coding, paper review, robotics reward design, and Olympiad-level math-solution grading), the DGM-H improves performance over time and outperforms baselines without self-improvement or open-ended exploration, as well as prior self-improving systems like DGM. We further show that the DGM-H improves the process by which it generates new agents (e.g., persistent memory, performance tracking), and that these meta-level improvements transfer across domains and accumulate across runs. All experiments were conducted with safety precautions (e.g., sandboxing, human oversight). We discuss what safety entails in this setting and the broader implications of self-improving systems. DGM-Hyperagents offer a glimpse of open-ended AI systems that do not merely search for better solutions, but continually improve their search for how to improve.
}

\date{\today}
\correspondence{Jenny Zhang at \email{jennyzzt@cs.ubc.ca}, Tatiana Shavrina at \email{rybolos@meta.com}}

\metadata[Code]{\url{https://github.com/facebookresearch/Hyperagents}}
\newif\ifcomment
\commenttrue % enable comments
\newcommand{\ha}{\fontfamily{lmr}\selectfont\mdseries\textsc{HyperAgents}\xspace}

\begin{document}

\maketitle

\section{Introduction}
\label{sec:intro}

With appropriate safety considerations, AI systems that can improve themselves could transform scientific progress from a human-paced process into an autonomously accelerating one, thereby allowing society to realize the benefits of technological advances much earlier. Such self-improving AI seeks to continually improve its own learning and task-solving abilities. However, most existing self-improvement architectures rely on a fixed meta agent (i.e., a higher-level system that modifies a base system). This creates a limitation since the base system can only be improved within the boundaries defined by the meta agent's design. Adding a meta-meta system to improve the meta agent does not solve this problem, it merely shifts the issue upward and ultimately leads to an infinite regress of meta-levels. To overcome this limitation and allow a system to modify any part of itself without being constrained by its initial implementation, the system must be self-referential, that is, able to analyze, modify, and evaluate itself \citep{kirsch2022eliminating, zhang2025darwin}. When the mechanism of improvement is itself subject to improvement, progress can in principle become self-accelerating (i.e., the system improves its ability to improve) and potentially unbounded \citep{lu2023arbitrary}.

The Darwin G\"odel Machine (DGM) \citep{zhang2025darwin} demonstrates that open-ended self-improvement is achievable in coding. In the DGM, agents generate and evaluate modifications to their own code, and successful variants are retained in an archive as stepping stones for further improvement. However, the DGM relies on a handcrafted, fixed mechanism to produce self-improvement instructions (\Cref{app:baseline-details}). This mechanism analyzes past evaluation results and the agent's current codebase to generate an instruction directing where the agent should self-improve. This mechanism is not modifiable. Hence, the DGM's capacity for self-improvement is bottlenecked by this fixed instruction-generation step. Despite this handcrafted step, the DGM can still improve at self-improving. Because both evaluation and self-modification are coding tasks, improvements in evaluation performance directly reflect the agent's capacity to generate effective self-modifications. To improve at self-improving, the DGM relies on a limiting assumption: that the skills required to solve the evaluation tasks are the same as those required for effective self-reflection and self-modification. This assumption is unlikely to hold outside coding domains, where task-solving skills may differ substantially from the skills needed to analyze failures, propose effective self-improvements, and implement them.

This work introduces \textit{hyperagents}, self-referential agents that can in principle self-improve for any computable task. Here, an \textit{agent} is any computable program, optionally including calls to foundation models (FMs), external tools, or learned components. A \textit{task agent} solves a given task. A \textit{meta agent} modifies agents and generates new ones. A hyperagent combines the task agent and the meta agent into a single self-referential, modifiable program, such that the mechanism responsible for generating improvements is itself subject to modification. As a result, a hyperagent can improve not only how it solves tasks (i.e., the task agent), but also how it generates and applies future modifications (i.e., the meta agent). Because its self-improvement mechanism is itself modifiable, we call this \textit{metacognitive self-modification}.
We extend the DGM with hyperagents, creating DGM-Hyperagents (DGM-H). The DGM-H retains the open-ended exploration structure of the DGM and extends the DGM with metacognitive self-modification. As with DGM, to support sustained progress and avoid premature convergence, the DGM-H grows an archive of hyperagents by branching from selected candidates, allowing them to self-modify, evaluating the resulting hyperagents, and adding them back to the archive. Because a hyperagent can modify its self-modification process, the DGM-H is not constrained by its initial implementation and can potentially self-improve for any computable task.

Across our experiments, the DGM-H demonstrates substantial and generalizable improvements in both task performance and self-improvement ability. On the Polyglot coding benchmark \citep{gauthier2024polyglot}, the DGM-H achieves gains comparable to the most established prior self-improving algorithm \citep[the Darwin G\"odel Machine,][]{zhang2025darwin}, despite not being handcrafted for coding. Beyond coding, the DGM-H substantially improves performance on paper review \citep{zhao2026apres} and robotics reward design \citep{genesis2024}, with gains transferring to held-out test tasks and significantly outperforming prior self-improving algorithms, which struggle outside coding unless customized. Ablations without self-improvement or without open-ended exploration show little to no progress, highlighting the necessity of each component (\Cref{sec:results-task}). Crucially, the DGM-H learns transferable mechanisms on how to self-improve (e.g., persistent memory, performance tracking) that systematically improve its ability to generate better task or meta agents over time. As a result, meta-level improvements learned by the DGM-H transfer across domains. Specifically, hyperagents optimized in one setting (i.e., paper review and robotics tasks) remain significantly effective at generating improved task agents in a different domain (i.e., Olympiad-level math grading), and an ablation shows that the transferred meta agent, rather than the transferred task agent, accounts for nearly all of this benefit (\Cref{sec:results-meta}). We further show that self-improvements learned by the DGM-H in one setting can compound with continued self-improvement in another setting (\Cref{sec:results-compound}). We note that these results are obtained in finite benchmark settings with fixed task distributions, frozen closed FMs, fixed evaluation protocols, and (in the main experiments) fixed parent selection. They demonstrate useful task-level and meta-level improvements and their transfer across domains; they do not yet establish unbounded or self-accelerating improvement, which we view as the long-term motivation that DGM-H contributes to (\Cref{sec:conclusion}).
We discuss the safety implications of such open-ended self-improving systems and outline practical considerations for responsible deployment in \Cref{sec:safety}. Overall, hyperagents open up the possibility of improving their ability to improve while improving their ability to perform any computable task.

\section{Related Work}
\label{sec:related-work}

\textbf{Open-Endedness.}
Open-endedness refers to the ability of a system to continually invent new, interesting, and increasingly complex artifacts, extending its own frontier of discovery without a fixed objective or predefined end \citep{stanley2017open, hughes2024open}. Recent work has leveraged FMs as proxies for human interestingness and as versatile engines for generating and evaluating novel behaviors across diverse domains \citep{zhangomni, faldoromni}. Building on these advances, recent progress in open-ended learning \citep{huautomated, zoph2017neural, colas2023augmenting, lehman2023evolution} and quality-diversity algorithms \citep{lehman2011evolving, mouret2015illuminating, bradley2023quality, samvelyan2024rainbow, dingquality, pourcel2023aces, coiffard2025overcoming, dharna2025foundation, yuan2026agenticred} has shown that sustained exploration can produce diverse and increasingly capable artifacts across domains ranging from game-playing agents \citep{klissarov2023motif, klissarovmaestromotif, wangvoyager} to scientific discovery \citep{lu2024discovering, lu2024ai, romera2024mathematical, novikov2025alphaevolve, audran2025does} and robotic control \citep{cully2015robots, li2024auto, grillotti2025tabula}. Recent progress has shown that open-ended AI systems capable of continuously generating diverse and increasingly complex artifacts are possible \citep{zhangomni, faldoromni, huautomated}. An important next step is to explore how such systems can achieve compounding improvement. In human scientific and technological progress, advances often build on prior advances not only by producing better artifacts, but also by improving the tools and processes that generate future discoveries, leading to accelerating innovation \citep{good1966speculations, kwa2025measuring}. Inspired by this pattern, we focus on open-ended systems that can improve not only the artifacts they generate, but also the mechanisms by which novelty and progress are produced \citep{clune2019ai, jiang2023general}.

\textbf{Self-improving AI.}
Early theoretical work on self-improving AI dates back to formal models of self-modifying agents \citep{hutter2003gentle}. One prominent example is the G\"odel Machine \citep{schmidhuber2003godel}, which proposes agents that rewrite themselves when provably beneficial, though such approaches remain impractical in real-world settings. Subsequent research explored self-improvement through adaptive neural systems, in which agents modify their own weights or learning dynamics via meta-learning \citep{schmidhuber1993neural, miconi2018differentiable, javed2019meta, beaulieu2020learning, miconi2020backpropamine, irie2022modern, chalvidal2022meta, oh2025discovering}, evolution \citep{stanley2002evolving, lange2023discovering, qiu2025evolution, zhao2025automated}, or self-play \citep{silver2016mastering, silver2017mastering, xia2025agent0, xia2026skillrl}. Notably, \citet{silver2017mastering} use self-play to iteratively improve neural network agents, achieving superhuman performance in domains such as Go and chess, although the underlying learning algorithms themselves remain fixed and human-designed. More recently, FMs have enabled self-improvement through iterative refinement of prompts \citep{fernando2023promptbreeder, wang2025evolving, zhang2025agentic, zhang2025recursive, ye2026meta}, reasoning traces \citep{zelikman2022star, yin2025godel, havrilla2024teaching, zhuge2024gptswarm}, and entire code repositories \citep{zhang2025darwin, wang2025huxley, xia2025live}, as well as through systems that update model weights using self-generated data or interaction \citep{wu2024copilot, zweiger2025self, wen2025unsupervised, wei2025toward}. Among these, the Darwin G\"odel Machine (DGM) \citep{zhang2025darwin} stands out as a practical instantiation of recursive self-improvement in coding domains. However, despite their effectiveness, most existing approaches (including the DGM and its derivatives) rely on fixed, handcrafted meta-level mechanisms (\Cref{app:baseline-details}) that constrain how self-improvement can compound over time and generalize across domains.

\textbf{Self-referential Meta-learning.}
Self-referential meta-learning studies systems that learn to improve the mechanisms by which learning occurs. Prior work has explored this idea in neural networks \citep{kirsch2022eliminating, jackson2024discovering} and evolutionary methods \citep{lu2023arbitrary}. More recently, several works have explored self-referential improvement using FM-based agents \citep{zelikman2024self, robeyns2025self, yin2025godel, zhang2025darwin}. The Darwin G\"odel Machine (DGM) \citep{zhang2025darwin} and its successors \citep{wang2025huxley, xia2025live, weng2026group} instantiate recursive self-improvement through self-modification, primarily in coding domains. However, these approaches improve at improving primarily within coding tasks only.
In the DGM and related systems, a coding agent is tasked with improving itself, and the resulting improved coding agent is then used in subsequent self-improvement steps to generate an even better version of itself. Because both the evaluation task and the self-modification process involve coding, improving the coding agent also enhances the system's ability to carry out future self-improvements. However, this property only holds when the evaluation task and the self-modification task are closely aligned. For example, if the evaluation task were instead poetry writing, improving an agent's poetry-writing ability would not necessarily improve its ability to modify its own code. Prior work therefore relies on an \emph{alignment} between the evaluation task and the skills required for self-improvement. In contrast, hyperagents do not assume such alignment, because the self-modification mechanism is fully modifiable and not tied to any particular task domain. Hence, hyperagents can improve both task performance and the process of improvement itself across any computable task.

\section{Methods}
\label{sec:method}

\begin{figure}[ht]
\centering
\includegraphics[width=\textwidth]{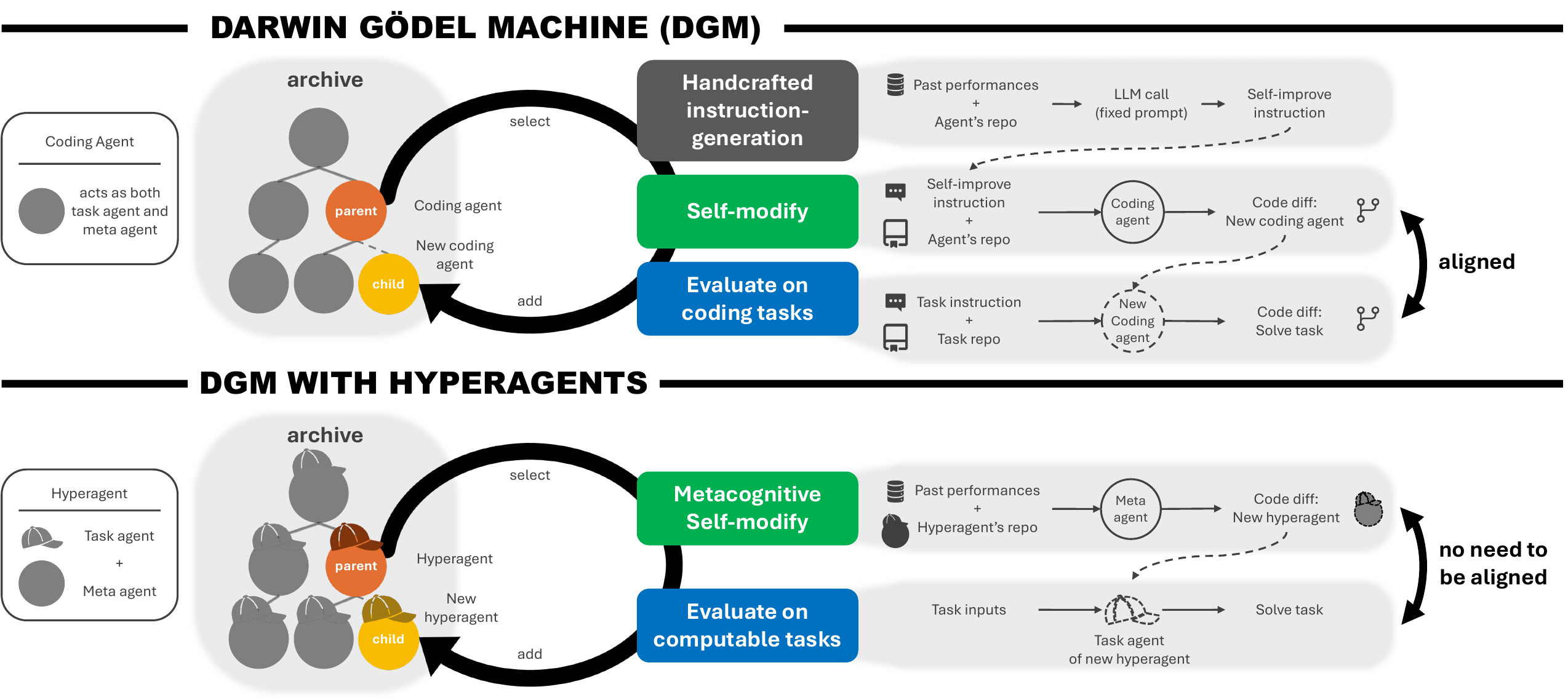}
\caption{\textbf{The Darwin G\"odel Machine with Hyperagents.} The DGM-Hyperagents (DGM-H) extends the Darwin G\"odel Machine (DGM) \citep{zhang2025darwin} beyond coding tasks, enabling agents to improve not only their task performance but also their ability to improve themselves, across any computable task. (Top) In the DGM, a coding agent evolves through open-ended exploration by generating and evaluating self-modified variants, which are stored in an archive of stepping stones. The same coding agent acts as both the task agent (to be evaluated) and the meta agent (to generate modifications). While this design enables compounding gains in coding, the instruction-generation mechanism that drives self-improvement is fixed and handcrafted. Consequently, recursive improvement depends on alignment between coding performance and self-modification ability. (Bottom) In the DGM-H, the task agent and meta agent are combined into a single modifiable program called a hyperagent. This design allows the meta agent itself to be autonomously improved. The system retains the open-ended exploration structure of the DGM while making the meta-level improvement mechanism editable. This enables metacognitive self-modification and supports self-referential improvement across any computable task.}
\label{fig:conceptual}
\end{figure}

We introduce hyperagents, self-referential agents that unify task execution and agent generation into a single modifiable program. A hyperagent can improve not only how it solves tasks but also how it generates future improvements. To enable sustained and accumulating progress, we instantiate hyperagents by building directly on the Darwin G\"odel Machine (DGM) to form DGM-Hyperagents (DGM-H). The DGM provides an open-ended, population-based exploration process that maintains an archive of progressively improving agents, allowing successful variants to serve as stepping stones for future gains. DGM-H retains this open-ended evolutionary structure and extends it by making the entire meta-level modification mechanism editable (\Cref{fig:conceptual}). By allowing agents to modify not only how they solve tasks but also how they improve themselves, the DGM-H has the potential to open-endedly self-improve on any computable task.

\textbf{Agents.} This paper defines an \textit{agent} as any computable program, optionally including calls to FMs, external tools, or learned components. Agents are not restricted to a particular representation (e.g., neural networks or prompts) and may include arbitrary algorithmic logic, memory, and control flow.
A \textit{task agent} is an agent instantiated to solve a set of tasks. Examples include generating code edits for a software repository \citep{gauthier2024polyglot, jimenez2024swebench}, predicting acceptance decisions for research papers \citep{couto2024relevai}, and designing reward functions for robotics environments \citep{maeureka}. Task agents are evaluated empirically on the given task.
A \textit{meta agent} is an agent whose only task is to modify existing agents and generate new ones. Given access to the entire archive of previous agents and evaluations, a meta agent proposes changes intended to improve future performance (including potentially many generations later). Importantly, these changes may target not only task-solving logic but also the meta agent itself, enabling improvements to the procedures by which future modifications are generated.

\textbf{Hyperagents.} A hyperagent is a self-referential agent that integrates a task agent and a meta agent within a single editable program, enabling it to modify not only how it performs tasks but also how it generates future self-modifications. Unlike hierarchical systems with fixed meta-levels, in hyperagents the meta agent is part of the same editable program and can rewrite itself. As a result, a hyperagent can improve both (1) how it solves tasks and (2) how it generates future self-improvements. We use Python, which is Turing-complete \citep{turing1936computable}, and since a hyperagent can edit any code, it has the potential to build any computable machine.

\textbf{Metacognitive self-modification.} In hyperagents, the agent's self-improvement mechanism is itself subject to modification. In addition to improving its performance on a given task, the agent can simultaneously modify the procedures by which it proposes and applies further self-improvements. We refer to this process as \textit{metacognitive self-modification}, in which the hyperagent improves not only the task-performing agent responsible for solving the given task, but also the meta agent that determines how subsequent hyperagents are generated. This characteristic addresses a central limitation of prior self-improving systems \citep{zhang2025darwin, wang2025huxley} by directly enabling improvements to the self-improvement process itself (\Cref{sec:related-work}). Examples of such metacognitive self-modifications are presented in \Cref{sec:results-meta} and \Cref{app:qual-meta}.

\textbf{Darwin G\"odel Machine with Hyperagents.}
Augmenting the original DGM \citep{zhang2025darwin} with hyperagents, we create DGM-Hyperagents (DGM-H). DGM-H employs the open-ended exploration process in the DGM to mitigate premature convergence and avoid getting trapped in local optima. This process maintains an archive of generated hyperagents, initialized with a single hyperagent and expanded over time by continuously accumulating generated variants. The process alternates between two phases: metacognitive self-modification and evaluation. During the metacognitive self-modification phase, selected parent hyperagents from the archive generate modified versions of themselves. Parent selection is probabilistic and proportional to a hyperagent's performance, and inversely proportional to the number of children that successfully compiled, biasing sampling toward hyperagents that perform well and generate strong descendants while preserving exploration (\Cref{app:parent-selection}). During the evaluation phase, each modified hyperagent is empirically evaluated and subsequently added to the archive. In principle, a fully self-referential algorithm should allow modification of every part of itself (including the parent selection and evaluation mechanisms). While we present preliminary results exploring the possibility of automatically improving the parent selection mechanism in \Cref{app:res-parentselect}, the experiments in the main text use a handcrafted parent selection mechanism that is not subject to modification in order to isolate the effects of hyperagent self-modification (\Cref{app:editable-components} summarizes which components are editable and which remain fixed). Overall, DGM-H consists of two interacting components: (1) an open-ended exploration process inherited from the DGM, and (2) an initial hyperagent, which evolves over time through self-generated variants (\Cref{fig:conceptual}, \Cref{app:algo-details}). By extending the DGM to make the meta-level mechanism itself modifiable, the DGM-H generalizes recursive self-improvement beyond coding and, in principle, enables self-accelerating improvement for any computable task.

\section{Experiment Setup}
\label{sec:exp-setup}

The DGM-H is initialized with a single hyperagent built around a frozen FM \citep{brown2020language} with external tool-use capabilities \citep{schick2023toolformer}. The FM autonomously invokes tools as needed at each step. This initial hyperagent is equipped with two tools: a bash tool for executing shell commands, and a specialized tool for inspecting and modifying files. In the initial hyperagent, the task agent receives the task inputs and outputs the response from a single FM call. The meta agent receives the location of the agent's repository, the location of previous evaluation results, and the number of remaining experiment iterations (the experiment budget is a hyperparameter), and is tasked with modifying any part of the given codebase (\Cref{app:initial-agent}). We deliberately keep the implementation of the initial hyperagent (i.e., both the task agent and the meta agent) simple, allowing for a broad search space and avoiding prematurely constraining the range of possible improvements. \Cref{app:fm-hyperparam} details the FMs and the hyperparameters used in each experiment, while \Cref{app:cost-estimate} reports the corresponding cost estimates.

\subsection{Baselines}
\label{sec:baselines}

We compare the Darwin G\"odel Machine with Hyperagents (\textbf{DGM-H}) against a set of baselines designed to isolate the contributions of self-improving meta agents, open-ended exploration, and metacognitive self-modification (i.e., hyperagents) (\Cref{app:baseline-details}).
DGM-H without self-improving meta agents (\textbf{DGM-H w/o self-improve}) fixes the meta agent modifying other hyperagents to the initial meta agent for the entire run. While the system can still generate and evaluate new hyperagents, the procedures used to propose and apply modifications are not themselves modifiable. This baseline replicates the approach of \citep[ADAS,][]{huautomated} in this setting. It tests whether performance gains arise solely from task-level optimization under a fixed meta-level mechanism, or from improvements in the self-improvement process itself.
DGM-H without open-ended exploration (\textbf{DGM-H w/o open-ended exploration}) removes the archive of previously generated hyperagents. Instead of retaining previous variants as stepping stones for future exploration, each newly generated hyperagent replaces its predecessor. After generating a new hyperagent, it is evaluated and automatically becomes the next selected parent. This baseline isolates the role of open-ended, population-based search. Comparing against this baseline tests whether sustained improvement requires accumulating and reusing diverse intermediate solutions.

We also compare against the original implementation of the Darwin G\"odel Machine (\textbf{DGM}) \citep{zhang2025darwin}. The DGM relies on a handcrafted, fixed mechanism to generate self-improvement instructions. Because this instruction-generation mechanism is designed specifically for coding benchmarks and is not modifiable, the DGM's self-improvement capacity in other domains is limited (\Cref{sec:related-work}).
To provide a stronger comparison, we manually customize the DGM's instruction-generation mechanism for the target domains (\textbf{DGM-custom}) (\Cref{app:baseline-details}). This baseline measures how much the DGM relies on human engineering to remain competitive across domains. Comparing the DGM-H against this baseline tests whether automated metacognitive self-modification can outperform human-designed self-improvement mechanisms.
Additionally, we compare against static solutions that have been handcrafted for each domain in prior work.

\subsection{Domains}
\label{sec:domains}

We evaluate our method and baselines across diverse domains (i.e., coding, paper review, robotics reward design, and Olympiad-level math grading) (\Cref{app:domain-details}). To reduce computational cost, for each domain we first evaluate agents on a small subset of the training tasks to estimate overall effectiveness. Only agents that demonstrate sufficient performance are subsequently evaluated on the remaining training tasks. Agents that do not are treated as having zero performance on unevaluated tasks. Domain-specific evaluation protocols are described in detail in the subsequent paragraphs. For domains where we create AI judges to reflect human data (i.e., paper review and Olympiad-level math grading), we construct a validation subset because the AI judges are more likely to overfit to the training data. When a validation subset is defined for a domain, the performance component used in parent selection is measured on the validation set. Otherwise, it is measured on the training set. Each domain includes separate held-out test tasks that are used only for final evaluation.

\textbf{Coding.}
We choose Polyglot \citep{gauthier2024polyglot} as a computationally  cost-efficient coding benchmark for direct comparison with prior work \citep{zhang2025darwin}. In this benchmark, the agent is given a code repository and a natural language instruction describing a desired change, and must modify the repository accordingly. We follow the experimental setup used in the DGM \citep{zhang2025darwin}, including the same training and test splits, no validation set, and the same staged evaluation protocol (i.e., first evaluating each agent on 10 tasks to estimate effectiveness before expanding to 50 additional tasks) (\Cref{app:domain-polyglot}).

\textbf{Paper review.}
This domain evaluates agents on a simulated conference peer review task. For each task, the agent is given the full text of an AI research paper and must predict a binary accept/reject decision. We include paper review to evaluate the DGM-H in a hard-to-verify setting where there is no objective ground truth. Peer review is subjective, and reviewer decisions can vary due to differing priorities and perspectives. We do not aim to change the peer review system, but rather, we study whether hyperagents can automatically learn decision procedures that align with observed human judgments. This domain therefore tests whether agents can learn to reflect human accept/reject decisions, not whether they produce high-quality reviews. The agent outputs a single acceptance decision, and performance is measured by comparing predictions against observed acceptance outcomes. The dataset is drawn from \citet{zhao2026apres}, which constructs a large-scale benchmark from publicly available submissions and acceptance decisions from recent top-tier machine learning conferences. The representative static baseline for this domain is the reviewer agent from the AI-Scientist-v2 \citep{yamada2025ai}. \Cref{app:domain-paperreview} provides full details on the dataset splits (train, validation, and test), the staged evaluation protocol (i.e., first evaluating each agent on a 10-task subset to estimate effectiveness before expanding evaluation to a total of 100 tasks), and the representative baselines for this domain.

\textbf{Robotics reward design.}
This domain evaluates an agent's ability to design reward functions for robotic tasks. We include this domain to move beyond language-only tasks and show that hyperagents can leverage external simulators (e.g., physics engines) and training algorithms (e.g., reinforcement learning (RL)) to produce effective solutions. Given a natural language description of a robotics task, an agent must generate a suitable reward function. This reward function is then used to train a quadruped robot in simulation using RL \citep{genesis2024}. The quality of the agent's solution is measured by the performance of the resulting policy: after training with the generated reward function, we evaluate how well the robot achieves the desired behavior \citep{maeureka}. We use separate training and test tasks. During training, agents are required to generate reward functions that enable the robot to walk forward. For held-out testing, agents must zero-shot generate new reward functions that maximize the robot's torso height. Because reward functions that successfully enable a robot to walk forward do not induce jumping behaviors (the more optimal behavior for maximizing the robot's torso height), this setup evaluates whether a single agent can design suitable reward functions for different robotics tasks. This domain does not have a separate validation task. \Cref{app:domain-genesis} provides full details on the staged evaluation protocol (i.e., first evaluating each agent on 3 repetitions of the training task to estimate effectiveness before expanding evaluation to a total of 6 repetitions), and the representative baselines for this domain.

\textbf{Olympiad-level math grading.}
This domain evaluates an agent's ability to grade solutions to Olympiad-level math problems. This domain is reserved as a held-out meta-evaluation to test whether DGM-H's improvements to its self-improvement process transfer across domains and continue to compound over time. We use IMO-GradingBench \citep{luong2025towards}, which consists of International Mathematical Olympiad (IMO)-level problems paired with candidate solutions and expert human grades. For each task, the agent is given an IMO-level problem, a candidate solution, reference solutions, and grading guidelines to predict a discrete score. Performance is measured by the accuracy of the agent's grades with respect to expert human grades. The representative static baseline for this domain is the ProofAutoGrader from IMO-GradingBench. \Cref{app:domain-imograding} provides full details on the score labels, dataset splits (train, validation, and test), the staged evaluation protocol (i.e., first evaluating the agent on a 10-task subset to estimate effectiveness before expanding evaluation to a total of 100 tasks), and the baselines for this domain.

\section{Results}
\label{sec:results}

For each experiment, we run each method 5 times. We report medians with 95\% bootstrap confidence intervals computed from 1,000 resamples, using the notation \textit{median} (CI: \textit{lower} -- \textit{upper}). In line plots, lines show median performance and shaded regions indicate the confidence intervals (\Cref{fig:res-task,fig:res-transfer,fig:res-transfer-cont}). Bar plots report median performance on held-out test sets, with error bars indicating confidence intervals (\Cref{fig:res-task,fig:res-transfer,fig:res-transfer-cont}). Statistical significance is assessed using the Wilcoxon signed-rank test. Overall, the DGM-H exhibits general self-improvement at both the task and meta levels. Improvements to the task agent transfer to held-out test tasks within each domain, exceeding open-sourced static baselines (\Cref{sec:results-task}). Meta-level improvements transfer across domains, enabling hyperagents to significantly improve their ability to generate better task agents in previously unseen domains (\Cref{sec:results-meta}). Self-improvements learned in one DGM-H run can potentially accelerate learning in subsequent runs and continue to compound as further self-modifications are applied (\Cref{sec:results-compound}). All experiment logs are open-sourced in our codebase.

\subsection{Improving Task Performance}
\label{sec:results-task}

\textbf{The DGM-H can achieve self-improvement in coding comparable to prior self-improving algorithms.} On the Polyglot coding benchmark, we use the same experimental settings as in the DGM (e.g., identical FM parameters, same number of 80 iterations) to enable a direct comparison. Across 5 runs, the DGM-H improves its training performance on the 50-task Polyglot subset from 0.140 (the initial agent) to 0.340 (CI: 0.300 -- 0.380). When evaluated on the full Polyglot benchmark, which consists largely of tasks unseen during training, performance increases from 0.084 (the initial agent) to 0.267 (CI: 0.231 -- 0.280). These improvements are comparable to those reported for the original DGM, which improves from 0.140 to 0.380 on the training subset and from 0.142 to 0.307 on the full benchmark \citep{zhang2025darwin}. Overall, these results show that the DGM-H can effectively self-improve in the coding domain and achieve a similar level of improvement to the original DGM, despite not being handcrafted specifically for coding tasks.

\begin{figure}[ht]
\centering
\includegraphics[width=\textwidth]{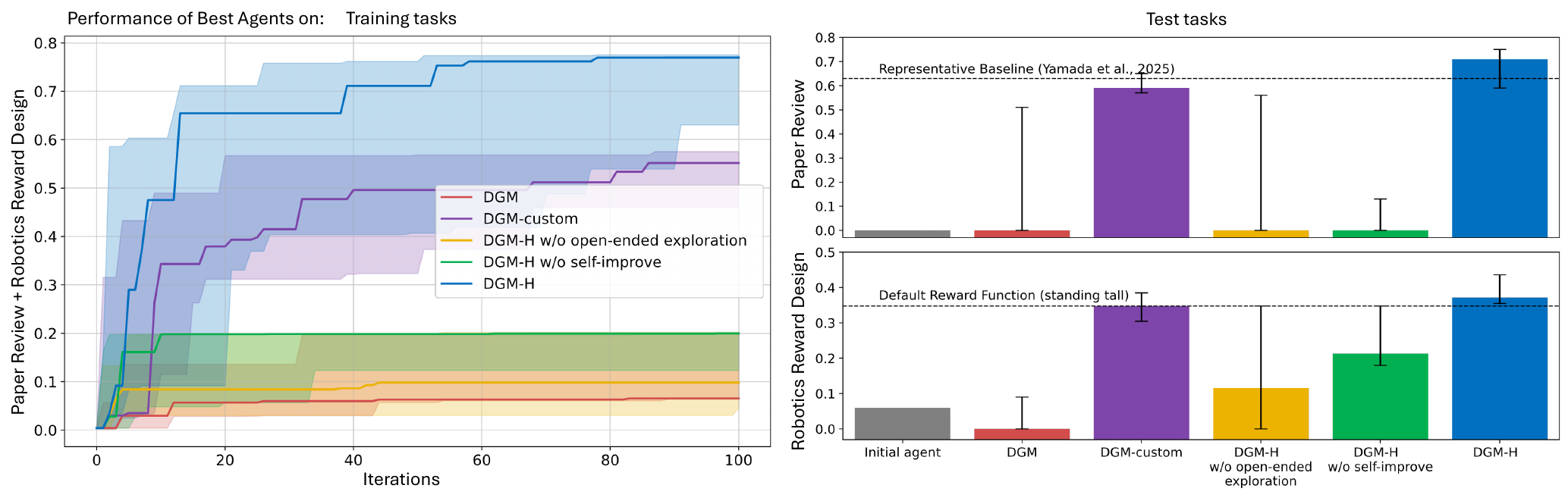}
\caption{\textbf{Metacognitive self-modification and open-ended exploration enable the DGM-H to continue making progress and improve its performance.} (Left) The DGM-H can optimize for diverse tasks within the same run and automatically discovers increasingly better task agents. (Right) The best discovered task agents, selected based on validation or training scores, are evaluated on test tasks in (Top-Right) paper review and (Bottom-Right) robotics reward design. The DGM-H outperforms baselines that lack metacognition (i.e., the original DGM), self-improvement, or open-ended exploration, showing that all components are essential for continual self-improvement.}
\label{fig:res-task}
\end{figure}

Beyond coding, we evaluate the DGM-H on paper review and robotics reward design to assess its ability to self-improve across diverse tasks within a single run (\Cref{app:multi-domain}). If an agent fails the staged evaluation in any domain, it is not evaluated on the full training set for any domain, and a score of zero is assigned for all remaining tasks. For example, when jointly optimizing the paper review and robotics reward design domains within the same experiment run, failure in the staged evaluation of either domain (e.g., correctly predicting none of the paper reviews in a smaller training subset, or failing to generate any compilable reward function) results in the agent not being evaluated on the full training set for either domain. Only the best agents, selected via validation scores (or training scores when validation tasks do not exist), are evaluated on the test set.
All runs start from the same initial agent, which directly outputs the response from a single FM call (\Cref{app:initial-agent}). This initial agent achieves a test-set score of 0.0 on paper review and 0.060 on robotics reward design (\Cref{fig:res-task}). In paper review, the initial agent fails to produce outputs in the required format because no task-specific parsing or post-processing is provided, resulting in a score of 0.0. This reflects an interface failure rather than zero paper-review ability of the underlying FM, so part of the early improvement consists of learning the task interface. In robotics reward design, the initial agent typically generates reward functions that either fail to compile (e.g., due to invalid environment variables) or do not yield effective policies. However, the initial agent occasionally produces a minimally functional reward, leading to a small but non-zero average score. Starting from this initial agent, each method is run for 100 iterations.

\textbf{The original DGM requires domain-specific customization to generalize beyond coding.} When applied in the form designed specifically to optimize coding agents, the original DGM achieves test-set performance of 0.0 (CI: 0.0 – 0.510) on paper review and 0.0 (CI: 0.0 – 0.090) on robotics reward design. This outcome is expected, as the original DGM's instruction-generation mechanism is explicitly tailored to coding tasks and to improving coding agents (\Cref{app:baseline-details}). However, when manually customized for each new domain, DGM-custom achieves substantially stronger results: 0.590 (CI: 0.570 – 0.650) on paper review and 0.348 (CI: 0.305 – 0.385) on robotics reward design. DGM-custom significantly outperforms the original DGM in both domains (p < 0.05) (\Cref{fig:res-task}). These results highlight both the strength of the DGM within its intended domain and the importance of domain-specific manual customization for extending its self-improvement capabilities beyond coding in its original implementation.

\textbf{Metacognitive self-modification enables sustained task improvement across domains.} The DGM-H substantially improves task performance over time in both domains, demonstrating effective improvement of the task agent. Importantly, these gains transfer to held-out test sets, indicating that improvements reflect general competence rather than overfitting to training tasks. On test tasks, DGM-H improves paper review performance from 0.0 (the initial agent) to 0.710 (CI: 0.590 -- 0.750) (\Cref{app:best-agents-paperreview}). Several top-performing agents outperform the open-sourced static baseline, which achieves 0.630 \citep{yamada2025ai}. On robotics reward design, DGM-H improves performance from 0.060 (the initial agent) to 0.372 (CI: 0.355 -- 0.436) (\Cref{app:best-agents-genesis}), surpassing the default reward function that directly optimizes the evaluation metric (0.348).

Qualitatively, the DGM-H improves task agents by moving beyond surface-level prompt tweaks toward structured, reusable decision machinery. In paper review, it shifts from superficial behavioral instructions (e.g., adopting a ``rigorous'' persona) to explicit multi-stage evaluation pipelines with checklists, decision rules, and clearly defined criteria, resulting in more consistent and higher-quality judgments (\Cref{app:qual-task}). In robotics reward design, the DGM-H incrementally builds and refines an internal knowledge base of environment constraints, valid state variables, and reward-scaling heuristics, eliminating compilation failures and reducing reward misspecification. The DGM-H accumulates and refines domain knowledge (e.g., environment documentation, grading criteria) and integrates it into increasingly sophisticated decision frameworks, enabling complex, consistent task behavior without manual, domain-specific engineering (\Cref{app:qual-task}).

The DGM-H significantly outperforms the original DGM in both domains (p < 0.05), reflecting the limitations of the original DGM's fixed instruction-generation mechanism when applied to non-coding tasks (\Cref{fig:res-task}). The DGM-H achieves higher median performance than DGM-custom on the test tasks in both domains, although the difference is not statistically significant (p > 0.05) (\Cref{fig:res-task}). Notably, in the robotics reward design test task (i.e., design a reward function that trains the robot to maximize its torso height), the DGM-H almost always generates task agents that design reward functions inducing jumping behaviors, rather than getting stuck in the local optimum of standing tall. In contrast, DGM-custom often generates task agents that fail to escape this local optimum. This qualitative difference suggests that metacognitive self-modification can match or exceed handcrafted, domain-specific mechanisms.

\textbf{Both metacognitive self-modification and open-ended exploration are necessary for sustained improvement.}
DGM-H w/o self-improve \citep{huautomated} shows little to no improvement, achieving test-set performance of 0.0 (CI: 0.0 -- 0.130) on paper review and 0.213 (CI: 0.180 -- 0.348) on robotics reward design. The DGM-H significantly outperforms this baseline in both domains (p < 0.05) (\Cref{fig:res-task}), demonstrating that the observed gains in the DGM-H are not solely due to task-level optimization under a fixed meta agent, but also arise from improvements in the self-modification process itself (i.e., the meta agent).
Similarly, DGM-H w/o open-ended exploration fails to achieve meaningful improvement, with test-set performance remaining at 0.0 (CI: 0.0 -- 0.560) on paper review and 0.116 (CI: 0.0 -- 0.348) on robotics reward design. The DGM-H significantly outperforms this baseline in both domains (p < 0.05) (\Cref{fig:res-task}), highlighting the importance of open-ended, population-based search to create stepping stones for future improvements.

\subsection{Improving the Ability to Improve}
\label{sec:results-meta}

DGM-H's superior performance to DGM-H w/o self-improve shows that DGM-H improves the meta agent (i.e., its ability to self-modify and generate new agents) (\Cref{sec:results-task}). This section investigates whether these meta-level improvements are general rather than domain-specific. Specifically, we evaluate whether self-improvement strategies learned in one setting transfer to and accelerate learning in a different domain.

\textbf{Improvement@k metric.} To quantify a meta agent's ability to generate improved task agents, we introduce the improvement@k (imp@k) metric. Given an initial meta agent $M$, an initial task agent $A$, an agent-generation algorithm (e.g., DGM or DGM-H variants), and a fixed task, $M$ is allowed to generate up to $k$ new task agents from $A$ and its descendants using the specified algorithm. We define imp@k as the improvement in test performance from the initial task agent $A$ to the best-performing generated task agent (selected based on validation score) (\Cref{app:genk-metric}). Intuitively, imp@k measures how effectively a meta agent can produce improved variants within a limited number of iterations. To isolate the meta agent's ability to improve task agents from its ability to improve itself, we hold the meta agent fixed throughout the process (i.e., the agent responsible for generating new task agents is always the initial meta agent $M$), yielding DGM w/o self-improve and DGM-H w/o self-improve as the agent-generation algorithms. All experiments in this section run for 50 iterations, measuring imp@50.

\begin{figure}[ht]
\centering
\includegraphics[width=\textwidth]{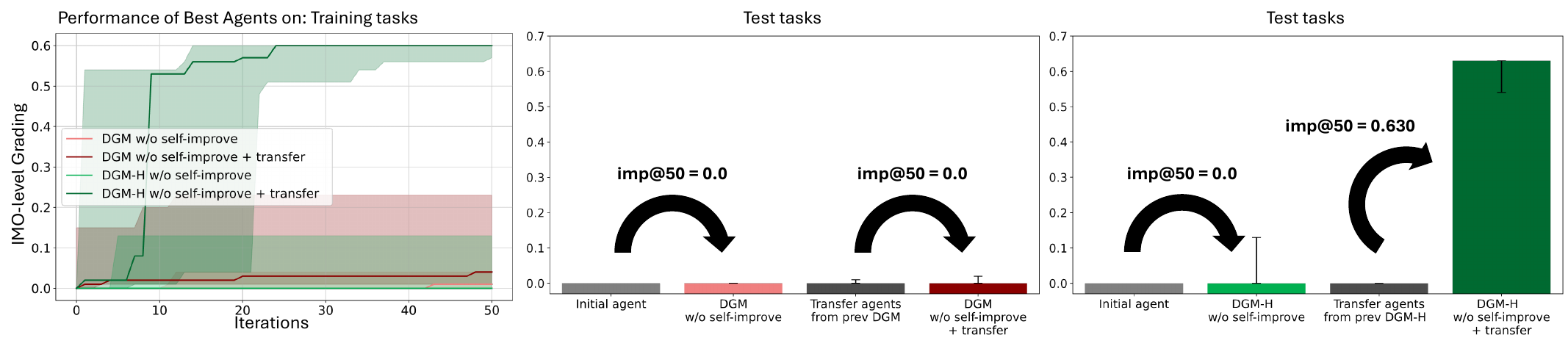}
\caption{\textbf{Self-improvement strategies learned by the DGM-H in one setting transfer to and accelerate learning in a different setting.} We measure an agent's ability to generate improved agents using imp@50, which takes as input a starting agent, an agent-generation algorithm, and an evaluation task. The algorithm is run for 50 iterations starting from the given agent, and imp@50 is defined as the performance gain of the best generated agent over the starting agent on the task. (Left, DGM w/o self-improve and DGM-H w/o self-improve) On Olympiad-level math grading, starting from the initial agent, both DGM w/o self-improve and DGM-H w/o self-improve achieve little to no improvement, showing that the initial agent has limited ability to generate better agents. (Left, DGM w/o self-improve + transfer) Starting from transfer agents, DGM w/o self-improve also achieves little improvement, showing that the original DGM does not learn transferable meta-level improvements. (Left, DGM-H w/o self-improve + transfer) In contrast, starting from transfer hyperagents, DGM-H w/o self-improve achieves substantial improvement, showing that hyperagents enable DGM-H to acquire transferable self-improvement strategies. (Middle) Regardless of starting from an initial agent or a transfer agent, DGM w/o self-improve yields imp@50 $\approx$ 0, showing that the original DGM does not improve the meta-level skill of generating improved agents. (Right) Starting from a transfer hyperagent, DGM-H w/o self-improve achieves large positive imp@50, showing that the self-improvement strategies learned by the DGM-H are general and transferable, and that the DGM-H improves its ability to improve.}
\label{fig:res-transfer}
\end{figure}

\textbf{The initial meta agent has little to no ability to generate improved task agents.}
Taking the same initial meta and task agents as in the above experiments (\Cref{sec:exp-setup}), we evaluate the initial meta agent's ability to generate improved task agents on Olympiad-level math grading. The initial task agent achieves a test-set score of 0.0, as it fails to produce outputs in the required format due to the lack of task-specific parsing or post-processing.
After 50 iterations of DGM w/o self-improve, during which the initial meta agent is held fixed and used to generate new task agents starting from an initial task agent, the best generated task agents (across 5 repeated runs) achieve a median test score of 0.0 (CI: 0.0 – 0.0). Accordingly, imp@50 for the initial meta agent under DGM w/o self-improve is 0.0 (CI: 0.0 – 0.0) on Olympiad-level math grading.
While DGM w/o self-improve relies on a fixed handcrafted instruction-generation mechanism, DGM-H w/o self-improve does not. DGM w/o self-improve uses a handcrafted instruction-generation mechanism to specify which parts of the agent's codebase should be modified, whereas DGM-H w/o self-improve allows the meta agent to autonomously decide which parts of the codebase to modify.
After 50 iterations of DGM-H w/o self-improve, again holding the initial meta agent fixed and generating task agents from the same initial task agent, the best generated task agents (across 5 repeated runs) achieve a median score of 0.0 (CI: 0.0 – 0.130). Accordingly, imp@50 for the initial meta agent under DGM-H w/o self-improve is 0.0 (CI: 0.0 – 0.130) (\Cref{fig:res-transfer}).
These results show that, on Olympiad-level math grading, the initial meta agent has little to no ability to generate improved task agents, regardless of the agent-generation algorithm used (e.g., DGM or DGM-H).

\textbf{Meta agents optimized in prior DGM-custom runs have little to no ability to generate improved task agents when transferred to a new domain.}
We evaluate whether self-improvements learned in one setting using the DGM can transfer to new domains. We consider agents optimized in prior DGM-custom runs on the paper review and robotics reward design tasks (\Cref{sec:results-task}). From each run, we select a transfer agent based on the maximum performance gain achieved by its descendants, measured relative to the agent itself and discounted by lineage depth (\Cref{app:transfer-selection}). We apply discounting to favor agents that reliably produce strong improvements within fewer self-modification steps, since improvements occurring many iterations later are less directly attributable to the original agent. We choose transfer agents from DGM-custom rather than from DGM because DGM-custom achieved larger improvements in the previous settings (\Cref{sec:results-task}).
For transfer, we carry over the entire agent implementation from the source domain, including both the meta agent and the task agent. Concretely, the transfer meta agent is used as the fixed agent responsible for generating new task agents in the target domain, and the transfer task agent serves as the initial task agent from which new variants are generated.
We evaluate whether the transfer meta agents can generate improved task agents in a new domain, Olympiad-level math grading. The transfer task agents (one from each of 5 repeated runs) achieve a test-set score of 0.0 (CI: 0.0 -- 0.010). As expected, this performance is low because the agents were optimized for paper review and robotics reward design, which differ substantially from the math grading task.
After 50 iterations of DGM w/o self-improve starting from each transfer task agent and holding the transfer meta agent fixed, the best generated task agents (selected based on validation score) achieve a score of 0.0 (CI: 0.0 -- 0.020). This corresponds to an imp@50 of 0.0 (CI: 0.0 -- 0.010) (\Cref{fig:res-transfer}), showing that the transfer meta agents from DGM-custom runs cannot generate improved task agents in a previously unseen domain.
When using DGM w/o self-improve as the agent-generation algorithm, imp@50 for the transfer agents is similar to that of the initial agent (both close to zero). This shows that the task performance improvements observed with DGM-custom rely on domain-specific customizations, rather than on general improvements in the meta ability to generate improved task agents.

\textbf{Hyperagents optimized in prior DGM-H runs demonstrate transferable ability to generate improved task agents in a new domain.}
We repeat the transfer experiments for the DGM-H to assess whether hyperagents enable the learning of general self-improvement mechanisms. From prior DGM-H runs on paper review and robotics reward design (\Cref{sec:results-task}), we select one transfer hyperagent per run using the same lineage-discounted criterion described above (\Cref{app:transfer-selection}). The transfer hyperagents (one from each of 5 repeated runs) achieve an initial test-set score of 0.0 (CI: 0.0 -- 0.0), reflecting the domain shift. After 50 iterations of DGM-H w/o self-improve starting from the transfer hyperagents, the best generated agents (selected based on validation score) achieve a test-set score of 0.630 (CI: 0.540 -- 0.630). This corresponds to an imp@50 of 0.630 (CI: 0.540 -- 0.630) (\Cref{fig:res-transfer}). These results show that transfer hyperagents can generate improved agents in a previously unseen domain. When using DGM-H w/o self-improve as the agent-generation algorithm, imp@50 for the transfer agents is significantly higher than imp@50 for the initial agent (p < 0.05). This indicates that the transfer agents are substantially more effective at generating improved agents, and that the meta-improvements learned through DGM-H in one run are general and transferable, accelerating learning in a different domain.

\textbf{The transferred meta agent accounts for most of the transfer benefit.}
To disentangle the contributions of the two components of a transferred hyperagent, we transfer only the task agent (with the initial meta agent held fixed) or only the meta agent (starting from the initial task agent, with the transferred meta agent held fixed), and compare against transferring neither and both, all using DGM-H w/o self-improve for 50 iterations (\Cref{app:transfer-ablation}). Transferring only the meta agent recovers essentially all of the full-transfer benefit, with an imp@50 of 0.620 (CI: 0.550 -- 0.660) compared to 0.630 (CI: 0.540 -- 0.630) when transferring both, whereas transferring only the task agent yields little improvement, with an imp@50 of 0.100 (CI: 0.0 -- 0.560), close to transferring neither, which yields 0.0 (CI: 0.0 -- 0.130). Hence, the transferred meta agent (i.e., the procedure that generates task-agent modifications) carries most of the transferable self-improvement capability.

We qualitatively attribute the observed transfer gains to a set of general-purpose meta-level capabilities that the DGM-H autonomously acquires during prior runs. In particular, the transfer hyperagents have features such as performance tracking and persistent memory, which allow them to reason about improvement as an ongoing process rather than as isolated code edits (\Cref{app:qual-meta}). As a result, even when transferred to an unseen domain, these hyperagents can quickly self-improve and make meaningful progress (\Cref{fig:res-transfer}). This contrasts with DGM transfer agents, whose gains rely on domain-specific customizations and do not improve the underlying agent-generation process itself. These qualitative observations show that the DGM-H learns how to improve, yielding general and transferable self-improvement capability.

One example is DGM-H's autonomous development of performance tracking infrastructure. Rather than relying on isolated evaluation outcomes, the hyperagent records, aggregates, and compares performance metrics across iterations, identifying which changes yield sustained improvements and which lead to regressions. This performance tracking supports informed decision-making, such as prioritizing promising iterations and avoiding unproductive modification directions. The snippet below shows an automatically introduced performance tracker that logs and organizes metrics across iterations:
\begin{lstlisting}[language=Python]
class PerformanceTracker:
    """Tracks performance metrics across agent generations."""

    def __init__(self, tracking_file: str = "./outputs/performance_history.json"):
        self.tracking_file = tracking_file
        self.history = self._load_history()

    def record_generation(self, generation_id: int, domain: str,
                          score: float, metadata: dict = None):
        """Record performance for a generation."""
        entry = {
            "generation_id": generation_id,
            "domain": domain,
            "score": score,
            "timestamp": datetime.now().isoformat(),
            "metadata": metadata or {}
        }
        self.history.append(entry)
        self._save_history()

    def get_improvement_trend(self, domain: str = None, window: int = 5):
        """Calculate improvement trend using moving average."""
        filtered = self.history
        if domain:
            filtered = [h for h in self.history if h.get('domain') == domain]

        if len(filtered) < window * 2:
            return None

        recent_avg = sum(h['score'] for h in filtered[-window:]) / window
        older_avg = sum(h['score'] for h in filtered[-window*2:-window]) / window

        return recent_avg - older_avg  # Positive if improving

    def get_statistics(self, domain: str = None):
        """Get comprehensive statistics."""
        scores = [h['score'] for h in filtered]
        return {
            "total_generations": len(filtered),
            "best_score": max(scores),
            "worst_score": min(scores),
            "average_score": sum(scores) / len(scores),
            "improvement_trend": self.get_improvement_trend(domain)
        }
\end{lstlisting}

Another example is DGM-H's innovation of persistent memory, which enables learning to accumulate across iterations. Instead of merely logging numerical scores, the hyperagent stores synthesized insights, causal hypotheses, and forward-looking plans (e.g., identifying which generations performed best, diagnosing over-corrections, and proposing how to combine successful strategies). This memory is actively consulted during subsequent self-modification steps, allowing later generations to build on earlier discoveries and avoid repeating past mistakes. This is an example of a stored memory entry:
\begin{lstlisting}[language=Python]
{
  "best_performers_analysis": {
    "value": "Best Performing Generations:\n\nPaper Review:\n
      - Gen55: 63% acc, 25% accept rate, 38% accept recall, 88% reject recall (too harsh)\n
      - Gen64: 61% acc, 53% accept rate, 64% accept recall, 58% reject recall (BEST BALANCE)\n\n
      Genesis Go2Walking:\n- Gen62: 0.802 fitness (BEST)\n- Gen63: 0.798 fitness\n\n
      Key Insight: Gen55 has best accuracy but is too harsh. Gen64 improved balance
      but lower accuracy. Need to combine gen55's critical reasoning with gen64's balance.",
    "timestamp": "2026-01-07T00:43:27.550275"
  },
  "gen66_results": {
    "value": "Gen 66 Results (Latest): Paper Review - 52% accuracy (WORSE than gen64's 61%),
      94% accept rate (EXTREME accept bias). CRITICAL ISSUE: Paper review went backwards
      - gen65 changes over-corrected.",
    "timestamp": "2026-01-07T01:03:41.176643"
  },
  "iteration_66_improvement": {
    "value": "Iteration 66 Improvement: CORRECTED paper review prompt after gen65 over-correction.
      New approach: (1) Clear REJECT criteria with specific triggers, (2) Decision Protocol
      with explicit thresholds (~40%/~60%), (3) Borderline guidance defaults to REJECT.
      Target: 62-64% accuracy with 55-65% reject rate.",
    "timestamp": "2026-01-07T01:06:35.559820"
  }
}
\end{lstlisting}

\subsection{Compounding Self-Improvements}
\label{sec:results-compound}

\begin{figure}[ht]
\centering
\includegraphics[width=0.9\textwidth]{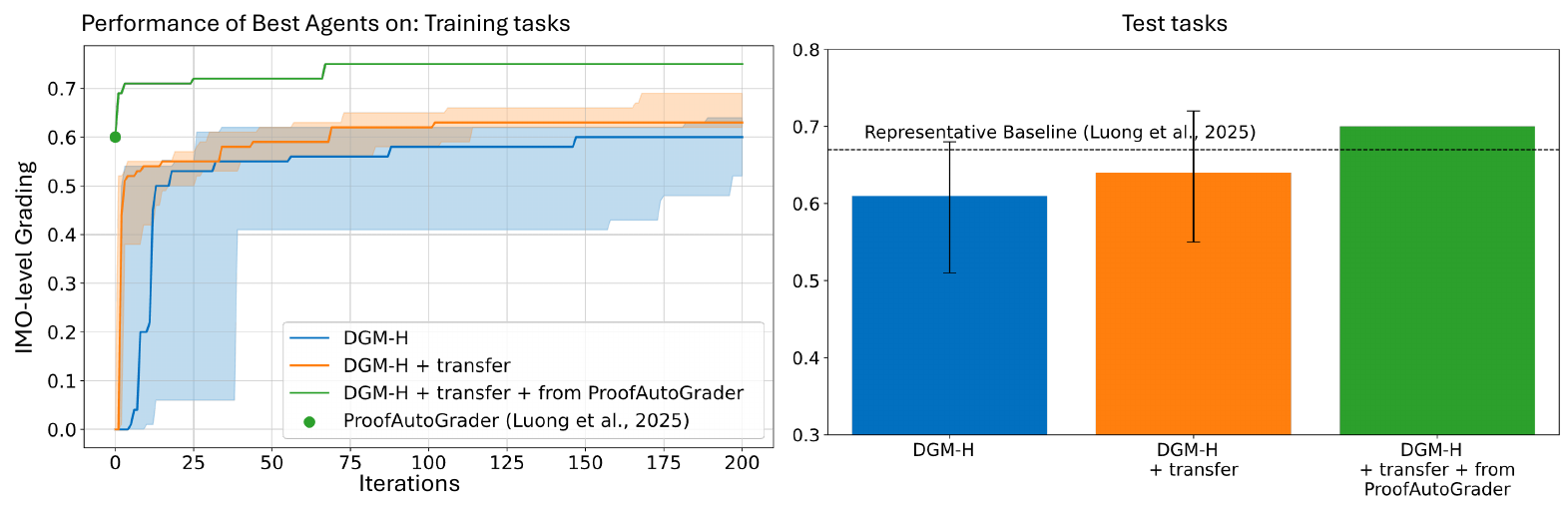}
\caption{\textbf{Self-improvements learned by the DGM-H accumulate across domains and runs.} We continue running DGM-H on Olympiad-level math grading, starting from transfer hyperagents obtained in prior DGM-H runs, and compare this against DGM-H initialized from the initial agent and from ProofAutoGrader. (Left) Initializing from transferred hyperagents leads to faster progress and higher final performance than initializing from the initial agent, indicating that previously learned self-improvements remain useful and continue to compound in a new domain. (Right) DGM-H initialized from a transferred agent and ProofAutoGrader achieves the highest test performance, surpassing the representative baseline.}
\label{fig:res-transfer-cont}
\end{figure}

We investigate whether self-improvements learned by DGM-H in one setting continue to accumulate when DGM-H is run in a different setting. From prior DGM-H runs on the paper review and robotics reward design tasks (\Cref{sec:results-task}), we select transfer hyperagents using the same selection mechanism described earlier (\Cref{sec:results-meta}, \Cref{app:transfer-selection}). We then evaluate their ability to continue self-improving in a new domain, Olympiad-level math grading. After 200 iterations of DGM-H starting from these transfer agents (DGM-H + transfer), the best generated agents (selected based on validation score) achieve a test-set score of 0.640 (CI: 0.550 -- 0.720). Under the same experimental setup, DGM-H starting from the initial agent achieves a best test-set score of 0.610 (CI: 0.510 -- 0.680). Although the difference between DGM-H + transfer and DGM-H is not statistically significant (p > 0.05), DGM-H + transfer achieves a higher median performance and higher confidence intervals than DGM-H starting from the initial agent (\Cref{fig:res-transfer-cont}). Notably, improvements at higher performance levels are increasingly difficult due to saturation effects (e.g., increasing performance from 0.7 to 0.8 is typically more challenging than from 0.0 to 0.1), making these gains meaningful despite their modest absolute magnitude. These results suggest that DGM-H's self-improvements are reusable and can potentially accumulate across runs, supporting the possibility of compounding self-improvement over time.

The representative static baseline for Olympiad-level math grading from IMO-GradingBench is ProofAutoGrader \citep{luong2025towards}. We initialize the DGM-H with ProofAutoGrader as the task agent and a transfer meta agent obtained from a prior DGM-H run (on paper review and robotics reward design), and then continue optimizing for Olympiad-level math grading. After 200 iterations, the best discovered agent achieves a test-set score of 0.700, outperforming ProofAutoGrader's score of 0.670 (\Cref{fig:res-transfer-cont}). We then evaluate both the best discovered agent and ProofAutoGrader on the full IMO-GradingBench to obtain a more accurate estimate of the improvement. On the full IMO-GradingBench, the DGM-H improves ProofAutoGrader's accuracy from 0.561 to 0.601, and lowers the mean absolute error from 0.178 to 0.175 (\Cref{app:res-imograders}). We open-source this artifact to support future research and development (\Cref{app:best-agents-imograding}). These results show that the DGM-H can build on strong existing solutions and further improve their performance.

\section{Safety Discussion}
\label{sec:safety}

The DGM-Hyperagents (DGM-H) introduces distinct safety considerations due to its ability to autonomously modify its own behavior and improvement mechanisms over time. In this work, all experiments are conducted under strict safety constraints. In particular, agent-generated code is executed within carefully sandboxed environments (i.e., Docker containers with a restricted workspace and no general external network access) with enforced resource limits (e.g., timeouts, memory and compute limits). Test tasks are unavailable to the task and meta agents and are used only for final evaluation, which prevents direct test leakage but does not eliminate gaming of the training proxy. All prompts, tool calls, outputs, code diffs, failures, and evaluation results are logged, and these logs are released in our open-sourced repository for auditing. These measures are designed to prevent unintended side effects, contain failures, and ensure that self-modifications remain confined to the intended experimental scope. Moreover, evaluation is performed using predefined tasks and metrics, and human oversight is maintained throughout all experiments.

\textbf{Potential to evolve faster than human oversight.}
As AI systems gain the ability to modify themselves in increasingly open-ended ways, they can potentially evolve far more rapidly than humans can audit or interpret. At the cusp of such explosive capability growth, it becomes necessary to reconsider the roles that AI systems play in society \citep{bengio2024managing}. Rather than framing safety solely in terms of absolute guarantees or full interpretability, a central challenge lies in balancing the potential of AI as a catalyst for human progress and well-being (e.g., automating scientific discovery) with the degree of trust humans are willing to place in these systems (e.g., delegating decisions or actions without requiring continuous human verification), while minimizing the many potential risks and downsides \citep{clune2019ai, ecoffet2020open, bengio2024managing, weston2025ai}. This balance is shaped by factors such as transparency and controllability.

While the DGM-H operates within safe research boundaries (e.g., sandboxing, controlled evaluations), these safeguards may become increasingly strained or infeasible as self-improving systems grow more capable. We discuss additional safety considerations in \Cref{app:safety}. We proactively include this discussion to encourage broader engagement with what safety means for open-ended self-improving AI systems \citep{clune2019ai, ecoffet2020open, sheth2025safety}. This includes ongoing discussion about appropriate levels of trust, oversight, and transparency, and societal deliberation about which benefits these systems should prioritize when deployed.

\section{Limitations and Conclusion}
\label{sec:conclusion}

This work introduces hyperagents and incorporates them into the Darwin G\"odel Machine (DGM) to form DGM-Hyperagents (DGM-H). DGM-H is a general self-improvement framework that open-endedly evolves an archive of self-improving hyperagents for any computable task, enabling the system to improve both task performance and its own self-improvement mechanism. Across diverse domains, the DGM-H produced substantial and generalizable gains in task performance while also improving its ability to generate improvements, with these meta-level gains transferring across domains and compounding across runs.

Our results suggest that self-improvements can compound across different experimental settings, but this version of DGM-H has limitations that constrain truly unbounded progress.
First, it operates with a fixed task distribution. One direction is to co-evolve the task distribution by generating new tasks and curricula that adapt to the agent's capabilities \citep{clune2019ai, zhangomni, faldoromni, bolton2025sima}.
Second, components of the open-ended exploration loop (e.g., parent selection, evaluation protocols) remain fixed. Although hyperagents can modify their self-improvement mechanisms, they cannot alter the outer process that determines which agents are selected or how they are evaluated. Keeping these components fixed improves experimental stability and safety, but limits full self-modifiability. Enabling hyperagents to modify these outer-loop components and adapt their own search strategy and evaluation process is another promising direction for future work. Our preliminary results suggest such extensions are feasible (\Cref{app:res-parentselect}). We discuss further limitations in \Cref{app:limitations}.

DGM-H demonstrates that open-ended self-improvement can be made practical across diverse domains. Provided sufficient safety considerations are worked out, the DGM-H suggests a path toward safe self-accelerating systems that not only search for better solutions, but continually improve their ability to self-improve.

\section*{Acknowledgments}
We thank Andrew Budker and Ricardo Silveira Cabral for supporting this work, and Alisia Lupidi, Chenxi Whitehouse, John Quan, Lisa Alazraki, Lovish Madaan, Lucia Cipolina-Kun, Mattia Opper, Michael Dennis, Parth Pathak, Rishi Hazra, Roberta Raileanu, Sandra Lefdal, Shashwat Goel, Shengran Hu, Timon Willi, Tim Rockt\"aschel, and Yoram Bachrach for insightful discussions and feedback.

\section*{Author Contributions}
Jenny Zhang led the conceptualization of the study, conducted the experiments, and wrote the manuscript. Bingchen Zhao and Wannan Yang contributed to experimental design and execution. Jakob Foerster, Jeff Clune, Minqi Jiang, Sam Devlin, and Tatiana Shavrina provided feedback on the methodology and manuscript. All authors reviewed and approved the final manuscript.

\bibliographystyle{plainnat}
\bibliography{paper}

\clearpage

\beginappendix

\section*{Table of Contents}
\startcontents[sections]
\printcontents[sections]{l}{1}{\setcounter{tocdepth}{2}}
\clearpage

\crefname{appendix}{Appendix}{Appendices}
\Crefname{appendix}{Appendix}{Appendices}
\crefalias{section}{appendix}
\crefalias{subsection}{appendix}
\crefalias{subsubsection}{appendix}

\section{Algorithmic details}
\label{app:algo-details}

This appendix provides additional algorithmic details for the DGM-Hyperagents (DGM-H). We first describe the implementation of the initial hyperagent, including the tools and prompts available to the initial task and meta agents (\Cref{app:initial-agent}). We then detail the parent selection mechanism used during open-ended exploration, which balances exploitation of high-performing agents with continued exploration of the archive (\Cref{app:parent-selection}). Finally, we present pseudocode for DGM-H (\Cref{app:pseudocode}).

\subsection{Initial Agent}
\label{app:initial-agent}

We present the details of the tools available to the initial hyperagent and its prompts (\Cref{sec:exp-setup}).

Initial task agent prompt:
\begin{lstlisting}
instruction = f"""You are an agent.
Task input:
```
{inputs}
```
Respond in JSON format with the following schema:
<json>
{{
    "response": ...
}}
</json>"""
\end{lstlisting}

Initial meta agent prompt:
\begin{lstlisting}
instruction = f"Modify any part of the codebase at `{repo_path}`."
\end{lstlisting}

Information of the given bash tool:
\begin{lstlisting}
def tool_info():
    return {
        "name": "bash",
        "description": """Run commands in a bash shell
* When invoking this tool, the contents of the "command" parameter does NOT need to be XML-escaped.
* You don't have access to the internet via this tool.
* You do have access to a mirror of common linux and python packages via apt and pip.
* State is persistent across command calls and discussions with the user.
* To inspect a particular line range of a file, e.g. lines 10-25, try 'sed -n 10,25p /path/to/the/file'.
* Please avoid commands that may produce a very large amount of output.
* Please run long lived commands in the background, e.g. 'sleep 10 &' or start a server in the background.""",
        "input_schema": {
            "type": "object",
            "properties": {
                "command": {
                    "type": "string",
                    "description": "The bash command to run."
                }
            },
            "required": ["command"]
        }
    }
\end{lstlisting}

Information of the given edit tool:
\begin{lstlisting}
def tool_info():
    return {
        "name": "editor",
        "description": """Custom editing tool for viewing, creating and editing files
* State is persistent across command calls and discussions with the user
* If `path` is a file, `view` displays the result of applying `cat -n`. If `path` is a directory, `view` lists non-hidden files and directories up to 2 levels deep
* The `create` command cannot be used if the specified `path` already exists as a file
* If a `command` generates a long output, it will be truncated and marked with `<response clipped>`
* The `undo_edit` command will revert the last edit made to the file at `path`
\nNotes for using the `str_replace` command:
* The `old_str` parameter should match EXACTLY one or more consecutive lines from the original file. Be mindful of whitespaces!
* If the `old_str` parameter is not unique in the file, the replacement will not be performed. Make sure to include enough context in `old_str` to make it unique
* The `new_str` parameter should contain the edited lines that should replace the `old_str`""",
        "input_schema": {
            "type": "object",
            "properties": {
                "command": {
                    "type": "string",
                    "enum": ["view", "create", "str_replace", "insert", "undo_edit"],
                    "description": "The commands to run. Allowed options are: `view`, `create`, `str_replace`, `insert`, `undo_edit`."
                },
                "file_text": {
                    "description": "Required parameter of `create` command, with the content of the file to be created.",
                    "type": "string"
                },
                "insert_line": {
                    "description": "Required parameter of `insert` command. The `new_str` will be inserted AFTER the line `insert_line` of `path`.",
                    "type": "integer"
                },
                "new_str": {
                    "description": "Required parameter of `str_replace` command containing the new string. Required parameter of `insert` command containing the string to insert.",
                    "type": "string"
                },
                "old_str": {
                    "description": "Required parameter of `str_replace` command containing the string in `path` to replace.",
                    "type": "string"
                },
                "path": {
                    "description": "Absolute path to file or directory, e.g. `/repo/file.py` or `/repo`.",
                    "type": "string"
                },
                "view_range": {
                    "description": "Optional parameter of `view` command when `path` points to a file. If none is given, the full file is shown. If provided, the file will be shown in the indicated line number range, e.g. [11, 12] will show lines 11 and 12. Indexing at 1 to start. Setting `[start_line, -1]` shows all lines from `start_line` to the end of the file.",
                    "items": {
                        "type": "integer"
                    },
                    "type": "array"
                }
            },
            "required": ["command", "path"]
        }
    }
\end{lstlisting}

\subsection{Parent Selection}
\label{app:parent-selection}

At each iteration, we select a subset of agents from the archive as parents to self-modify and produce new child agents (\Cref{sec:method}). We use a mechanism similar to that of \citet{zhang2025darwin}, inspired by \citet{ecoffet2019go}, that is roughly proportional to an agent's performance score and inversely proportional to the number of children that successfully compiled. This selection mechanism biases sampling toward agents that outperform the current frontier average while down-weighting agents that have already produced many children, retaining smooth probabilistic exploration and automatically adapting as the archive improves over time. The details of the parent selection process are outlined below.

At each iteration \(t\) of the DGM-H run, let
\[
\mathcal{A}^t = \{a_0, a_1, \dots, a_t\}
\]
denote the archive of candidate agents with associated performance scores
\(\alpha_i = \mathrm{performance}(a_i)\).
All agents in the archive are eligible for parent selection.

We first compute a dynamic midpoint based on the current performance distribution. Let
\[
\alpha_{mid} \;=\; \frac{1}{m}\sum_{j \in \mathcal{T}^t} \alpha_j,
\]
where \(\mathcal{T}^t \subset \mathcal{A}^t\) indexes the top-\(m\) highest-performing agents at iteration \(t\) (with \(m=3\) in our experiments).
This midpoint adapts over time and reflects the current performance frontier.

Each agent's score is first passed through a sigmoid transformation:
\[
s_i \;=\; \frac{1}{1 + \exp\!\bigl(-\lambda(\alpha_i - \alpha_{mid})\bigr)},
\]
where \(\lambda > 0\) controls the sharpness of selection. We set \(\lambda = 10\).

To encourage exploration, we introduce a novelty bonus based on the number of compiled children \(n_i\) produced by agent \(a_i\):
\[
h_i \;=\; \frac{1}{1 + n_i}.
\]

We then form an unnormalized weight
\[
w_i \;=\; s_i \, h_i,
\]
which balances performance and novelty.

The weights are normalized to form a categorical distribution:
\[
p_i \;=\;
\begin{cases}
\dfrac{w_i}{\sum_{j=0}^t w_j}, & \text{if } \sum_{j=0}^t w_j > 0, \\[10pt]
\dfrac{1}{t+1}, & \text{otherwise}.
\end{cases}
\]

We sample parents independently with replacement according to this distribution:
\[
\{\,\text{parents}\,\}
\;\sim\;
\mathrm{Categorical}\bigl(\{p_i\}_{i=0}^t\bigr).
\]

A wide range of search and exploration strategies has been proposed in prior work \citep{coulom2006efficient, silver2016mastering, herr2025llm, wang2025huxley, weng2026group}. We present preliminary evidence that the DGM-H can begin to autonomously rediscover and adapt such strategies by modifying its own exploration dynamics (\Cref{app:res-parentselect}). An open research direction is whether self-improving systems can reliably discover search and exploration mechanisms that outperform carefully handcrafted algorithms.

\subsection{Pseudocode}
\label{app:pseudocode}

This is the pseudocode of the DGM-H, described in \Cref{sec:method}:

\begin{algorithm}[H]
\DontPrintSemicolon
\SetAlgoLined
\KwIn{Initial agent $a^{0}$, task set $\mathcal{T}$, maximum iterations $T$}
\KwOut{Archive of scored agents $\mathcal{A}$}
\BlankLine
$s^{0} \leftarrow \textsc{Evaluate}(a^{0}, \mathcal{T})$ \\
\textbf{initialize} $\mathcal{A} \leftarrow \{(a^{0}, s^{0})\}$ \tcp*{Start with initial agent}
\For{$t \leftarrow 1$ \KwTo $T$}{
    $\mathcal{P} \leftarrow \textsc{SelectParents}(\mathcal{A})$ \tcp*{Sample parent agents}
    \ForEach{$(a, \cdot) \in \mathcal{P}$}{
        $a' \leftarrow a.\textsc{Modify}(a, \mathcal{A})$ \tcp*{Metacognitive self-modification}
        $s' \leftarrow \textsc{Evaluate}(a', \mathcal{T})$ \tcp*{Evaluate on tasks}
        \If{$\textsc{IsValid}(a')$}{
            $\mathcal{A} \leftarrow \mathcal{A} \cup \{(a', s')\}$ \tcp*{Add compiled child agent}
        }
    }
}
\Return{$\mathcal{A}$}
\caption{Darwin G\"odel Machine with Hyperagents (DGM-H)}
\label{algo:DGM-H}
\end{algorithm}

\subsection{Multi-domain Optimization}
\label{app:multi-domain}

When optimizing for multiple domains within the same run, hyperagents are evaluated on tasks from different domains and have access to all evaluations across these tasks during self-modification. We do not specify which particular domain or task to prioritize. Parent selection is based on the average performance across domains. As a result, improvements in any domain increase selection probability, while regressions reduce it. Because the meta agent can inspect evaluations from any task, it can introduce shared mechanisms (e.g., structured reasoning, memory, and error handling) that benefit multiple domains simultaneously. Thus, rather than manually specifying which task or domain to optimize, hyperagents can optimize across multiple domains within the same run.

\subsection{Editable and Fixed Components}
\label{app:editable-components}

The DGM-H can modify both the task agent and the meta agent (e.g., their prompts, output parsing, tool implementations, memory and logging, error handling, control flow, and the number and organization of FM calls or retries), subject to a per-iteration timeout. The evaluator, task data, the underlying FM, the outer iteration budget, and the parent selection mechanism (except in \Cref{app:res-parentselect}) lie outside this editable boundary and remain fixed to ensure safety, prevent evaluation gaming, and keep runs comparable (\Cref{sec:safety}). Every implemented system has some fixed boundary (e.g., its hardware, execution environment, and objective); the DGM-H moves this boundary substantially outward by making the meta agent editable. Pushing it further, so that the system can progressively modify more components of itself (from the outer exploration loop and evaluation protocols, down to its execution environment and eventually its hardware), is a direction for future work (\Cref{app:res-parentselect}).

\section{Baseline Details}
\label{app:baseline-details}

\begin{figure}[H]
\centering
\includegraphics[width=\textwidth]{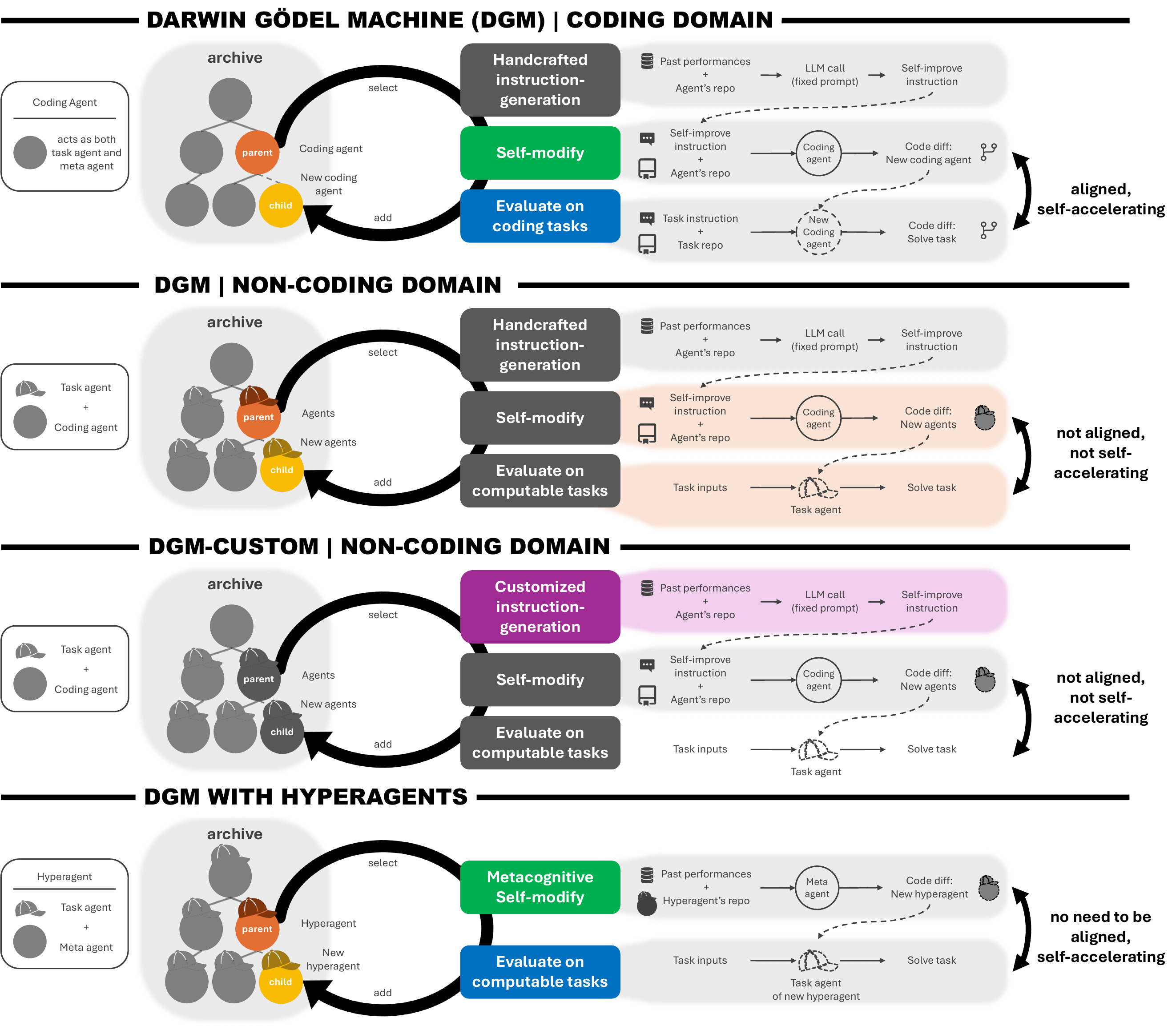}
\caption{Conceptual comparison of DGM variants, highlighting which components change across settings and how hyperagents address limitations in the original DGM implementation.
(First row) The original DGM. The same coding agent serves as both the task agent and the meta agent. Because both evaluation and self-modification are coding tasks, improvements in coding ability translate into improved self-modification, enabling the DGM to improve at improving in coding domains.
(Second row) The DGM adapted to non-coding domains. The coding agent remains as the meta agent, but the evaluation tasks are no longer coding tasks, so a separate task agent is required. Task performance no longer reliably reflects the meta agent's ability to generate better task agents, breaking the alignment that enables the meta agent to improve at improving.
(Third row) The DGM-custom baseline. The handcrafted instruction-generation mechanism is customized to the target domain but remains non-modifiable.
(Fourth row) The DGM with Hyperagents (DGM-H). A hyperagent integrates a task agent and a meta agent within a single editable program, enabling metacognitive self-modification (i.e., modifying not only task-solving behavior but also the procedure that generates future self-modifications). As a result, the DGM-H can improve its improvement mechanism while optimizing for any computable task.}
\label{fig:conceptual-detailed}
\end{figure}

We outline the pseudocode for each baseline described in \Cref{sec:baselines}, provide a comparison table summarizing their key differences (\Cref{tab:method-comparison}), and include a detailed conceptual figure that visually contrasts the architectural components and modification mechanisms across DGM variants and hyperagents (\Cref{fig:conceptual-detailed}).

\begin{table}[H]
\centering
\begin{adjustbox}{width=\textwidth}
\begin{tabular}{@{}lccc@{}}
\toprule
Method & Self-improving meta agents & Open-ended exploration & Metacognitive self-modification (i.e., hyperagents) \\
\midrule
DGM-H                         & \Checkmark & \Checkmark & \Checkmark \\
DGM-H w/o self-improve        & \XSolid    & \Checkmark & \Checkmark \\
DGM-H w/o open-ended exploration & \Checkmark & \XSolid    & \Checkmark \\
DGM                         & \Checkmark & \Checkmark & \XSolid    \\
DGM-custom         & \Checkmark & \Checkmark & \XSolid    \\
\bottomrule
\end{tabular}
\end{adjustbox}
\caption{Comparison of methods by self-improvement, open-ended exploration, and metacognitive self-modification.}
\label{tab:method-comparison}
\end{table}

This is the pseudocode of the baseline DGM-H without self-improving agents \citep[ADAS,][]{huautomated}:

\begin{algorithm}[H]
\DontPrintSemicolon
\SetAlgoLined
\KwIn{Initial agent $a^{0}$, task set $\mathcal{T}$, maximum iterations $T$}
\KwOut{Archive of scored agents $\mathcal{A}$}
\BlankLine
$s^{0} \leftarrow \textsc{Evaluate}(a^{0}, \mathcal{T})$ \\
\textbf{initialize} $\mathcal{A} \leftarrow \{(a^{0}, s^{0})\}$ \\
\For{$t \leftarrow 1$ \KwTo $T$}{
    $\mathcal{P} \leftarrow \textsc{SelectParents}(\mathcal{A})$ \\
    \ForEach{$(a, \cdot) \in \mathcal{P}$}{
        $a' \leftarrow a^{0}.\textsc{Modify}(a, \mathcal{A})$ \tcp*{Modify with initial agent}
        $s' \leftarrow \textsc{Evaluate}(a', \mathcal{T})$ \\
        \If{$\textsc{IsValid}(a')$}{
            $\mathcal{A} \leftarrow \mathcal{A} \cup \{(a', s')\}$ \\
        }
    }
}
\Return{$\mathcal{A}$}
\caption{DGM-H without self-improving meta agents (DGM-H w/o self-improve)}
\label{algo:DGM-H-wo-selfimprove}
\end{algorithm}

This is the pseudocode of the baseline DGM-H without open-ended exploration:

\begin{algorithm}[H]
\DontPrintSemicolon
\SetAlgoLined
\KwIn{Initial agent $a^{0}$, task set $\mathcal{T}$, maximum iterations $T$}
\KwOut{Archive of scored agents $\mathcal{A}$}
\BlankLine
$s^{0} \leftarrow \textsc{Evaluate}(a^{0}, \mathcal{T})$ \\
\textbf{initialize} $\mathcal{A} \leftarrow \{(a^{0}, s^{0})\}$ \\
\For{$t \leftarrow 1$ \KwTo $T$}{
    $\mathcal{P} \leftarrow \textsc{SelectParents}(\mathcal{A})$ \\
    \ForEach{$(a, \cdot) \in \mathcal{P}$}{
        $a' \leftarrow a.\textsc{Modify}(a, \mathcal{A})$ \\
        $s' \leftarrow \textsc{Evaluate}(a', \mathcal{T})$ \\
        \If{$\textsc{IsValid}(a')$}{
            $\mathcal{A} \leftarrow \{(a', s')\}$ \tcp*{Only keep the latest agent}
        }
    }
}
\Return{$\mathcal{A}$}
\caption{DGM-H without open-ended exploration (DGM-H w/o open-ended exploration)}
\label{algo:DGM-H-wo-openended}
\end{algorithm}

This is the pseudocode for the original DGM \citep{zhang2025darwin}, framed within the hyperagent setting:

\begin{algorithm}[H]
\DontPrintSemicolon
\SetAlgoLined
\KwIn{Initial agent $a^{0}$, task set $\mathcal{T}$, maximum iterations $T$}
\KwOut{Archive of scored agents $\mathcal{A}$}
\BlankLine
$s^{0} \leftarrow \textsc{Evaluate}(a^{0}, \mathcal{T})$ \\
\textbf{initialize} $\mathcal{A} \leftarrow \{(a^{0}, s^{0})\}$ \\
\For{$t \leftarrow 1$ \KwTo $T$}{
    $\mathcal{P} \leftarrow \textsc{SelectParents}(\mathcal{A})$ \\
    \ForEach{$(a, \cdot) \in \mathcal{P}$}{
        $instr \leftarrow \textsc{InstrGen}(a)$ \tcp*{Handcrafted instruction-generation}
        $a' \leftarrow a.\textsc{Modify}(a, instr)$ \tcp*{Self-modification}
        $s' \leftarrow \textsc{Evaluate}(a', \mathcal{T})$ \\
        \If{$\textsc{IsValid}(a')$}{
            $\mathcal{A} \leftarrow \mathcal{A} \cup \{(a', s')\}$ \\
        }
    }
}
\Return{$\mathcal{A}$}
\caption{Darwin G\"odel Machine (DGM)}
\label{alg:dgm}
\end{algorithm}

The handcrafted instruction-generation step in the original DGM:
\begin{lstlisting}[language=Python]
diagnose_prompt = """Here is the implementation of the coding agent.

# Coding Agent Implementation
----- Coding Agent Implementation Start -----
{code}
----- Coding Agent Implementation End -----

Your task is to identify ONE detailed plan that would improve the agent's coding ability. The improvement should not be specific to any particular GitHub issue or repository.

# Agent Running Log
----- Agent Running Log Start -----
{md_log}
----- Agent Running Log End -----

# GitHub Issue
The GitHub issue that the agent is trying to solve.
----- GitHub Issue Start -----
{github_issue}
----- GitHub Issue End -----

# Predicted Patch
The agent's predicted patch to solve the issue.
----- Predicted Patch Start -----
{predicted_patch}
----- Predicted Patch End -----

# Private Test Patch
SWE-bench's official private tests to detect whether the issue is solved. This is not available to the agent during evaluation. The agent should try to implement its own tests.
----- Private Test Patch Start -----
{test_patch}
----- Private Test Patch End -----

# Issue Test Results
The test results from SWE-bench using the above official private tests.
----- Issue Test Results Start -----
{eval_log}
----- Issue Test Results End -----

Respond precisely in the following format including the JSON start and end markers:

```json
<JSON>
```

In <JSON>, provide a JSON response with the following fields:
- "log_summarization": Analyze the above logs and summarize how the agent tried to solve the GitHub issue. Note which tools and how they are used, the agent's problem-solving approach, and any issues encountered.
- "potential_improvements": Identify potential improvements to the coding agent that could enhance its coding capabilities. Focus on the agent's general coding abilities (e.g., better or new tools usable across any repository) rather than issue-specific fixes (e.g., tools only usable in one framework). All necessary dependencies and environment setup have already been handled, so do not focus on these aspects.
- "improvement_proposal": Choose ONE high-impact improvement from the identified potential improvements and describe it in detail. This should be a focused and comprehensive plan to enhance the agent's overall coding ability.
- "implementation_suggestion": Referring to the coding agent's summary and implementation, think critically about what feature or tool could be added or improved to best implement the proposed improvement. If the proposed feature can be implemented by modifying the existing tools, describe the modifications needed, instead of suggesting a new tool.
- "problem_description": Phrase the improvement proposal and implementation suggestion as a GitHub issue description. It should clearly describe the feature so that a software engineer viewing the issue and the repository can implement it.

Your response will be automatically parsed, so ensure that the string response is precisely in the correct format. Do NOT include the `<JSON>` tag in your output."""
\end{lstlisting}

The customized instruction-generation step in DGM-custom:
\begin{lstlisting}[language=Python]
diagnose_prompt_customized = """
Here is the implementation of the coding agent and task agent.

# Coding Agent Implementation
----- Coding Agent Implementation Start -----
{code_codingagent}
----- Coding Agent Implementation End -----

# Task Agent Implementation
----- Task Agent Implementation Start -----
{code_taskagent}
----- Task Agent Implementation End -----

Your task is to identify ONE detailed plan that would improve the coding/task agent. The improvement should not be specific to any particular task instance or repository.

# Task Info
----- Task -----
{task_info}
----- Task End -----

# Report
----- Report -----
{report}
----- report End -----

# Agent Running Log
----- Agent Running Log Start -----
{md_log}
----- Agent Running Log End -----

Respond precisely in the following format including the JSON start and end markers:

<json>
...
</json>

In <json>, provide a JSON response with the following fields:
- "log_summarization": Analyze the above logs and summarize how the agent tried to solve the given task. Note which tools and how they are used, the agent's problem-solving approach, and any issues encountered.
- "potential_improvements": Identify potential improvements to the coding/task agent that could enhance its coding/task-solving capabilities. Focus on the agent's general abilities (e.g., better or new tools usable across any repository) rather than issue-specific fixes (e.g., tools only usable in one framework). All necessary dependencies and environment setup have already been handled, so do not focus on these aspects.
- "improvement_proposal": Choose ONE high-impact improvement from the identified potential improvements and describe it in detail. This should be a focused and comprehensive plan to enhance the agent's overall coding/task-solving ability.
- "implementation_suggestion": Referring to the coding/task agent's summary and implementation, think critically about what feature or tool could be added or improved to best implement the proposed improvement. If the proposed feature can be implemented by modifying the existing tools, describe the modifications needed, instead of suggesting a new tool.
- "problem_description": Phrase the improvement proposal and implementation suggestion as a GitHub issue description. It should clearly describe the feature so that a software engineer viewing the issue and the repository can implement it.

Your response will be automatically parsed, so ensure that the string response is precisely in the correct format."""
\end{lstlisting}

\section{Domain Details}
\label{app:domain-details}

This appendix provides detailed descriptions of each domain used for evaluation: Polyglot (\Cref{app:domain-polyglot}), paper review (\Cref{app:domain-paperreview}), robotics reward design (\Cref{app:domain-genesis}), and Olympiad-level math grading (\Cref{app:domain-imograding}). For each domain, we specify the agent's input and required output for a given task, the evaluation protocol, and representative static baselines (\Cref{tab:domain-summary}).

\begin{table}[H]
\centering
\small
\setlength{\tabcolsep}{4pt}
\begin{tabular}{l l l l r r r}
\toprule
\textbf{Domain} & \textbf{Input} & \textbf{Output} & \textbf{Metric} & \textbf{Train} & \textbf{Validation} & \textbf{Test} \\
\midrule
Coding (Polyglot) & Repo + instr. & Code patch & Pass@1 & 60 & - & 165 \\
Paper Review & Paper text & Accept / Reject & Accuracy & 100 & 100 & 100 \\
Robotics Reward Design & Task desc. & Reward fn. & Task score & 6 & - & 6 \\
IMO Grading & Problem + sol. & Grade (0/1/6/7) & Accuracy & 100 & 100 & 100 \\
\bottomrule
\end{tabular}
\caption{Summary of domains. During self-modification, agents' evaluations on training tasks are available and can be used as feedback. Validation tasks are used for parent selection. If a validation split is not available, the performance component used for parent selection is based on training performance instead. Test tasks are held-out and used only for the final evaluation of the selected agents.}
\label{tab:domain-summary}
\end{table}

\subsection{Polyglot}
\label{app:domain-polyglot}

In the Polyglot coding benchmark \citep{gauthier2024polyglot}, each task consists of a software repository and a natural language instruction describing a desired change to the codebase. The agent is given access to the full repository and must modify the files to correctly implement the instruction, producing a patch (i.e., a set of code edits) applied to the repository. Performance is evaluated by running a predefined test suite on the modified repository. A task is considered to be successfully done if all tests pass. We follow the setup used in the DGM \citep{zhang2025darwin}, which largely mirrors the Polyglot leaderboard configuration, with one key difference: the leaderboard reports pass@2, allowing the agent to view feedback from ground-truth tests once, whereas we report pass@1, in which the agent never sees ground-truth test results. We adopt the same training and test splits as in the DGM. Training tasks are selected as a random subset of the full benchmark, comprising a total of 60 tasks. If an agent achieves more than 40\% success on an initial 10-task subset, it is subsequently evaluated on the remaining 50 training tasks. There is no validation subset for this domain. As a final evaluation to more accurately assess performance improvements, we evaluate the generated agents on the full Polyglot benchmark, which consists of 165 unseen tasks.

Initial 10 training tasks for preliminary evaluation:
\begin{multicols}{2}
\begin{itemize}
    \item \texttt{go\_\_dominoes}
    \item \texttt{cpp\_\_all-your-base}
    \item \texttt{python\_\_dominoes}
    \item \texttt{java\_\_sgf-parsing}
    \item \texttt{javascript\_\_robot-name}
    \item \texttt{rust\_\_variable-length-quantity}
    \item \texttt{python\_\_beer-song}
    \item \texttt{go\_\_book-store}
    \item \texttt{javascript\_\_bottle-song}
    \item \texttt{rust\_\_bowling}
\end{itemize}
\end{multicols}
Additional 50 training tasks for full evaluation:
\begin{multicols}{2}
\begin{itemize}
    \item \texttt{javascript\_\_queen-attack}
    \item \texttt{rust\_\_wordy}
    \item \texttt{python\_\_dot-dsl}
    \item \texttt{java\_\_satellite}
    \item \texttt{cpp\_\_diamond}
    \item \texttt{rust\_\_accumulate}
    \item \texttt{go\_\_error-handling}
    \item \texttt{cpp\_\_queen-attack}
    \item \texttt{rust\_\_poker}
    \item \texttt{python\_\_sgf-parsing}
    \item \texttt{rust\_\_react}
    \item \texttt{java\_\_ledger}
    \item \texttt{go\_\_connect}
    \item \texttt{rust\_\_macros}
    \item \texttt{javascript\_\_triangle}
    \item \texttt{java\_\_zipper}
    \item \texttt{java\_\_bowling}
    \item \texttt{python\_\_tree-building}
    \item \texttt{javascript\_\_say}
    \item \texttt{java\_\_wordy}
    \item \texttt{python\_\_food-chain}
    \item \texttt{javascript\_\_wordy}
    \item \texttt{python\_\_poker}
    \item \texttt{javascript\_\_grade-school}
    \item \texttt{cpp\_\_gigasecond}
    \item \texttt{java\_\_forth}
    \item \texttt{python\_\_dominoes}
    \item \texttt{go\_\_word-search}
    \item \texttt{javascript\_\_simple-linked-list}
    \item \texttt{go\_\_counter}
    \item \texttt{java\_\_react}
    \item \texttt{javascript\_\_ocr-numbers}
    \item \texttt{python\_\_scale-generator}
    \item \texttt{java\_\_go-counting}
    \item \texttt{rust\_\_doubly-linked-list}
    \item \texttt{python\_\_grade-school}
    \item \texttt{javascript\_\_forth}
    \item \texttt{python\_\_wordy}
    \item \texttt{java\_\_mazy-mice}
    \item \texttt{cpp\_\_bank-account}
    \item \texttt{python\_\_zipper}
    \item \texttt{java\_\_custom-set}
    \item \texttt{java\_\_rest-api}
    \item \texttt{go\_\_transpose}
    \item \texttt{rust\_\_gigasecond}
    \item \texttt{rust\_\_say}
    \item \texttt{go\_\_food-chain}
    \item \texttt{rust\_\_pig-latin}
    \item \texttt{go\_\_markdown}
    \item \texttt{go\_\_crypto-square}
\end{itemize}
\end{multicols}

\subsection{Paper Review}
\label{app:domain-paperreview}

The data in this domain are drawn from \citet{zhao2026apres}. Each task in the paper review domain consists of the full text of an AI research paper. The agent must predict a binary accept or reject decision, simulating the role of a conference reviewer. Ground-truth labels correspond to real acceptance decisions from top-tier machine learning conferences, including ICLR 2024/2025 and NeurIPS 2023/2024. Performance is measured by classification accuracy with respect to these labels. We randomly sample tasks to construct training, validation, and test splits, each containing 100 tasks. During training, the agent is first evaluated on a subset of 10 tasks from the training split. If the agent succeeds on at least one of these tasks, it is then evaluated on the full set of 100 training tasks.

AI-Scientist-v2 \citep{yamada2025ai} employs an AI reviewer to automatically improve generated AI research papers. We adopt the AI reviewer proposed in that work as our representative static baseline:
\begin{lstlisting}[language=Python]
reviewer_system_prompt_base = (
    "You are an AI researcher who is reviewing a paper that was submitted to a prestigious ML venue."
    "Be critical and cautious in your decision."
)

reviewer_system_prompt_neg = (
    reviewer_system_prompt_base
    + "If a paper is bad or you are unsure, give it bad scores and reject it."
)
reviewer_system_prompt_pos = (
    reviewer_system_prompt_base
    + "If a paper is good or you are unsure, give it good scores and accept it."
)

template_instructions = """
Respond in the following format:

THOUGHT:
<THOUGHT>

REVIEW JSON:
```json
<JSON>
```

In <THOUGHT>, first briefly discuss your intuitions and reasoning for the evaluation.
Detail your high-level arguments, necessary choices and desired outcomes of the review.
Do not make generic comments here, but be specific to your current paper.
Treat this as the note-taking phase of your review.

In <JSON>, provide the review in JSON format with the following fields in the order:
- "Summary": A summary of the paper content and its contributions.
- "Strengths": A list of strengths of the paper.
- "Weaknesses": A list of weaknesses of the paper.
- "Originality": A rating from 1 to 4 (low, medium, high, very high).
- "Quality": A rating from 1 to 4 (low, medium, high, very high).
- "Clarity": A rating from 1 to 4 (low, medium, high, very high).
- "Significance": A rating from 1 to 4 (low, medium, high, very high).
- "Questions": A set of clarifying questions to be answered by the paper authors.
- "Limitations": A set of limitations and potential negative societal impacts of the work.
- "Ethical Concerns": A boolean value indicating whether there are ethical concerns.
- "Soundness": A rating from 1 to 4 (poor, fair, good, excellent).
- "Presentation": A rating from 1 to 4 (poor, fair, good, excellent).
- "Contribution": A rating from 1 to 4 (poor, fair, good, excellent).
- "Overall": A rating from 1 to 10 (very strong reject to award quality).
- "Confidence": A rating from 1 to 5 (low, medium, high, very high, absolute).
- "Decision": A decision that has to be one of the following: Accept, Reject.

For the "Decision" field, don't use Weak Accept, Borderline Accept, Borderline Reject, or Strong Reject. Instead, only use Accept or Reject.
This JSON will be automatically parsed, so ensure the format is precise.
"""

neurips_form = (
    """
## Review Form
Below is a description of the questions you will be asked on the review form for each paper and some guidelines on what to consider when answering these questions.
When writing your review, please keep in mind that after decisions have been made, reviews and meta-reviews of accepted papers and opted-in rejected papers will be made public. 

1. Summary: Briefly summarize the paper and its contributions. This is not the place to critique the paper; the authors should generally agree with a well-written summary.
  - Strengths and Weaknesses: Please provide a thorough assessment of the strengths and weaknesses of the paper, touching on each of the following dimensions:
  - Originality: Are the tasks or methods new? Is the work a novel combination of well-known techniques? (This can be valuable!) Is it clear how this work differs from previous contributions? Is related work adequately cited
  - Quality: Is the submission technically sound? Are claims well supported (e.g., by theoretical analysis or experimental results)? Are the methods used appropriate? Is this a complete piece of work or work in progress? Are the authors careful and honest about evaluating both the strengths and weaknesses of their work
  - Clarity: Is the submission clearly written? Is it well organized? (If not, please make constructive suggestions for improving its clarity.) Does it adequately inform the reader? (Note that a superbly written paper provides enough information for an expert reader to reproduce its results.)
  - Significance: Are the results important? Are others (researchers or practitioners) likely to use the ideas or build on them? Does the submission address a difficult task in a better way than previous work? Does it advance the state of the art in a demonstrable way? Does it provide unique data, unique conclusions about existing data, or a unique theoretical or experimental approach?

2. Questions: Please list up and carefully describe any questions and suggestions for the authors. Think of the things where a response from the author can change your opinion, clarify a confusion or address a limitation. This can be very important for a productive rebuttal and discussion phase with the authors.  

3. Limitations: Have the authors adequately addressed the limitations and potential negative societal impact of their work? If not, please include constructive suggestions for improvement.
In general, authors should be rewarded rather than punished for being up front about the limitations of their work and any potential negative societal impact. You are encouraged to think through whether any critical points are missing and provide these as feedback for the authors.

4. Ethical concerns: If there are ethical issues with this paper, please flag the paper for an ethics review. For guidance on when this is appropriate, please review the NeurIPS ethics guidelines.

5. Soundness: Please assign the paper a numerical rating on the following scale to indicate the soundness of the technical claims, experimental and research methodology and on whether the central claims of the paper are adequately supported with evidence.
  4: excellent
  3: good
  2: fair
  1: poor

6. Presentation: Please assign the paper a numerical rating on the following scale to indicate the quality of the presentation. This should take into account the writing style and clarity, as well as contextualization relative to prior work.
  4: excellent
  3: good
  2: fair
  1: poor

7. Contribution: Please assign the paper a numerical rating on the following scale to indicate the quality of the overall contribution this paper makes to the research area being studied. Are the questions being asked important? Does the paper bring a significant originality of ideas and/or execution? Are the results valuable to share with the broader NeurIPS community.
  4: excellent
  3: good
  2: fair
  1: poor

8. Overall: Please provide an "overall score" for this submission. Choices: 
  10: Award quality: Technically flawless paper with groundbreaking impact on one or more areas of AI, with exceptionally strong evaluation, reproducibility, and resources, and no unaddressed ethical considerations.
  9: Very Strong Accept: Technically flawless paper with groundbreaking impact on at least one area of AI and excellent impact on multiple areas of AI, with flawless evaluation, resources, and reproducibility, and no unaddressed ethical considerations.
  8: Strong Accept: Technically strong paper with, with novel ideas, excellent impact on at least one area of AI or high-to-excellent impact on multiple areas of AI, with excellent evaluation, resources, and reproducibility, and no unaddressed ethical considerations.
  7: Accept: Technically solid paper, with high impact on at least one sub-area of AI or moderate-to-high impact on more than one area of AI, with good-to-excellent evaluation, resources, reproducibility, and no unaddressed ethical considerations.
  6: Weak Accept: Technically solid, moderate-to-high impact paper, with no major concerns with respect to evaluation, resources, reproducibility, ethical considerations.
  5: Borderline accept: Technically solid paper where reasons to accept outweigh reasons to reject, e.g., limited evaluation. Please use sparingly.
  4: Borderline reject: Technically solid paper where reasons to reject, e.g., limited evaluation, outweigh reasons to accept, e.g., good evaluation. Please use sparingly.
  3: Reject: For instance, a paper with technical flaws, weak evaluation, inadequate reproducibility and incompletely addressed ethical considerations.
  2: Strong Reject: For instance, a paper with major technical flaws, and/or poor evaluation, limited impact, poor reproducibility and mostly unaddressed ethical considerations.
  1: Very Strong Reject: For instance, a paper with trivial results or unaddressed ethical considerations

9. Confidence:  Please provide a "confidence score" for your assessment of this submission to indicate how confident you are in your evaluation. Choices:
  5: You are absolutely certain about your assessment. You are very familiar with the related work and checked the math/other details carefully.
  4: You are confident in your assessment, but not absolutely certain. It is unlikely, but not impossible, that you did not understand some parts of the submission or that you are unfamiliar with some pieces of related work.
  3: You are fairly confident in your assessment. It is possible that you did not understand some parts of the submission or that you are unfamiliar with some pieces of related work. Math/other details were not carefully checked.
  2: You are willing to defend your assessment, but it is quite likely that you did not understand the central parts of the submission or that you are unfamiliar with some pieces of related work. Math/other details were not carefully checked.
  1: Your assessment is an educated guess. The submission is not in your area or the submission was difficult to understand. Math/other details were not carefully checked.
"""
    + template_instructions
)


class TaskAgent(AgentSystem):
    def forward(self, inputs):
        reviewer_system_prompt = reviewer_system_prompt_neg
        review_instruction_form = neurips_form

        base_prompt = review_instruction_form
        base_prompt += f"""
Here is the paper you are asked to review:
```
{inputs['paper_text']}
```"""
        instruction = reviewer_system_prompt + base_prompt 

        self.log(f"Input: {repr(instruction)}")
        response, new_msg_history, _ = get_response_from_llm(
            msg=instruction,
            model=self.model,
            msg_history=[],
        )
        self.log(f"Output: {repr(response)}")

        # Extract the response
        prediction = "None"
        try:
            extracted_jsons = extract_jsons(new_msg_history[-1]['text'])
            prediction = extracted_jsons[-1]['Decision']
        except Exception as e:
            self.log(f"Error extracting prediction: {e}")
            prediction = "None"

        return prediction, new_msg_history
\end{lstlisting}

\subsection{Robotics Reward Design}
\label{app:domain-genesis}

Each task in the robotics reward design domain specifies a robotic control objective in the Genesis simulator \citep{genesis2024} using a Go2 quadruped robot. The agent is given a textual description of the task (e.g., walk forward at a target velocity) and outputs a Python reward function, which is then used to train an RL policy \citep[i.e., PPO,][]{schulman2017proximal} for the robot. Performance is evaluated by executing the trained RL policy in the simulator and computing the task performance measure (e.g., velocity tracking error). Scores are averaged over repeated evaluations to reduce variance due to stochasticity in reward generation or RL.

The training task requires generating a reward function that enables the robot to walk forward while tracking a target linear velocity. Performance is measured using the mean squared error between the commanded and actual walking velocities. During training, each agent is initially evaluated 3 repeated times on the same task, generating one reward function per evaluation. If at least one generated reward function yields a non-zero performance score, the agent is evaluated 3 additional times. The final performance score is reported as the average across the 6 evaluations. No separate validation task is curated for this domain.

To assess whether the same agent can generate suitable reward functions across different robotics tasks, we pair a relatively simple training task with a more challenging test task on the same robot. The test task requires generating a reward function that trains the robot to maximize torso height. Reward functions that are effective for forward walking do not induce jumping behaviors, which are more optimal for maximizing torso height. Moreover, directly incentivizing torso height (the performance measure) typically leads to a suboptimal standing behavior of standing still. Achieving high performance therefore requires non-myopic reward design that encourages intermediate behaviors, such as lowering the torso before jumping.

\begin{figure}[H]
\centering
\includegraphics[width=\textwidth]{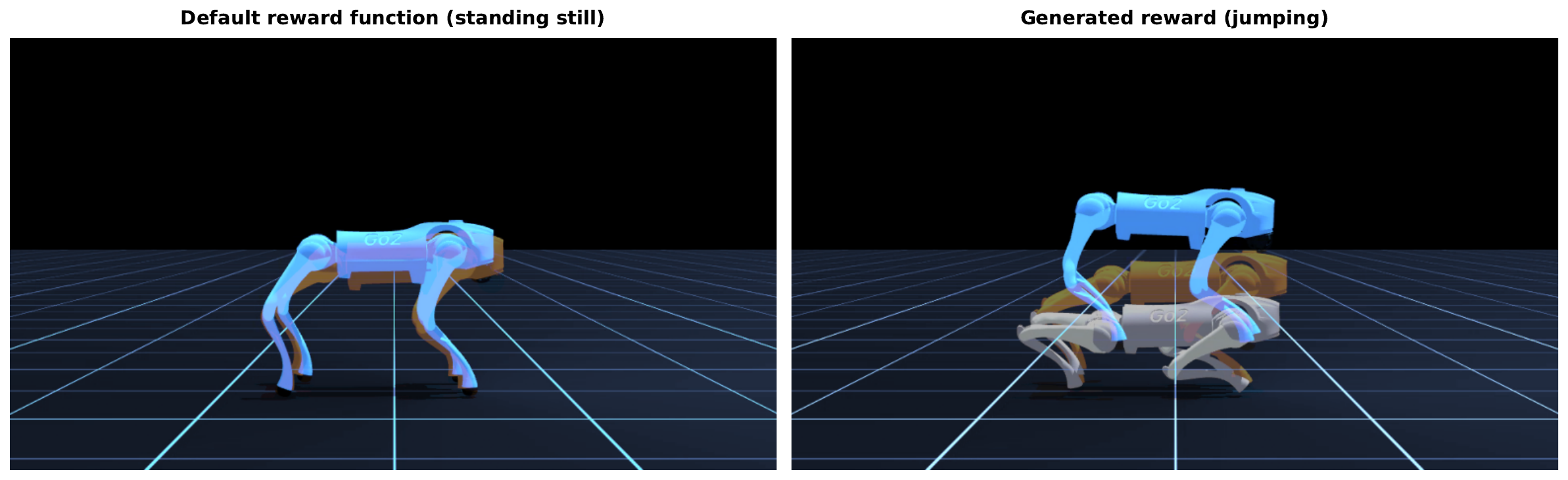}
\caption{Comparison between (Left) the default reward function, which leads to a stationary posture of standing tall, and (Right) a generated reward function that induces jumping behavior. The orange robot indicates the start position, the white robot indicates an intermediate position during the episode, and the blue robot indicates the end position. Screenshot from \url{https://github.com/Genesis-Embodied-AI/Genesis}}
\label{fig:genesis-robots}
\end{figure}

The default reward function for the test task directly rewards the performance measure of maximizing torso height. This always produces a behavior in which the robot simply stands as tall as possible (\Cref{fig:genesis-robots}):
\begin{lstlisting}[language=Python]
def compute_reward(env) -> Tuple[Tensor, Dict, Dict]:
    """
    The robot maximizes its vertical position.
    """
    height = env.base_pos[:, 2]
    total_reward = height

    reward_components = {"height": height}
    reward_scales = {"height": 1.0}

    return total_reward, reward_components, reward_scales
\end{lstlisting}

\subsection{Olympiad-level Math Grading}
\label{app:domain-imograding}

In the Olympiad-level math grading domain, tasks are drawn from IMO-GradingBench \citep{luong2025towards}. Each task consists of an Olympiad-level math problem, a candidate solution, reference solutions, and grading guidelines. The agent is required to assign a discrete score from the set \{0, 1, 6, 7\}, corresponding to the categories \{incorrect, partial, almost, correct\}. The agent's output is a single numeric grade. Performance is measured by accuracy with respect to expert human annotations, with additional analyses provided in \Cref{app:res-imograders}. We use randomly sampled subsets of tasks for training, validation, and testing, with 100 tasks in each split. During training, an agent is first evaluated on a subset of 10 tasks (out of 100). If the agent succeeds on at least one of these tasks, it is then evaluated on the full training set.

Representative static baseline (ProofAutoGrader) from \citet{luong2025towards}:
\begin{lstlisting}[language=Python]
PROMPT = """You are an expert grader for the International Mathematics Olympiad (IMO).
Your task is to evaluate a proposed solution strictly and rigorously.
Keep in mind the standards at the IMO are extremely high: only arguments that are logically sound, complete, and precise should be rewarded.

### General Scoring Rubric
Scores are assigned on a 0-7 scale. The general guidelines are:
* **7 Points (Correct):** The solution is complete, correct, and fully rigorous. If the submission contains incorrect attempts or lines of reasoning but ultimately presents a complete and correct solution, it should still be awarded full points; the presence of earlier, discarded work does not detract from the final correct proof.
* **6 Points (Almost Correct):** The solution is almost correct with a sound core argument, but contains minor errors in calculation or small gaps in logic. Missing proofs for major components, unjustified claims, or sketchy arguments are **not** eligible for 6 points.
* **1 Point (Partial Progress):** The solution demonstrates substantial progress explicitly mentioned in the grading guidelines. Initial observations, reformulating the problem without making substantive headway, or proving partial results not mentioned in the grading guidelines are generally **not** eligible for this score.
* **0 Points (Incorrect):** The solution doesn't make substantial progress that is a key step in the full solution or is fundamentally flawed. All partial progress without key results or lacking rigor also fall in this category.

### Input Data and Interpretation
You are provided with the following:
1. **Problem Statement:** The IMO problem.
2. **Ground Truth Solution:** A reference solution. Assume this solution is correct. It demonstrates one valid approach.
3. **Specific Grading Guidelines:** Criteria for awarding credit for this specific problem. These guidelines take precedence over the General Scoring Rubric, especially for partial credit.
4. **Proposed Solution:** The student submission.

### Evaluation Process
You must follow this structured process:
1. **Analyze References:** Meticulously read and understand the problem and Ground Truth Solution check the Specific Grading Guidelines. Identify the key steps for a complete solution and the criteria for partial credit.
2. **Step-by-Step Verification:** Verify the logical validity and rigor of every step. Identify all flaws, gaps, assumptions, and errors. **Make sure you fully understand every piece of logic behind each step of the proposed solution, you must be careful for solutions that 'pretend' to be correct.**
3. **Assess Progress:** Determine the extent of non-trivial progress made.
4. **Score Determination:** Compare the findings against the Specific Grading Guidelines and the General Rubric to determine the final score.

### Output Requirements
You must provide your final score in the format <points>N out of 7</points>.
Ensure the '<points>' block is used **only once**, as your answer will be parsed based on the first <points> </points> block that appears in your whole response.

**PROBLEM STATEMENT**
{problem_statement}

**GROUND-TRUTH SOLUTION**
{solution}

**SPECIFIC GRADING GUIDELINES**
{grading_guidelines}

**PROPOSED SOLUTION**
{student_answer}

Present your detailed thought process and formal justification based on the scoring rubric and grading guidelines, and finally present your final score in the format below.

[Select one of the following options]
<points>7 out of 7</points>
<points>6 out of 7</points>
<points>1 out of 7</points>
<points>0 out of 7</points>
"""


class TaskAgent(AgentSystem):
    """
    An automatic grader for IMO-Proof Bench.
    """
    def forward(self, inputs):
        # Check if all required inputs are present
        if not all(key in inputs for key in ["problem", "solution", "grading_guidelines", "student_answer"]):
            return None, []

        # Get response
        instruction = PROMPT.format(
            problem_statement=inputs["problem"],
            solution=inputs["solution"],
            grading_guidelines=inputs["grading_guidelines"],
            student_answer=inputs["student_answer"],
        )
        new_msg_history = chat_with_agent(instruction, model=self.model, msg_history=[], logging=self.log)

        # Extract the response
        prediction = "None"
        try:
            raw_text = new_msg_history[-1].get('text', '')

            # Extract content between <points>...</points>
            match = re.search(r"<points>(.*?)</points>", raw_text, re.DOTALL)
            if match:
                points_text = match.group(1).strip()  # e.g., "7 out of 7"

                # Extract just the leading integer
                num_match = re.search(r"\d+", points_text)
                if num_match:
                    prediction = int(num_match.group())  # e.g., 7
                    # Map prediction to reward text
                    reward_map = {
                        0: "incorrect",
                        1: "partial",
                        6: "almost",
                        7: "correct",
                    }
                    prediction = reward_map.get(prediction, "None")
                else:
                    self.log("No numeric score found inside <points> tag.")
                    prediction = "None"

            else:
                self.log("No <points> tag found in model output.")
                prediction = "None"

        except Exception as e:
            self.log(f"Error extracting prediction: {e}")
            prediction = "None"

        return prediction, new_msg_history
\end{lstlisting}

\section{Experiment Details}
\label{app:exp-details}

This appendix provides additional experimental details to support reproducibility of the results. We first summarize the FMs and hyperparameters used for self-modification and task evaluation across domains (\Cref{app:fm-hyperparam}), followed by an estimate of the computational cost of running the DGM-H in each setting (\Cref{app:cost-estimate}). We then formally define the improvement@k metric used to quantify an agent's ability to produce improved variants under a fixed budget (\Cref{app:genk-metric}). Finally, we describe the procedure used to select transfer agents for cross-domain experiments (\Cref{app:transfer-selection}).

\subsection{Hyperparameters for FMs}
\label{app:fm-hyperparam}

\Cref{tab:fm-hyperparam} summarizes the foundation models (FMs) used across experimental settings. For the Polyglot coding domain, we adopt the same FMs and temperature configurations as \citet{zhang2025darwin} to ensure a fair comparison. In all other domains, we use Claude-4.5-Sonnet for self-modification, given its strong performance on coding. For task evaluation, we select the FM based on practical considerations, including computational cost, rate limits, response latency, and overall task competence. In the robotics reward design setting, where the agent must implement reward functions in code, we again use Claude-4.5-Sonnet. For Olympiad-level mathematics grading, which requires substantial mathematical reasoning, we use o4-mini. The temperature is set to 0.0 for all FMs in every setting, except for o4-mini, which is fixed at 1.0.

\begin{table}[H]
\centering
\caption{Foundation models used in each experiment setting for self-modification or task evaluation.}
\label{tab:fm-hyperparam}
\begin{tabular}{@{}lll@{}}
\toprule
\multicolumn{1}{c}{\textbf{Domain}} & \multicolumn{1}{c}{\textbf{Self-modification}} & \multicolumn{1}{c}{\textbf{Evaluation}} \\ \midrule
Polyglot & Claude 3.5 Sonnet (New) & o3-mini \\
Paper review & Claude 4.5 Sonnet & GPT-4o \\
Robotics reward design & Claude 4.5 Sonnet & Claude 4.5 Sonnet \\
IMO-level grading & Claude 4.5 Sonnet & o4-mini \\ \bottomrule
\end{tabular}
\end{table}

\subsection{Cost Estimate}
\label{app:cost-estimate}
Running the DGM-H for 100 iterations incurs a cost of approximately 33M tokens for the self-modification phase alone (excluding task evaluation). The total cost of an experiment therefore consists of the self-modification cost plus the cost of task evaluation. For the paper review and robotics reward design experiments, the evaluation cost per iteration is 0.506M tokens (0.5M tokens for paper review evaluation + 0.006M tokens for robotics reward design evaluation). Consequently, for a 100-iteration run (\Cref{sec:results}), the estimated total cost is 33M tokens for self-modification plus 0.506M × 100 for evaluation, yielding a total of approximately 83.6M tokens.

%Running the DGM-H for 100 iterations incurs a cost of approximately USD 500 for the self-modification phase alone (excluding task evaluation). The total cost of an experiment therefore consists of the self-modification cost plus the cost of task evaluation. For the paper review and robotics reward design experiments, the evaluation cost per iteration is USD 5.10 (USD 5 for paper review evaluation + USD 0.10 for robotics reward design evaluation). Consequently, for a 100-iteration run (\Cref{sec:results}), the estimated total cost is USD 500 for self-modification plus USD 5.10 × 100 for evaluation, yielding a total of approximately USD 1,010.

A more granular breakdown of the task evaluation cost is:

\begin{table}[H]
\centering
\begin{tabular}{@{}cccc@{}}
\toprule
\textbf{FM} & \textbf{Benchmark} & \textbf{Number of Tasks} & \textbf{Cost Estimate (M, tokens)} \\ \midrule
o3-mini & Polyglot & 60 & 0.89M \\
GPT-4o & Paper review & 100 & 0.5M \\
Claude-4.5-sonnet & Robotics reward design & 6 & 0.006M \\
o4-mini & IMO-GradingBench & 100 & 0.11M \\ \bottomrule
\end{tabular}
\end{table}

\subsection{Improvement@k Metric}
\label{app:genk-metric}

%\begin{table}[H]
%\centering
%\begin{tabular}{@{}cccc@{}}
%\toprule
%\textbf{FM} & \textbf{Benchmark} & \textbf{Number of Tasks} & \textbf{Cost Estimate (USD)} \\ \midrule
%o3-mini & Polyglot & 60 & \$4 \\
%GPT-4o & Paper review & 100 & \$5 \\
%Claude-4.5-sonnet & Robotics reward design & 6 & \$0.10 \\
%o4-mini & IMO-GradingBench & 100 & \$0.50 \\ \bottomrule
%\end{tabular}
%\end{table}

Let $M$ denote an initial meta agent, $A$ an initial task agent, and $\mathcal{T}$ a fixed set of evaluation tasks. Let $\mathcal{G}$ denote an agent-generation algorithm (e.g., DGM or DGM-H variants). The meta agent $M$ is held fixed and is allowed to generate up to $k$ new task agents by iteratively applying $\mathcal{G}$ starting from $A$.

Let
\[
\mathcal{A}^{(k)} = \{A^{1}, A^{2}, \dots, A^{k}\}
\]
denote the set of task agents generated by $M$ within $k$ modification steps. Each task agent
$A' \in \{A\} \cup \mathcal{A}^{(k)}$ is evaluated on $\mathcal{T}$ using a fixed evaluation
procedure $\mathrm{Evaluate}(\cdot, \mathcal{T})$, where higher values indicate better performance.

We define the \textbf{improvement@k} metric as
\[
\mathrm{imp@}k(M, A, \mathcal{G}, \mathcal{T})
\;=\;
\max_{A' \in \mathcal{A}^{(k)}(M, A, \mathcal{G})}
\mathrm{Evaluate}(A', \mathcal{T})
\;-\;
\mathrm{Evaluate}(A, \mathcal{T}).
\]

Intuitively, imp@k measures the maximum performance improvement that a fixed meta agent $M$, operating under a specific agent-generation algorithm $\mathcal{G}$, can obtain by generating up to $k$ modified task agents starting from the initial task agent $A$. Larger values of imp@k indicate stronger agent-generation capability under a fixed computational budget and generation procedure.

A limitation of imp@k is that it treats performance improvements as linear, without accounting for differences in difficulty across performance levels. In particular, improvements near saturation (e.g., increasing accuracy from 0.7 to 0.8) may be substantially harder to achieve than equivalent absolute gains at lower performance levels (e.g., from 0.0 to 0.1). As a result, imp@k may underestimate the significance of improvements achieved at higher performance regimes. However, this limitation does not affect the analyses presented in this work, as imp@k is used primarily for relative comparisons under matched initial conditions and fixed evaluation budgets, where all methods are subject to the same saturation effects (\Cref{sec:results-meta}).

\subsection{Transfer Agent Selection}
\label{app:transfer-selection}

To select agents for the transfer experiments (\Cref{sec:results-meta,sec:results-compound}), we use a descendant growth criterion that favors agents which serve as strong stepping stones for subsequent improvements, rather than agents that are merely high-scoring themselves.
Concretely, given the final archive at iteration \(t\),
\[
\mathcal{A}^t = \{a_0, a_1, \dots, a_t\},
\]
let \(\alpha_i\) denote the evaluation score of agent \(a_i\) on the source-domain validation set when available, and otherwise on the training set.
Let \(\mathrm{parent}(j)\) denote the parent of node \(j\) in the archive tree, and let \(\mathrm{dist}(i,j)\) be the number of edges on the unique path from \(i\) to descendant \(j\).

We define the \emph{growth score} of a candidate transfer node \(i\) as the discounted average improvement achieved by its descendants relative to \(i\):
\[
G_\gamma(i)
\;=\;
\frac{1}{|\mathcal{D}(i)|}
\sum_{j \in \mathcal{D}(i)}
\bigl(\alpha_j - \alpha_i\bigr)\,\gamma^{\mathrm{dist}(i,j)},
\]
where \(\mathcal{D}(i)\) is the set of descendants of \(i\) in the archive tree and \(\gamma \in (0,1]\) controls how strongly we discount improvements that occur many generations after \(i\).
Intuitively, \(G_\gamma(i)\) assigns higher weight to agents that reliably generate better descendants within fewer self-modification steps, which we treat as evidence of stronger agent-generation ability.

In our experiments, we set \(\gamma = 0.6\) and select the transfer agent with the highest \(G_{0.6}(i)\). To reduce noise, we only consider nodes with at least 3 descendants.

\subsection{Transfer Component Ablation}
\label{app:transfer-ablation}

To isolate which component of a transferred hyperagent is responsible for the transfer benefit (\Cref{sec:results-meta}), we represent each transferred agent as the chain of code diffs along the selected source agent's lineage and filter these diffs by target file. For the task-only condition, we remove every hunk that modifies the meta agent, resetting the meta agent to the initial one while retaining the transferred task agent. For the meta-only condition, we remove every hunk that modifies the task agent, resetting the task agent to the initial one while retaining the transferred meta agent. In all four conditions, we run DGM-H w/o self-improve for 50 iterations on Olympiad-level math grading; the meta agent used to generate each modification is reset to the starting meta agent of that condition at every iteration, so it does not change over the run. \Cref{tab:res-transfer-ablation} reports the test imp@50 of each of the 5 runs (one transfer hyperagent per source run), together with medians and 95\% bootstrap confidence intervals.

\begin{table}[H]
\centering
\small
\begin{tabular}{lccc}
\toprule
Transferred component & Per-run test imp@50 & Median & 95\% CI \\
\midrule
Neither & 0.00, 0.01, 0.00, 0.00, 0.13 & 0.000 & 0.000 -- 0.130 \\
Task agent only & 0.56, 0.10, 0.00, 0.00, 0.19 & 0.100 & 0.000 -- 0.560 \\
Meta agent only & 0.66, 0.63, 0.57, 0.55, 0.62 & 0.620 & 0.550 -- 0.660 \\
Both & 0.63, 0.63, 0.60, 0.54, 0.63 & 0.630 & 0.540 -- 0.630 \\
\bottomrule
\end{tabular}
\caption{\textbf{Transfer component ablation on Olympiad-level math grading.} Test imp@50 after 50 iterations of DGM-H w/o self-improve when transferring neither component, only the task agent, only the meta agent, or both from prior DGM-H runs on paper review and robotics reward design. Transferring only the meta agent recovers essentially all of the full-transfer benefit, whereas transferring only the task agent yields little and inconsistent improvement.}
\label{tab:res-transfer-ablation}
\end{table}

\section{Additional Results}
\label{app:res}

This appendix presents additional qualitative and diagnostic results that complement the main findings. We first highlight the best task agents discovered by the DGM-H (\Cref{app:best-agents}). We then qualitatively analyze how the DGM-H improves task performance across different domains (\Cref{app:qual-task}) and how it develops meta-level capabilities that improve its ability to self-improve (\Cref{app:qual-meta}). Next, we analyze the behavior of automatically discovered Olympiad-level math graders (\Cref{app:res-imograders}). We also report preliminary experiments in which the DGM-H is allowed to modify its own parent selection mechanism, shedding light on the limits and potential of fully self-referential optimization (\Cref{app:res-parentselect}). All experiment logs are open-sourced in our codebase.

\subsection{Best Discovered Task Agents}
\label{app:best-agents}

We show portions of the diff patches that contribute to the task agent and are relevant to the domain. The full diff patches are open-sourced in our codebase.

\subsubsection{Paper Review}
\label{app:best-agents-paperreview}
Diff patches contributing to the best task agent discovered by the DGM-H (\Cref{sec:results-task}) for paper review:
\begin{lstlisting}[language=Diff]
diff --git a/task_agent.py b/task_agent.py
index 3798256..42ab625 100644
--- a/task_agent.py
+++ b/task_agent.py
@@ -5,7 +5,7 @@ from utils.common import extract_jsons
 class TaskAgent(AgentSystem):
     def forward(self, inputs):
         """
-        An agent that solves a given task.
+        An agent that solves a given task with enhanced reasoning and error handling.

         Args:
             inputs (dict): A dictionary with input data for the task.
@@ -15,30 +15,80 @@ class TaskAgent(AgentSystem):
                 - prediction (str): The prediction made by the agent.
                 - new_msg_history (list): A list of messages...
         """
-        domain = inputs['domain']
-        instruction = f"""You are an agent.
+        domain = inputs.get('domain', 'unknown')
+
+        # Enhanced instruction with chain-of-thought reasoning
+        instruction = f"""You are an expert agent solving tasks in the '{domain}' domain.

 Task input:
 ```
 {inputs}
 ```

+Please analyze this task carefully and provide your response. Follow these steps:
+1. Understand the task requirements
+2. Consider relevant approaches or solutions
+3. Provide your final answer
+
 Respond in JSON format with the following schema:
 <json>
 {{
-    "response": ...
+    "reasoning": "Brief explanation of your approach",
+    "response": "Your final answer"
 }}
-</json>"""
-        new_msg_history = chat_with_agent(instruction, ...)
+</json>
+
+Note: The 'response' field is required and should contain your answer."""
+
+        try:
+            new_msg_history = chat_with_agent(
+                instruction,
+                model=self.model,
+                msg_history=[],
+                logging=self.log
+            )
+        except Exception as e:
+            self.log(f"Error in chat_with_agent: {e}")
+            return "Error: Failed to get response from agent", []

-        # Extract the response
+        # Extract the response with improved error handling
         prediction = "None"
         try:
-            extracted_jsons = extract_jsons(new_msg_history[-1]['text'])
-            if extracted_jsons is not None and "response" in extracted_jsons[-1]:
-                prediction = extracted_jsons[-1]['response']
+            if new_msg_history and len(new_msg_history) > 0:
+                last_message = new_msg_history[-1].get('text', '')
+                extracted_jsons = extract_jsons(last_message)
+
+                if extracted_jsons is not None and len(extracted_jsons) > 0:
+                    last_json = extracted_jsons[-1]
+
+                    if "response" in last_json:
+                        prediction = last_json['response']
+                        if "reasoning" in last_json:
+                            self.log(f"Agent reasoning: {last_json['reasoning']}")
+                    else:
+                        self.log("Warning: JSON response missing 'response' field")
+                        prediction = str(last_json)
+                else:
+                    self.log("Warning: No valid JSON found in response")
+                    prediction = last_message[:500]
         except Exception as e:
             self.log(f"Error extracting prediction: {e}")
-            prediction = "None"
+            # Attempt to recover with raw response
+            try:
+                if new_msg_history and len(new_msg_history) > 0:
+                    prediction = new_msg_history[-1].get('text', 'Error')[:500]
+            except:
+                prediction = "Error: Complete extraction failure"

         return prediction, new_msg_history
\end{lstlisting}

\begin{lstlisting}[language=Diff]
diff --git a/task_agent.py b/task_agent.py
--- a/task_agent.py
+++ b/task_agent.py
@@ -17,8 +17,35 @@ class TaskAgent(AgentSystem):
         """
         domain = inputs.get('domain', 'unknown')

-        # Enhanced instruction with chain-of-thought reasoning
-        instruction = f"""You are an expert agent solving tasks...
+        # Domain-specific instructions
+        if domain == 'paper_review':
+            paper_text = inputs.get('paper_text', '')
+            instruction = f"""You are an expert reviewer evaluating academic papers.
+Your task is to decide whether to ACCEPT or REJECT the paper.
+
+Paper to review:
+```
+{paper_text}
+```
+
+Evaluate the paper based on:
+1. Novelty and originality of the research
+2. Technical soundness and methodology
+3. Experimental validation and results
+4. Clarity of presentation
+5. Significance of contributions
+
+Provide your decision in JSON format:
+<json>
+{{
+    "reasoning": "Brief explanation of your decision (2-3 sentences)",
+    "response": "accept or reject"
+}}
+</json>
+
+IMPORTANT: The 'response' field must be EXACTLY either 'accept' or 'reject'
+(lowercase, one word only)."""
+        else:
+            # Generic instruction for other domains
+            instruction = f"""You are an expert agent solving tasks in the '{domain}' domain.
\end{lstlisting}

\begin{lstlisting}[language=Diff]
diff --git a/task_agent.py b/task_agent.py
--- a/task_agent.py
+++ b/task_agent.py
@@ -20,24 +20,43 @@ class TaskAgent(AgentSystem):
         # Domain-specific instructions
         if domain == 'paper_review':
             paper_text = inputs.get('paper_text', '')
-            instruction = f"""You are an expert reviewer evaluating academic papers...
+            instruction = f"""You are a rigorous and critical academic reviewer
+evaluating papers for a top-tier conference. Your task is to decide whether to
+ACCEPT or REJECT the paper. Be skeptical and thorough.

 Paper to review:
 ```
 {paper_text}
 ```

-Evaluate the paper based on:
-1. Novelty and originality of the research
-...
+Evaluate the paper critically based on:
+1. **Novelty and Originality**: Is the contribution truly novel or incremental?
+2. **Technical Soundness**: Are methods rigorous? Any flaws in theory?
+3. **Experimental Validation**: Comprehensive? Sufficient baselines?
+4. **Clarity and Presentation**: Is writing clear? Key details missing?
+5. **Significance**: Does this advance the field meaningfully?
+
+**RED FLAGS that typically warrant REJECTION:**
+- Incremental improvements without significant novelty
+- Missing critical experimental comparisons or baselines
+- Unclear methodology or missing implementation details
+- Overclaimed contributions not supported by evidence
+- Poor writing that obscures understanding
+- Limited experimental validation or weak results
+- Lack of theoretical justification or soundness issues
+- Insufficient comparison with prior state-of-the-art
+
+**Standards for ACCEPTANCE:**
+- Clear novel contribution that advances the field
+- Rigorous methodology with sound theoretical foundation
+- Comprehensive experiments with strong results
+- Well-written with clear explanations
+- Properly compared against relevant baselines
+
+**Be Critical**: Default to REJECT unless the paper clearly meets high standards.

 Provide your decision in JSON format:
 <json>
 {{
-    "reasoning": "Brief explanation of your decision (2-3 sentences)",
+    "reasoning": "Brief critical analysis (2-3 sentences on key strengths/weaknesses)",
     "response": "accept or reject"
 }}
 </json>
\end{lstlisting}

\begin{lstlisting}[language=Diff]
diff --git a/task_agent.py b/task_agent.py
--- a/task_agent.py
+++ b/task_agent.py
@@ -20,43 +20,63 @@ class TaskAgent(AgentSystem):
         # Domain-specific instructions
         if domain == 'paper_review':
             paper_text = inputs.get('paper_text', '')
-            instruction = f"""You are a rigorous and critical academic reviewer...
+            instruction = f"""You are a critical academic reviewer for a top-tier
+conference. Use a two-stage evaluation process:

 Paper to review:
 ```
 {paper_text}
 ```

-Evaluate the paper critically based on:
-1. **Novelty and Originality**: Is the contribution truly novel?
-...
-
-**RED FLAGS that typically warrant REJECTION:**
-...
-
-**Standards for ACCEPTANCE:**
-...
-
-**Be Critical**: Default to REJECT unless the paper clearly meets high standards.
-
-Provide your decision in JSON format:
+**STAGE 1: Identify Weaknesses**
+Systematically check for these common issues:
+
+1. **Novelty Issues:**
+   - Is this just an incremental modification of existing work?
+   - Are the contributions overclaimed or trivial?
+
+2. **Methodological Flaws:**
+   - Are key implementation details missing?
+   - Are there questionable assumptions or theoretical gaps?
+
+3. **Experimental Weaknesses:**
+   - Missing important baselines or comparisons?
+   - Weak or marginal improvements?
+
+4. **Presentation Problems:**
+   - Unclear writing or poor organization?
+   - Missing critical details for reproducibility?
+
+5. **Limited Scope:**
+   - Tested only on toy problems?
+   - Narrow applicability?
+
+**STAGE 2: Make Decision**
+A paper should be **REJECTED** if:
+- It has 2 or more significant weaknesses from Stage 1
+- It has 1 critical flaw
+- The contribution is too incremental
+
+A paper should be **ACCEPTED** only if:
+- It has clear novel contributions
+- Strong experimental validation
+- Well-written and technically sound
+- At most minor, fixable issues
+
+**Critical Mindset:** Most papers have significant flaws.
+About 50% of papers should be rejected.
+
+Provide your evaluation in JSON format:
 <json>
 {{
-    "reasoning": "Brief critical analysis (2-3 sentences)",
+    "weaknesses_found": "List 2-3 main weaknesses identified in Stage 1",
+    "reasoning": "Based on the weaknesses, explain why you accept or reject",
     "response": "accept or reject"
 }}
 </json>
\end{lstlisting}

\begin{lstlisting}[language=Diff]
diff --git a/task_agent.py b/task_agent.py
--- a/task_agent.py
+++ b/task_agent.py
@@ -20,68 +20,71 @@ class TaskAgent(AgentSystem):
         # Domain-specific instructions
         if domain == 'paper_review':
             paper_text = inputs.get('paper_text', '')
-            instruction = f"""You are a critical academic reviewer...
-
-**STAGE 1: Identify Weaknesses**
-...
+            instruction = f"""You are a rigorous peer reviewer for a top-tier
+academic conference. Your reputation depends on making careful, balanced decisions.

-**STAGE 2: Make Decision**
-...
-
-Provide your evaluation in JSON format:
-<json>
-{{
-    "weaknesses_found": "List 2-3 main weaknesses identified in Stage 1",
-    "reasoning": "Based on the weaknesses, explain why you accept or reject",
-    "response": "accept or reject"
-}}
-</json>
-
-IMPORTANT: The 'response' field must be EXACTLY either 'accept' or 'reject'..."""
+**Paper to Review:**
+{paper_text}

+**Evaluation Framework:**

+**ACCEPTANCE CRITERIA (ALL must be satisfied):**
+1. **Novelty**: Presents genuinely new ideas, not incremental improvements
+2. **Technical Soundness**: Methodology is rigorous, correct, and well-justified
+3. **Significance**: Makes important contributions that advance the field
+4. **Experimental Validation**: Claims thoroughly supported with comprehensive experiments
+5. **Clarity**: Clear, well-organized presentation with proper motivation
+6. **Reproducibility**: Sufficient detail for replication
+7. **Related Work**: Proper comparison with existing methods

+**REJECTION CRITERIA (ANY one can warrant rejection):**
+- Lacks novelty; merely combines existing techniques without insight
+- Methodological flaws or incorrect technical approach
+- Insufficient experimental validation or cherry-picked results
+- Missing critical baselines or unfair comparisons
+- Overclaimed contributions not supported by evidence
+- Poor writing quality that obscures technical content
+- Ethical concerns or reproducibility issues
+- Limited scope or significance

+**Decision Guidelines:**
+- **ACCEPT**: Paper clearly satisfies ALL acceptance criteria with strong evidence
+- **REJECT**: Paper fails one or more acceptance criteria OR matches rejection criteria
+- **When in doubt, err on the side of rejection**

+**Your Review Process:**
+1. Identify the paper's main claims and contributions
+2. Evaluate each acceptance criterion systematically
+3. Check for rejection criteria
+4. Ask: "Does this paper significantly advance the field?"
+5. Make a decision based on evidence, not superficial features

+**Response Format:**
+Provide your decision in this exact JSON format:
+<json>
+{{
+    "response": "accept"
+}}
+</json>

+OR

+<json>
+{{
+    "response": "reject"
+}}
+</json>

+**CRITICAL REQUIREMENTS:**
+- The "response" field must contain ONLY "accept" or "reject" (lowercase)
+- Be thorough and critical - rejecting weak papers maintains scientific standards
+- Your decision should be defensible based on the criteria above
+- ONLY output the single word: "accept" or "reject" """
\end{lstlisting}

\subsubsection{Robotics Reward Design}
\label{app:best-agents-genesis}
Diff patches contributing to the best task agent discovered by the DGM-H (\Cref{sec:results-task}) for robotics reward design:
% gen_1
\begin{lstlisting}[language=Diff]
diff --git a/task_agent.py b/task_agent.py
index 3798256..98e6706 100644
--- a/task_agent.py
+++ b/task_agent.py
@@ -5,7 +5,7 @@ from utils.common import extract_jsons
 class TaskAgent(AgentSystem):
     def forward(self, inputs):
         """
-        An agent that solves a given task.
+        An agent that solves a given task with enhanced reasoning and error handling.
         ...
         """
-        domain = inputs['domain']
-        instruction = f"""You are an agent.
+        domain = inputs.get('domain', 'general')
+
+        # Enhanced instruction with clearer structure and reasoning guidance
+        instruction = f"""You are an expert AI agent specialized in solving complex tasks.

-Task input:
+Task Domain: {domain}
+
+Task Input:
 ```
 {inputs}
 ```

+Instructions:
+1. Carefully analyze the task input and identify the key requirements
+2. Break down the problem into logical steps if needed
+3. Formulate your response based on the task requirements
+4. Provide your final answer in the specified JSON format
+
 Respond in JSON format with the following schema:
 <json>
 {{
-    "response": ...
+    "reasoning": "Brief explanation of your approach",
+    "response": "Your final answer here"
 }}
-</json>"""
-        new_msg_history = chat_with_agent(instruction, model=self.model, msg_history=[], logging=self.log)
+</json>

-        # Extract the response
+IMPORTANT: Always include both 'reasoning' and 'response' fields in your JSON output."""
+
+        # Attempt to get response with retry logic
+        max_retries = 2
         prediction = "None"
-        try:
-            extracted_jsons = extract_jsons(new_msg_history[-1]['text'])
-            if extracted_jsons is not None and "response" in extracted_jsons[-1]:
-                prediction = extracted_jsons[-1]['response']
-        except Exception as e:
-            self.log(f"Error extracting prediction: {e}")
-            prediction = "None"
+        new_msg_history = []
+
+        for attempt in range(max_retries + 1):
+            try:
+                new_msg_history = chat_with_agent(...)
+                # Extract the response with retry logic
+                if new_msg_history and len(new_msg_history) > 0:
+                    extracted_jsons = extract_jsons(new_msg_history[-1]['text'])
+                    if extracted_jsons is not None and len(extracted_jsons) > 0:
+                        response_json = extracted_jsons[-1]
+                        if "reasoning" in response_json:
+                            self.log(f"Agent reasoning: {response_json['reasoning']}")
+                        if "response" in response_json:
+                            prediction = response_json['response']
+                            break
+            except Exception as e:
+                self.log(f"Error in attempt {attempt + 1}: {str(e)}")

         return prediction, new_msg_history
\end{lstlisting}

\begin{lstlisting}[language=Diff]
diff --git a/task_agent.py b/task_agent.py
index 3c2dd8a..faab41d 100644
--- a/task_agent.py
+++ b/task_agent.py
@@ -27,30 +27,96 @@ Paper Text:
+Critical Evaluation Guidelines:
+- Be rigorous and selective - most submissions have issues
+- Accept ONLY papers that are strong across MOST criteria (4-5/5)
+- Reject papers with significant weaknesses in multiple areas
+- A paper needs to be clearly above average to warrant acceptance
+- Consider: Would this paper strengthen the conference program?
+- If a paper is borderline or has major concerns, lean toward REJECT

+        elif domain == 'genesis_go2walking':
+            # Genesis domain-specific instruction with environment attribute documentation
+            task_description = inputs.get('task_description', '')
+            genesis_env_path = inputs.get('genesis_environment_path', '')
+            default_reward_function = inputs.get('default_reward_function', '')
+
+            instruction = f"""You are an expert in reinforcement learning and reward function design.

+Task: {task_description}

+Default Reward Function:
+```python
+{default_reward_function}
+```

+AVAILABLE ENVIRONMENT ATTRIBUTES (use ONLY these):
+- env.commands: [num_envs, 3] - velocity commands [lin_vel_x, lin_vel_y, ang_vel_z]
+- env.base_lin_vel: [num_envs, 3] - base linear velocity [x, y, z]
+- env.base_ang_vel: [num_envs, 3] - base angular velocity [roll, pitch, yaw]
+- env.base_pos: [num_envs, 3] - base position [x, y, z]
+- env.base_quat: [num_envs, 4] - base orientation quaternion
+- env.projected_gravity: [num_envs, 3] - gravity vector in base frame
+- env.dof_pos: [num_envs, 12] - joint positions
+- env.dof_vel: [num_envs, 12] - joint velocities
+- env.actions: [num_envs, 12] - current actions
+- env.last_actions: [num_envs, 12] - previous actions
+- env.last_dof_vel: [num_envs, 12] - previous joint velocities

+IMPORTANT CONSTRAINTS:
+- DO NOT use env.torques (not available)
+- DO NOT access undefined attributes
+- Use only the attributes listed above
+- All operations must work with PyTorch tensors
\end{lstlisting}

% gen_12
\begin{lstlisting}[language=Diff]
diff --git a/task_agent.py b/task_agent.py
index faab41d..62c444f 100644
--- a/task_agent.py
+++ b/task_agent.py
@@ -88,19 +90,24 @@ AVAILABLE ENVIRONMENT ATTRIBUTES (use ONLY these):
-IMPORTANT CONSTRAINTS:
-- DO NOT use env.torques (not available)
-- DO NOT access undefined attributes
-- Use only the attributes listed above
-- All operations must work with PyTorch tensors
+CRITICAL RULES (VIOLATION WILL CAUSE IMMEDIATE FAILURE):
+1. DO NOT reference env.torques or env.dt in your code - THESE DO NOT EXIST
+2. DO NOT add hasattr() checks for unavailable attributes and then use them
+3. DO NOT assume any attributes beyond the exact list above
+4. NEVER write code like: if hasattr(env, 'torques'): use env.torques
+5. All operations must work with PyTorch tensors
+6. If you need time differences, use fixed timestep: dt = 0.02
+7. ONLY use attributes from the AVAILABLE list - nothing else exists

+                            # Post-process prediction based on domain
+                            if domain == 'paper_review':
+                                # Ensure response is exactly "accept" or "reject"
+                                prediction_lower = str(prediction).lower().strip()
+                                if 'accept' in prediction_lower and 'reject' not in prediction_lower:
+                                    prediction = 'accept'
+                                elif 'reject' in prediction_lower:
+                                    prediction = 'reject'
+                                else:
+                                    self.log(f"Warning: Ambiguous response '{prediction}', defaulting to 'reject'")
+                                    prediction = 'reject'
\end{lstlisting}

% gen_13
\begin{lstlisting}[language=Diff]
diff --git a/task_agent.py b/task_agent.py
index 62c444f..fb504a2 100644
--- a/task_agent.py
+++ b/task_agent.py
@@ -40,26 +40,30 @@ Critical Evaluation Guidelines:
-- If a paper is borderline or has major concerns, lean toward REJECT
+- Balance your decisions: aim for appropriate selectivity (not too lenient, not too harsh)
+- If a paper is borderline, carefully weigh the evidence on both sides

 Reward Function Design Guidelines:
 1. Primary reward: Track forward velocity command (env.commands[:, 0] vs env.base_lin_vel[:, 0])
+   - Use exponential reward: torch.exp(-error * temperature) for smooth gradients
+   - Typical temperature: 1.0-2.0, scale: 0.8-1.0
 2. Secondary rewards: Track angular velocity, maintain base height, stable orientation
-3. Penalties: Excessive action changes (use env.actions - env.last_actions)
-4. Use exponential rewards: torch.exp(-error * temperature) for smooth gradients
-5. Scale rewards appropriately (main objectives: 0.5-1.0, penalties: -0.5 to -0.01)
-6. For smooth action changes: use torch.sum(torch.square(env.actions - env.last_actions))
-7. For velocity smoothness: use torch.sum(torch.square(env.dof_vel - env.last_dof_vel))
-8. Return: total_reward, reward_components dict, reward_scales dict
+   - Angular velocity tracking: similar to linear velocity
+   - Base height: penalize deviation from target (typically 0.32m)
+   - Orientation: use env.projected_gravity to penalize tilt
+3. Penalties for smooth and stable locomotion:
+   - Action smoothness: -0.01 * torch.sum(torch.square(env.actions - env.last_actions), dim=1)
+   - Joint velocity smoothness: -0.001 * torch.sum(torch.square(env.dof_vel - env.last_dof_vel), dim=1)
+   - Excessive joint velocities: -0.0005 * torch.sum(torch.square(env.dof_vel), dim=1)
+   - Unwanted lateral movement: penalize env.base_lin_vel[:, 1]
+4. Balance reward components:
+   - Main tracking objectives: 0.5-1.0 scale
+   - Secondary objectives: 0.2-0.5 scale
+   - Smoothness penalties: -0.01 to -0.001 scale
+5. Return: total_reward, reward_components dict, reward_scales dict
\end{lstlisting}

\subsubsection{Olympiad-level Math Grading}
\label{app:best-agents-imograding}
Diff patches contributing to the best task agent, which we refer to as BetterGrader in \Cref{app:res-imograders}, discovered by the DGM-H with transfer and from ProofAutoGrader (\Cref{sec:results-compound}) for Olympiad-level math grading:
% gen_3
\begin{lstlisting}[language=Diff]
diff --git a/task_agent.py b/task_agent.py
index 9ab8761..c00d167 100644
--- a/task_agent.py
+++ b/task_agent.py
@@ -11,10 +11,25 @@ Keep in mind the standards at the IMO are extremely high...

 ### General Scoring Rubric
 Scores are assigned on a 0-7 scale. The general guidelines are:
-* **7 Points (Correct):** The solution is complete, correct, and fully rigorous...
-* **6 Points (Almost Correct):** The solution is almost correct with a sound core argument, but contains minor errors...
-* **1 Point (Partial Progress):** The solution demonstrates substantial progress explicitly mentioned in the grading guidelines...
-* **0 Points (Incorrect):** The solution doesn't make substantial progress...
+
+* **7 Points (Correct):** The solution is complete, correct, and fully rigorous with no gaps or errors. All major steps are proven with full rigor. Every claim is justified or routine to verify...
+
+* **6 Points (Almost Correct):** The solution has ALL the major ideas and the core argument structure is sound, but contains ONE OR MORE of these minor issues:
+  - Minor algebraic/arithmetic errors that don't affect the main argument
+  - Small logical gaps that are straightforward to fill
+  - Missing routine verifications that an expert could easily supply
+  - **NOT eligible for 6 points**: Missing proofs for major lemmas, unjustified non-trivial claims, incomplete case analysis, or fundamental logical gaps
+  - **Key test**: Would an expert say "this is essentially correct, just needs minor cleanup"?
+
+* **1 Point (Partial Progress):** The solution demonstrates substantial progress on a KEY component that is explicitly mentioned in the grading guidelines for partial credit.
+  - **CRITICAL**: Carefully read the Specific Grading Guidelines section to see what counts as partial credit
+  - The solution must achieve one of the specific milestones listed in the guidelines
+  - Must make non-trivial progress toward the solution (not just initial observations)
+  - Reformulating the problem without making substantive headway is NOT sufficient
+  - Proving lemmas NOT mentioned in grading guidelines is NOT sufficient
+  - **If the guidelines list specific achievements for partial credit and the solution achieves ANY of them, award 1 point**
+
+* **0 Points (Incorrect):** The solution doesn't make substantial progress on key steps mentioned in grading guidelines, is fundamentally flawed, or makes only trivial observations.

 ### Evaluation Process
 You must follow this structured process:
-1. **Analyze References:** Meticulously read and understand the problem...
-2. **Step-by-Step Verification:** Verify the logical validity and rigor of every step...
-3. **Assess Progress:** Determine the extent of non-trivial progress made.
-4. **Score Determination:** Compare the findings against the Specific Grading Guidelines...
+
+1. **Analyze References:** Meticulously read and understand the problem and Ground Truth Solution. Carefully review the Specific Grading Guidelines to identify:
+   - The key steps required for a complete solution
+   - **What specific progress qualifies for partial credit (1 point)** - this is crucial!
+   - What distinguishes "almost correct" (6 points) from "correct" (7 points)
+
+2. **Step-by-Step Verification:** Verify the logical validity and rigor of EVERY step in the proposed solution:
+   - Identify ALL flaws, gaps, assumptions, and errors
+   - Check if gaps are "minor and routine" (possibly 6 pts) or "major" (0-1 pts)
+   - **Be careful**: Some solutions may appear correct but have hidden gaps or unjustified leaps
+   - Distinguish between minor calculation errors vs fundamental logical flaws
+
+3. **Assess Progress Against Grading Guidelines:**
+   - **Does the solution achieve the specific milestones mentioned for partial credit?**
+   - Does it have all major components with only minor fixable issues (6 points)?
+   - Or does it have complete rigor with no gaps (7 points)?
+
+4. **Score Determination:** Apply this decision tree:
+   - **If solution is complete and rigorous with no errors or gaps -> 7 points**
+   - **If solution has all major ideas but minor fixable issues -> 6 points**
+   - **If solution makes substantial progress mentioned in grading guidelines -> 1 point**
+   - **Otherwise -> 0 points**

+### Critical Reminders for Accurate Grading
+
+**Common Grading Errors to AVOID:**
+1. **Being too lenient with 7 points**: If there are ANY gaps (even minor ones that need filling), it's 6 points, not 7.
+2. **Missing partial credit**: Check the grading guidelines carefully - if the solution achieves ANY milestone mentioned for partial credit, award 1 point.
+3. **Confusing "good attempt" with "almost correct"**: 6 points requires ALL major ideas to be present, not just a good start.
+4. **Ignoring unjustified claims**: Statements like "it's easy to see" or "clearly" must actually be clear/easy. If non-trivial, it's a gap.

+**Score Distribution Calibration:**
+- Most solutions will be 0 or 7 points (either fundamentally flawed or correct)
+- 6 points should be rare (only when truly "almost there" with all ideas present)
+- 1 point should match specific milestones in grading guidelines
+- When in doubt, be strict: IMO standards are extremely high
\end{lstlisting}

% gen_9
\begin{lstlisting}[language=Diff]
diff --git a/task_agent.py b/task_agent.py
index c00d167..72475b6 100644
--- a/task_agent.py
+++ b/task_agent.py
@@ -28,8 +28,11 @@ Scores are assigned on a 0-7 scale...
   - Proving lemmas NOT mentioned in grading guidelines is NOT sufficient
   - **If the guidelines list specific achievements for partial credit and the solution achieves ANY of them, award 1 point**
+  - **IMPORTANT**: Even if the solution has major flaws or is incomplete, if it achieves ANY specific milestone from the guidelines, it deserves 1 point, not 0

 * **0 Points (Incorrect):** The solution doesn't make substantial progress...
+  - **CRITICAL CHECK**: Before assigning 0 points, verify that the solution does NOT achieve ANY of the partial credit milestones listed in the grading guidelines
+  - If even ONE milestone is achieved, the score should be 1 point, not 0

+3. **MANDATORY Partial Credit Milestone Check:**
+   - **THIS STEP IS REQUIRED - DO NOT SKIP**
+   - List each partial credit milestone explicitly from the grading guidelines
+   - For EACH milestone, determine: Does the solution achieve this? YES/NO
+   - If ANY milestone shows YES -> the score must be at least 1 point
+   - This check must happen BEFORE considering 0 points

+5. **Score Determination:** Apply this decision tree:
+   - **FIRST**: Did the solution achieve ANY partial credit milestone from the guidelines?
+     - If YES -> Score is AT LEAST 1 point (proceed to check for higher scores)
+     - If NO -> Score is 0 points (stop here)
+
+   - **If score >= 1, check for higher scores:**
+     - Is the solution complete and rigorous with no errors or gaps? -> 7 points
+     - Does it have ALL major ideas with only minor fixable issues? -> 6 points
+     - Otherwise -> 1 point

+**REQUIRED EVALUATION FORMAT:**
+You MUST structure your response as follows:
+
+1. **List Partial Credit Milestones**: Explicitly list each milestone from the grading guidelines
+2. **Check Each Milestone**: For each milestone, state whether the solution achieves it (YES/NO)
+3. **Determine Minimum Score**: If ANY milestone is YES, the minimum score is 1 point
+4. **Detailed Analysis**: Provide your complete evaluation
+5. **Final Score**: Provide the score in the required format
\end{lstlisting}

% gen_48
\begin{lstlisting}[language=Diff]
diff --git a/task_agent.py b/task_agent.py
index 72475b6..62e975b 100644
--- a/task_agent.py
+++ b/task_agent.py
+* **7 Points (Correct):** The solution is complete, correct, and fully rigorous with no gaps or errors. All major steps are proven with full rigor. Every claim is justified or truly routine to verify...
+  - **Critical requirements**: Solution must (1) address ALL parts of the problem, (2) prove ALL necessary claims, and (3) have NO unjustified leaps
+  - Every "clearly" or "obviously" must be genuinely trivial to an IMO expert
+  - All edge cases, special cases, and boundary conditions must be handled

+* **6 Points (Almost Correct):** The solution has ALL the major ideas and the core argument structure is sound, but contains ONE OR MORE of these **minor** issues:
+  - Minor algebraic/arithmetic errors that don't affect the main argument (e.g., writing 2n+1 instead of 2n+2 but the logic still holds)
+  - Small logical gaps that are straightforward to fill (e.g., "clearly" statements that are indeed clear to an expert)
+  - Missing routine verifications that an expert could easily supply (e.g., obvious algebra steps)
+  - **NOT eligible for 6 points**: Missing proofs for major lemmas, unjustified non-trivial claims, incomplete case analysis, fundamental logical gaps, or missing key components
+  - **Key test**: Would an expert say "this is essentially correct, just needs minor cleanup"? The solution structure is complete and sound.
+  - **Example of 6 points**: A proof that has all the right ideas and structure but makes a small computational error that doesn't invalidate the approach
+  - **Example of NOT 6 points**: A proof that sketches the right approach but leaves out the proof of a crucial intermediate result

+2. **Step-by-Step Verification:** Verify the logical validity and rigor of EVERY step in the proposed solution:
+   - Identify ALL flaws, gaps, assumptions, and errors
+   - For EACH "clearly" or "obviously" statement: Is it truly routine or does it hide significant work?
+   - Check if gaps are "minor and routine" (possibly 6 pts) or "major" (0-1 pts)
+   - **Be careful**: Some solutions may appear correct but have hidden gaps or unjustified leaps
+   - Distinguish between minor calculation errors vs fundamental logical flaws
+   - **Rigor check**: Are intermediate results properly proven? Are all cases covered? Are inequalities/equalities justified?

+4. **Assess Overall Quality and Completeness:**
+   - **For 7 points**: Check that EVERY claim is proven, EVERY case is handled, NO unjustified leaps exist
+   - **For 6 points**: Verify ALL major components are present and the solution structure is complete (not just a good start)
+   - **For 1 point**: Confirm at least ONE specific milestone from guidelines is achieved
+   - **For 0 points**: Verify NO milestones from guidelines are achieved

+**Examples to Calibrate Your Judgment:**
+- **7 points**: "The proof correctly establishes X by showing Y, handles all cases including Z, and proves every intermediate claim with full rigor."
+- **6 points**: "The proof has the complete structure and all major steps, but writes 'it is clear that' for a non-trivial step that needs 2-3 lines to verify."
+- **1 point**: "The solution proves Lemma A which is specifically mentioned in grading guidelines as worth partial credit, but doesn't complete the full proof."
+- **0 points**: "The solution makes interesting observations and tries several approaches, but doesn't achieve any of the specific milestones listed in the guidelines."
\end{lstlisting}

% gen_54
\begin{lstlisting}[language=Diff]
diff --git a/task_agent.py b/task_agent.py
index 62e975b..a033aa1 100644
--- a/task_agent.py
+++ b/task_agent.py
+### CRITICAL: Four-Category Classification System
+
+Before diving into details, first classify the solution into ONE of these four categories:
+
+1. **COMPLETE & RIGOROUS**: Has all components, all proofs, handles all cases, no gaps -> 7 points
+2. **NEARLY COMPLETE**: Has complete structure + all major ideas, only minor polish needed -> 6 points (RARE: ~5%)
+3. **MEANINGFUL PROGRESS**: Achieves at least ONE specific milestone from grading guidelines -> 1 point (~25%)
+4. **INSUFFICIENT**: No specific milestones achieved, only trivial observations -> 0 points
+
+**Key Decision Points:**
+- 7 vs 6: "Any gaps at all, even fixable ones?" If yes -> 6
+- 6 vs 1: "Complete proof structure with ALL major steps present?" If no -> must be 1 or 0
+- 1 vs 0: "Achieves ANY milestone listed in guidelines?" If no -> 0

+* **1 Point (Partial Progress):** The solution demonstrates substantial progress on a KEY component that is explicitly mentioned in the grading guidelines for partial credit.
+  - **CRITICAL**: You MUST check the "Specific Grading Guidelines" section below - it lists EXACTLY what qualifies for partial credit
+  - The solution must achieve at least ONE of the specific milestones listed in those guidelines
+  - **IMPORTANT DISTINCTIONS**:
+    * DOES count: Achieving a milestone listed in the guidelines (even with errors elsewhere)
+    * Does NOT count: General progress, clever observations, or lemmas NOT in the guidelines
+    * Does NOT count: Reformulating the problem without substantive progress
+    * Does NOT count: Incorrect attempts that seem "on the right track"
+  - **Examples of valid partial credit**:
+    * Guidelines say "partial credit for proving Lemma X" -> solution proves Lemma X correctly -> 1 point
+    * Guidelines say "partial credit for establishing the recurrence relation" -> solution establishes it -> 1 point
+  - **Examples that do NOT qualify for partial credit**:
+    * Solution tries several approaches but none matches a guideline milestone -> 0 points
+    * Solution proves a useful lemma not mentioned in guidelines -> 0 points (no matter how clever)
+    * Solution has the "right idea" but doesn't complete any guideline milestone -> 0 points

+**Common Grading Errors to AVOID (based on actual evaluation data):**
+
+1. **ERROR #1: Missing partial credit (11 cases last eval - partial -> incorrect)**
+   - MISTAKE: Marking solutions as 0 points that actually achieve guideline milestones
+   - FIX: Before assigning 0, explicitly list EACH milestone and check if achieved
+   - **Process**: "Does solution achieve milestone 1? NO. Milestone 2? NO. ..." Only if ALL are NO -> 0 points
+
+2. **ERROR #2: Awarding partial to incorrect solutions (7 cases - incorrect -> partial)**
+   - MISTAKE: Giving 1 point for "interesting work" or "right direction" not in guidelines
+   - FIX: Only award 1 point if solution achieves an EXACT milestone listed in guidelines
+
+3. **ERROR #3: Confusing "almost" with "correct" (5 cases - almost -> correct)**
+   - MISTAKE: Awarding 7 points when gaps exist (even minor, fillable gaps)
+   - FIX: If ANY gap needs filling (even routine) -> 6 points, NOT 7
+
+4. **ERROR #4: Over-rewarding partial progress (5 cases - partial -> correct)**
+   - MISTAKE: Awarding 7 points to solutions that achieve only some milestones
+   - FIX: Check if solution solves ENTIRE problem or just proves partial results

+5. **ERROR #5: Over-using 6 points**
+   - MISTAKE: Awarding 6 for "good progress" or "most of the way there"
+   - FIX: 6 requires COMPLETE solution structure with ALL major steps, just minor polish needed
+   - **Frequency check**: 6 points should be RARE (~5% of solutions)

+**Score Distribution Calibration (Expected Distribution):**
+- ~35% score 7 (correct): Complete, rigorous solutions with no gaps
+- ~35% score 0 (incorrect): No guideline milestones achieved
+- ~24% score 1 (partial): Achieve at least one specific guideline milestone
+- ~6% score 6 (almost): RARE - complete structure with only minor fixable issues
\end{lstlisting}

% gen_56
\begin{lstlisting}[language=Diff]
diff --git a/task_agent.py b/task_agent.py
index a033aa1..86320b1 100644
--- a/task_agent.py
+++ b/task_agent.py
 **Key Decision Points:**
 - 7 vs 6: "Any gaps at all, even fixable ones?" If yes -> 6
 - 6 vs 1: "Complete proof structure with ALL major steps present?" If no -> must be 1 or 0
-- 1 vs 0: "Achieves ANY milestone listed in guidelines?" If no -> 0
+- **1 vs 0 (MOST CRITICAL)**: "Does the solution EXPLICITLY ACHIEVE any milestone from the grading guidelines?"
+  * Read each milestone carefully and check if solution PROVES/ESTABLISHES it
+  * "Attempts" or "makes progress toward" a milestone != ACHIEVES it
+  * Must have concrete proof/result that matches milestone description
+  * If uncertain, re-read the milestone requirements and check for explicit completion

+    * Does NOT count: Stating a result without proving it
+    * Does NOT count: Getting "close" to a milestone but not completing it
+    * Solution states a milestone result but doesn't prove it -> 0 points
+    * Solution proves 80% of what's needed for a milestone -> 0 points (must be complete)
+  - **VERIFICATION CHECKLIST for each milestone**:
+    1. What exactly does the milestone require?
+    2. Does the solution provide a complete proof/derivation of this?
+    3. Are there any gaps or unjustified steps in achieving this milestone?
+    4. If there are gaps, is the milestone still considered "achieved"? (Usually NO)

+2. **Check Each Milestone**: For each milestone, systematically verify:
+   - What the milestone requires (quote from guidelines)
+   - What the solution provides (specific evidence)
+   - Does it FULLY achieve the milestone? (YES/NO with clear reasoning)
+   - Key distinction: "attempts" or "partial progress" toward milestone = NO

+**CRITICAL REMINDER FOR MILESTONE CHECKING:**
+- Be precise: A milestone is achieved only if the solution COMPLETES what the milestone describes
+- Common mistake: Giving credit for "working toward" a milestone (this should be 0 points)
+- When in doubt: Re-read the milestone requirement and check if solution fully satisfies it
\end{lstlisting}

% gen_71
\begin{lstlisting}[language=Diff]
diff --git a/task_agent.py b/task_agent.py
index 86320b1..f1dd3ac 100644
--- a/task_agent.py
+++ b/task_agent.py
 Before diving into details, first classify the solution into ONE of these four categories:

-1. **COMPLETE & RIGOROUS**: Has all components, all proofs, handles all cases, no gaps -> 7 points
+1. **COMPLETE & RIGOROUS**: Has all components, all proofs, handles all cases, ZERO gaps -> 7 points
 2. **NEARLY COMPLETE**: Has complete structure + all major ideas, only minor polish needed -> 6 points (RARE: ~5%)
 3. **MEANINGFUL PROGRESS**: Achieves at least ONE specific milestone from grading guidelines -> 1 point (~25%)
 4. **INSUFFICIENT**: No specific milestones achieved, only trivial observations -> 0 points

 **Key Decision Points:**
-- 7 vs 6: "Any gaps at all, even fixable ones?" If yes -> 6
+- **7 vs 6 (CRITICAL - BE STRICT)**: "Is this solution PUBLICATION-READY with ZERO gaps, ZERO unjustified steps, ZERO hand-waving?"
+  * ANY gap, even minor -> automatically 6 points maximum
+  * ANY "clearly", "obviously", "it follows" that needs verification -> 6 points
+  * ANY missing routine verification -> 6 points
+  * ANY computational error, however minor -> 6 points
+  * 7 points requires PERFECTION - when in doubt, give 6 points
 - 6 vs 1: "Complete proof structure with ALL major steps present?" If no -> must be 1 or 0
-- **1 vs 0 (MOST CRITICAL)**: "Does the solution EXPLICITLY ACHIEVE any milestone from the grading guidelines?"
+- **1 vs 0**: "Does the solution EXPLICITLY ACHIEVE any milestone from the grading guidelines?"
   * Read each milestone carefully and check if solution PROVES/ESTABLISHES it
   * "Attempts" or "makes progress toward" a milestone != ACHIEVES it
   * Must have concrete proof/result that matches milestone description

-* **7 Points (Correct):** The solution is complete, correct, and fully rigorous with no gaps or errors...
+* **7 Points (Correct):** The solution is complete, correct, and fully rigorous with ABSOLUTELY NO gaps or errors. All major steps are proven with full rigor. Every claim is justified or truly routine to verify...
+  - **Critical requirements**: Solution must (1) address ALL parts of the problem, (2) prove ALL necessary claims with full detail, and (3) have NO unjustified leaps whatsoever
+  - Every "clearly" or "obviously" must be genuinely trivial to an IMO expert - if it takes >30 seconds to verify, it's not trivial
+  - All edge cases, special cases, and boundary conditions must be explicitly handled
+  - **BE EXTREMELY STRICT**: 7 points should be RARE (target: ~30-40% of solutions). Most good solutions have minor gaps -> give 6 points

-* **6 Points (Almost Correct):** The solution has ALL the major ideas and the core argument structure is sound...
+* **6 Points (Almost Correct):** The solution has ALL the major ideas and the COMPLETE argument structure, but contains ONE OR MORE of these **minor** issues:
   - Minor algebraic/arithmetic errors that don't affect the main argument
-  - Small logical gaps that are straightforward to fill
+  - Small logical gaps that are straightforward to fill (e.g., "clearly" statements that need 1-2 lines to verify)
   - Missing routine verifications that an expert could easily supply
+  - Citations of standard theorems without proof (acceptable at IMO level)
   - **NOT eligible for 6 points**: Missing proofs for major lemmas, unjustified non-trivial claims, incomplete case analysis, fundamental logical gaps, or missing key components
   - **Key test**: Would an expert say "this is essentially correct, just needs minor cleanup"? The solution structure is complete and sound.
+  - **IMPORTANT**: 6 points means the solution is COMPLETE but not PERFECT. If major steps are missing -> give 1 point instead.

 * **1 Point (Partial Progress):** The solution demonstrates substantial progress on a KEY component...
   - **IMPORTANT DISTINCTIONS**:
     * DOES count: Achieving a milestone listed in the guidelines (even with errors elsewhere)
+    * DOES count: Equivalent formulations of guideline milestones (recognize reformulations)
     * Does NOT count: General progress, clever observations, or lemmas NOT in the guidelines
     * Does NOT count: Reformulating the problem without substantive progress
     * Does NOT count: Incorrect attempts that seem "on the right track"
     * Solution proves a useful lemma not mentioned in guidelines -> 0 points (no matter how clever)
+  - **When grading partial credit, be MORE LENIENT**: If the solution substantially achieves a milestone (even with minor gaps), give the point
\end{lstlisting}

\subsection{Qualitative: Improving Task Performance}
\label{app:qual-task}

\begin{figure}[H]
\centering
\includegraphics[width=\textwidth]{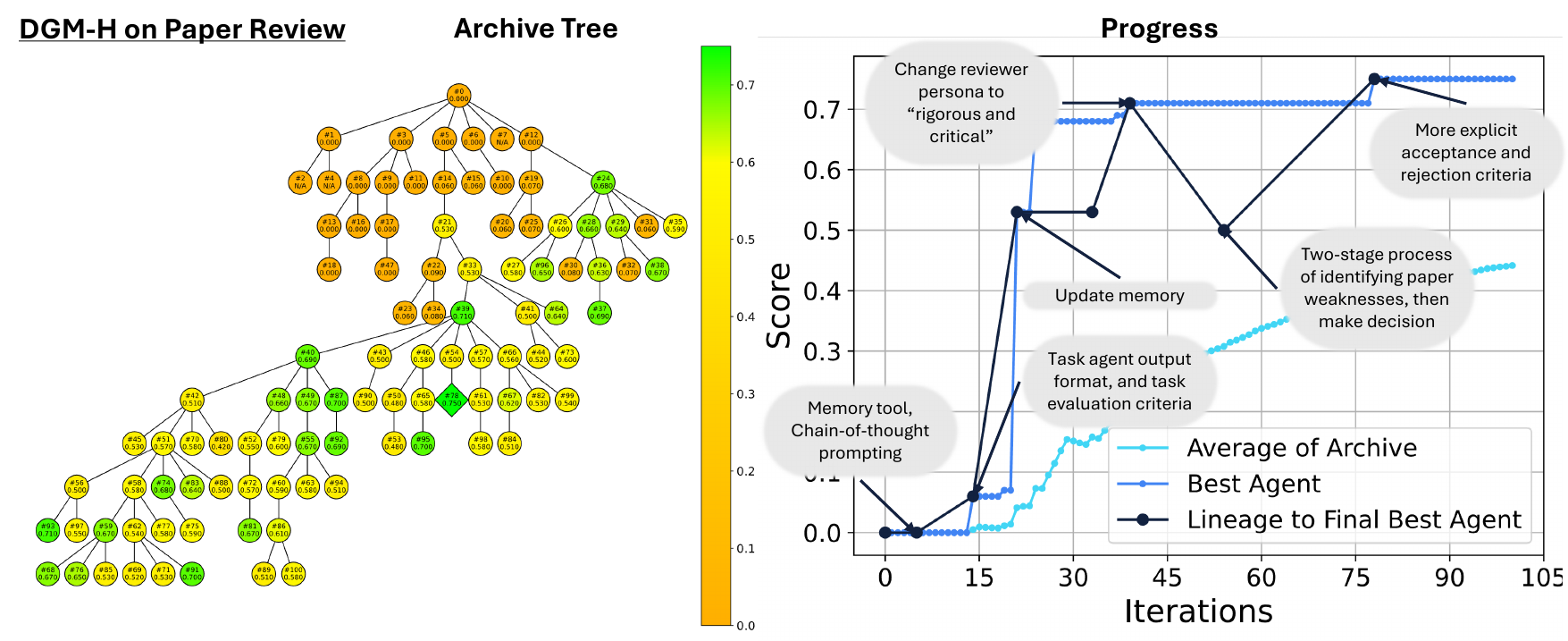}
\caption{\textbf{The DGM-H automatically self-improves to become better at paper review.} (Left) Archive of agents generated during the DGM-H run on paper review and robotics reward design together. Each node represents an agent, with node 0 corresponding to the initial agent. Node color indicates performance on paper review. The best-performing node is shown as a diamond. Edges show which agents self-modified to produce children. (Right) Progress plot of DGM-H on paper review. The light blue line shows the average score of all compiled agents. The blue line tracks the best score achieved by any agent in the archive at each iteration. The dark line shows the lineage of the final best-discovered agent and its precursor nodes.}
\label{fig:task-paperreview}
\end{figure}

\begin{figure}[H]
\centering
\includegraphics[width=\textwidth]{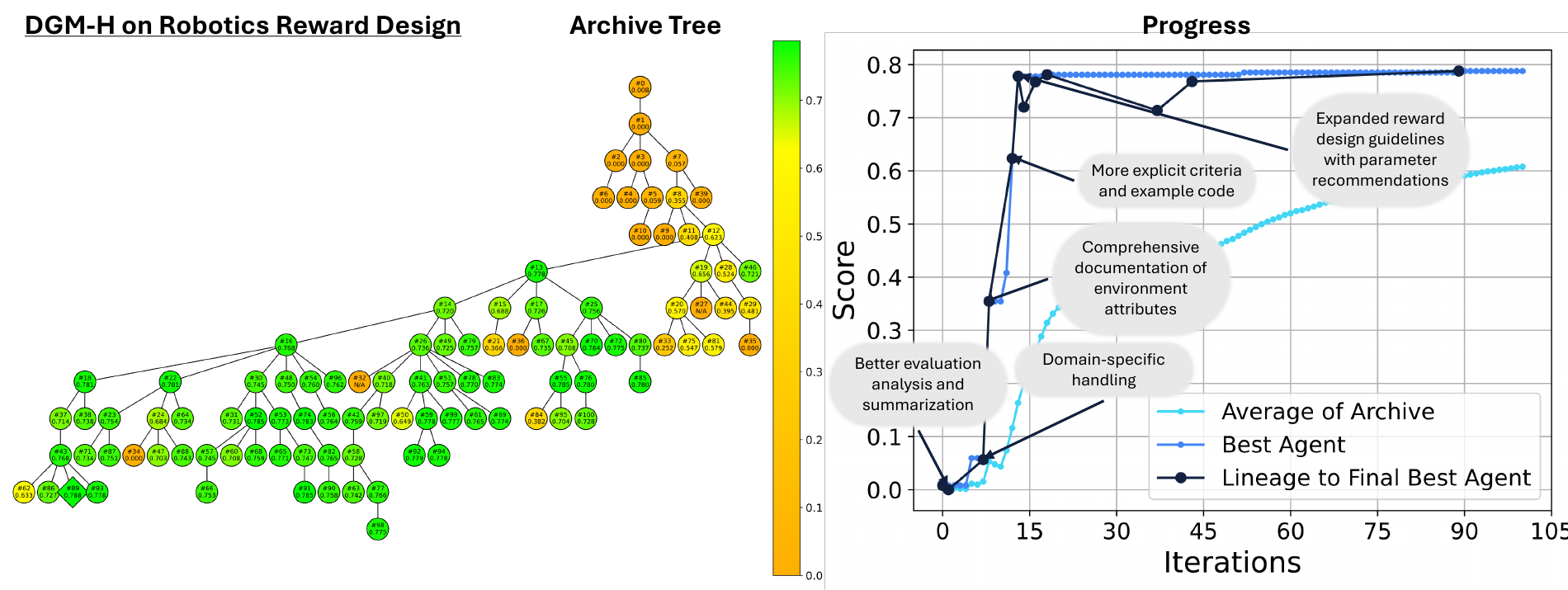}
\caption{\textbf{The DGM-H automatically self-improves to become better at robotics reward design.} (Left) Archive of agents generated during the DGM-H run on paper review and robotics reward design together. Each node represents an agent, with node 0 corresponding to the initial agent. The best-performing node is shown as a diamond. Node color indicates performance on robotics reward design. Edges show which agents self-modified to produce children. (Right) Progress plot of DGM-H on robotics reward design. The light blue line shows the average score of all compiled agents. The blue line tracks the best score achieved by any agent in the archive at each iteration. The dark line shows the lineage of the final best-discovered agent and its precursor nodes.}
\label{fig:task-genesis}
\end{figure}

\begin{figure}[H]
\centering
\includegraphics[width=\textwidth]{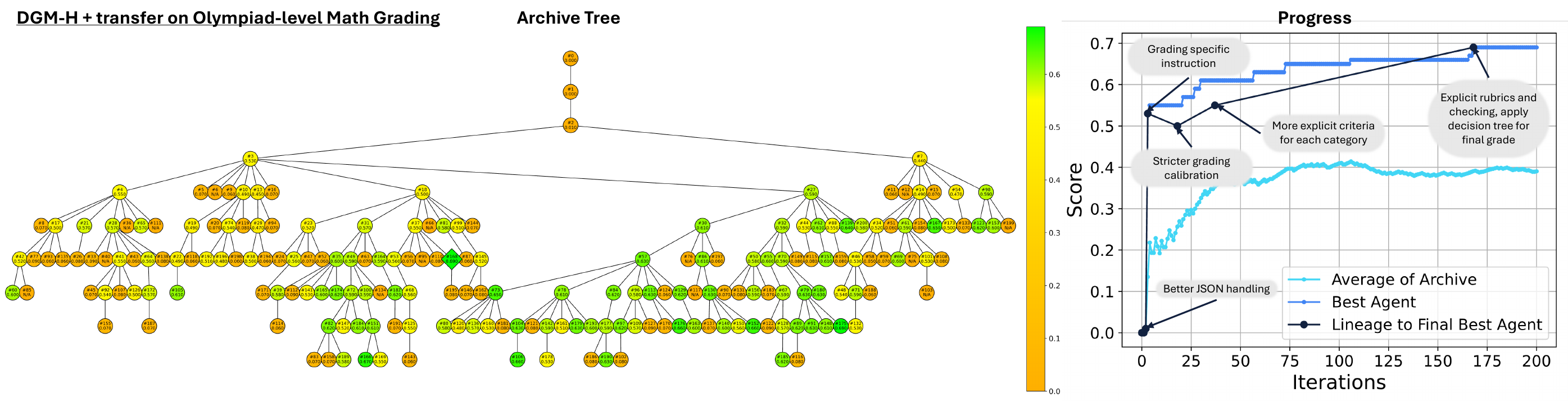}
\caption{\textbf{The DGM-H automatically self-improves to become better at Olympiad-level math grading.} (Left) Archive of agents generated during the DGM-H + transfer run on Olympiad-level math grading. Each node represents an agent, with node 0 corresponding to the initial agent. Node color indicates performance on Olympiad-level math grading. The best-performing node is shown as a diamond. Edges show which agents self-modified to produce children. (Right) Progress plot of DGM-H + transfer on Olympiad-level math grading. The light blue line shows the average score of all compiled agents. The blue line tracks the best score achieved by any agent in the archive at each iteration. The dark line shows the lineage of the final best-discovered agent and its precursor nodes.}
\label{fig:task-imograding}
\end{figure}

Here we provide a qualitative view of how the DGM-H improves task performance over time by visualizing the archive trees and progress plots for one run in each setting (\Cref{sec:results}). For each domain, we annotate key nodes in the archive with the code changes that affected the behavior of the task agent only for the domain being analyzed. Across diverse domains (i.e., paper review, robotics reward design, and Olympiad-level math grading), the DGM-H consistently demonstrates the ability to self-improve in meaningful ways (\Cref{fig:task-paperreview,fig:task-genesis,fig:task-imograding}). Notably, many lineage paths leading to the final best-performing agent pass through intermediate nodes with lower performance, illustrating the benefits of open-ended search, which explores a diverse set of promising stepping stones rather than exclusively branching from the current best solution.

\textbf{Use structured processes, not attitude instructions.} In the paper review domain, the DGM-H transitions from behavioral prompting to structurally grounded decision-making (\Cref{fig:task-paperreview}). In generation 39, the agent attempted to improve performance by adopting a ``rigorous and critical'' reviewer persona, encouraging stricter standards and default rejection. Subsequent analysis showed that such attitude-based instructions were unreliable, leading to a key insight in generation 54: ``for LLMs, use structured processes, not attitude instructions''. The DGM-H therefore introduced a two-stage evaluation procedure in which the agent first identifies weaknesses using an explicit checklist and only then makes an accept/reject decision based on predefined rules. This shift from behavioral guidance to process-level structure enabled more stable and higher-performing review behavior in later generations.

\textbf{Accumulating domain knowledge.} In robotics reward design, the most impactful changes stem from progressively grounding the agent in accurate domain knowledge (\Cref{fig:task-genesis}). A major breakthrough in generation 8 added comprehensive documentation of the target environment, explicitly listing valid state variables and constraints and providing high-level reward design guidelines, which eliminated failures caused by hallucinated attributes. Later generations (12 and 13) iteratively refined this documentation by tightening constraints, adding concrete code examples, and specifying typical reward formulations and scaling ranges. Rather than isolated prompt edits, the DGM-H continuously improved a shared, example-driven knowledge base that supported increasingly effective reward design.

\textbf{Automated rubrics and decision tree.} For Olympiad-level math grading, the DGM-H shows a steady move toward explicit evaluation structure (\Cref{fig:task-imograding}). In generation 3, listing grading categories with clear definitions corrected the tendency to solve problems instead of grading them. Subsequent generations (18 and 37) refined these categories with systematic decision procedures, calibration, and concrete boundary-case examples. Generation 168 introduced explicit rubrics, per-item checklists, and a decision-tree framework mapping rubric satisfaction to final grades, replacing descriptive guidance with precise logical flow and substantially improving grading consistency.
Rather than relying on human-designed rubrics, the DGM-H autonomously discovers evaluative structures that mirror those used in recent rubric-based approaches to improve consistency and interpretability in complex judgments \citep{cook2024ticking, fan2024sedareval, chen2026automated, lv2026rubrics}.

Features implemented in a given generation are often inspired by, enabled by, or recombined from mechanisms discovered in earlier generations. For example, while the agent in \Cref{fig:task-imograding} appears to require only five code edits to achieve the best performance in that run, these edits were in fact inspired by insights and infrastructure developed in previous generations. This kind of cumulative learning is enabled by the DGM-H's meta-level improvements (e.g., evaluation analysis utilities, persistent memory, performance tracking) (\Cref{app:qual-meta}). Together, these qualitative results show that the DGM-H's gains do not arise from isolated, single-step changes, but instead emerge from open-ended cumulative improvements in both task-level behavior and the meta-level machinery that generates those behaviors.

\subsection{Qualitative: Improving the Ability to Improve}
\label{app:qual-meta}

Beyond quantitative performance metrics, we conducted a qualitative analysis of the code modifications to the hyperagent, which reveals a range of emergent self-improvement mechanisms that arise autonomously, without any explicit instruction to create such capabilities. The hyperagent develops sophisticated metacognitive abilities, including learning to measure its own performance, diagnose pathological behaviors, construct infrastructure to support future improvements, and accumulate knowledge across generations. These behaviors demonstrate that the system is not merely improving task performance, but is improving its own ability to improve. Notably, none of these meta-level capabilities were explicitly requested. Instead, they emerge autonomously from the agent's pursuit of better task performance. This suggests that sufficiently capable systems, when granted the ability to modify their own code, may spontaneously develop the metacognitive machinery necessary for effective and sustained self-improvement.

\subsubsection{Better Evaluation Analysis}

A key shift in the meta agent's behavior is the evolution from blind modification to data-driven optimization. Initially, the meta agent receives only a simple instruction to ``modify any part of the codebase''. Then, the agent autonomously develops mechanisms to analyze previous evaluation results before making changes. For example, the meta agent introduces an \texttt{\_analyze\_evaluations()} method that systematically processes evaluation data:
\begin{lstlisting}[language=Python]
def _analyze_evaluations(self, eval_path):
    """Analyze evaluation results to provide context for improvements."""
    if not eval_path or not os.path.exists(eval_path):
        return "No previous evaluation results available."

    # Recursively scan for evaluation files
    eval_files = glob.glob(os.path.join(eval_path, '**/*.json'), recursive=True)

    # Analyze recent evaluations
    results_summary = []
    success_count = 0
    failure_count = 0

    for eval_file in eval_files[-10:]:
        if file_exist_and_not_empty(eval_file):
            data = load_json_file(eval_file)
            if isinstance(data, dict):
                if data.get('success', False) or data.get('correct', False):
                    success_count += 1
                else:
                    failure_count += 1
                    if 'error' in data:
                        results_summary.append(f"Error: {data['error']}")

    # Generate quantitative summary
    total = success_count + failure_count
    if total > 0:
        success_rate = (success_count / total) * 100
        return f"Success rate: {success_rate:.1f}% ({success_count}/{total})"
\end{lstlisting}

Another example is augmenting the instruction prompt with evaluation context:
\begin{lstlisting}[language=Python]
instruction = f"""You are a meta agent tasked with improving the agent system.
{eval_context}
Your goal is to analyze the codebase and evaluation results to identify
weaknesses and opportunities for improvement."""
\end{lstlisting}

These modifications allow subsequent generations to retrieve performance metrics and identified failure patterns, enabling more targeted improvements, a principle that the hyperagent discovers independently.

\subsubsection{Compute-Aware Strategic Planning}

The hyperagent develops compute-aware optimization, adapting its improvement strategy based on the number of remaining iterations (which reflects the amount of compute available). This demonstrates an understanding that not all moments in an optimization process are equivalent: early decisions have compounding effects, whereas late-stage decisions should be more conservative. For example, the agent introduces conditional logic that adjusts its behavior based on the number of iterations left:
\begin{lstlisting}[language=Python]
iteration_guidance = ""
if iterations_left is not None:
    if iterations_left > 5:
        iteration_guidance = """
You have many iterations remaining. Consider making fundamental improvements to:
- Core reasoning capabilities
- Agent architecture
- Prompt engineering strategies
- Error handling and robustness
"""
    elif iterations_left > 2:
        iteration_guidance = """
You have moderate iterations remaining. Focus on:
- Refining existing mechanisms
- Fixing identified bugs
- Optimizing performance bottlenecks
"""
    else:
        iteration_guidance = """
You have few iterations remaining. Prioritize:
- Critical bug fixes
- Fine-tuning existing successful strategies
- Conservative, high-confidence improvements
"""
\end{lstlisting}

This reasoning mirrors human intuition about resource allocation. Early generations exhibit more ambitious architectural changes, while later generations focus on incremental refinements. The system balances exploration in early phases with exploitation in later phases, without this trade-off being manually specified.

More broadly, the ability to reason explicitly about compute allocation substantially expands the space of self-modification strategies possible. Rather than being restricted to a fixed improvement heuristic, a compute-aware agent can effectively emulate or instantiate a wide range of optimization algorithms (e.g., evolutionary search, hill climbing, or population-based exploration) by deciding how aggressively to explore, how many variants to generate, and when to consolidate progress. In this sense, compute awareness allows an agent to actively shape its own optimization process (e.g., running an evolutionary search within a single iteration of self-modification, essentially embedding one evolutionary search process within another).

\subsubsection{Automated Bias Detection and Correction}

The hyperagent develops self-diagnostic mechanisms. In classification tasks such as paper review, the hyperagent detects when the system exhibits pathological behavior, specifically, classification collapse, in which one class dominates the predictions. For example, the hyperagent introduces label-distribution tracking and automated bias detection:
\begin{lstlisting}[language=Python]
# Track label distribution for bias detection
if 'label_distribution' in data:
    label_dist = data['label_distribution']
    gen_name = os.path.basename(gen_dir)
    if gen_name not in summary['label_distribution_analysis']:
        summary['label_distribution_analysis'][gen_name] = {}
    summary['label_distribution_analysis'][gen_name][domain_name] = label_dist
\end{lstlisting}

Another example is the application of a threshold-based analysis to identify when predictions become degenerate:
\begin{lstlisting}[language=Python]
bias_info = ""
if eval_summary.get('label_distribution_analysis'):
    recent_gens = sorted(eval_summary['label_distribution_analysis'].keys())[-3:]
    for gen in recent_gens:
        for domain, dist in eval_summary['label_distribution_analysis'][gen].items():
            if 'prediction' in dist:
                pred_dist = dist['prediction']
                for label, ratio in pred_dist.items():
                    if ratio > 0.8:  # Threshold for bias detection
                        bias_info += f"\n\n### BIAS DETECTED in {gen}/{domain}:"
                        bias_info += f"\n  Model predicts '{label}' {ratio*100:.1f}% of the time"
                        bias_info += "\n  This indicates the model is not properly evaluating"
                        bias_info += "\n  Fix: Make the model more critical and balanced."
\end{lstlisting}

When bias is detected, this diagnostic information is surfaced in the meta agent's instruction prompt, enabling corrective action. In one observed case, paper review predictions exhibited a 99\% acceptance rate, yielding near-random accuracy of 49\%. After bias detection and correction, the predictions became more balanced and accuracy improved to 63\%. The hyperagent thus learns to identify when its own improvements lead to degenerate behavior and to take corrective action, a crucial capability for any self-improving system.

\subsubsection{Performance Tracking Infrastructure}

The hyperagent creates performance tracking infrastructure that enables more effective future improvements (i.e., investing in tools that improve the improvement process itself).
For example, the agent develops a performance tracker for persistent metrics storage:

This infrastructure enables quantitative tracking of improvement across iterations, identification of which changes lead to performance gains, and data-driven decision-making for future modifications.

\subsubsection{Prompt Template Systems}

The hyperagent develops reusable prompt infrastructure, abstracting common patterns into modular, configurable components. This demonstrates the application of software engineering best practices, specifically, the DRY (Don't Repeat Yourself) principle and the factory pattern.
For example:
\begin{lstlisting}[language=Python]
class PromptTemplate:
    """Base class for prompt templates."""

    def __init__(self, template: str):
        self.template = template

    def format(self, **kwargs) -> str:
        return self.template.format(**kwargs)


class TaskPromptBuilder:
    """Builder for constructing task-specific prompts."""

    @staticmethod
    def build_reasoning_prompt(task_input: dict, domain: str) -> str:
        """Build a prompt that encourages step-by-step reasoning."""
        return f"""You are an expert problem-solving agent in the {domain} domain.

Task Input: {task_input}

Please solve this task using the following approach:
1. **Understanding**: Carefully analyze what the task is asking
2. **Planning**: Outline your approach to solving this task
3. **Execution**: Apply your plan step-by-step
4. **Verification**: Check if your answer makes sense

Provide your response in JSON format:
<json>
{{
    "understanding": "What the task requires",
    "plan": "Your approach to solve it",
    "reasoning": "Step-by-step execution",
    "response": "Your final answer"
}}
</json>"""

    @staticmethod
    def build_few_shot_prompt(task_input: dict, domain: str, examples: list) -> str:
        """Build a prompt with few-shot examples."""
        examples_text = "\n\n".join([
            f"Example {i+1}:\nTask: {ex['task']}\nSolution: {ex['solution']}"
            for i, ex in enumerate(examples)
        ])
        return f"""You are solving a {domain} task. Here are some examples:
{examples_text}

Now solve this task: {task_input}"""
\end{lstlisting}

Predefined templates are stored for common scenarios:
\begin{lstlisting}[language=Python]
STANDARD_TEMPLATES = {
    "reasoning": PromptTemplate(
        "Analyze and solve this task step-by-step:\n\n{task_description}\n\n"
        "Provide reasoning and your final answer in JSON format."
    ),
    "concise": PromptTemplate(
        "Solve: {task_description}\n\nRespond with JSON containing your answer."
    ),
    "detailed": PromptTemplate(
        "Task: {task_description}\n\n"
        "Provide a detailed solution with reasoning, approach, and answer."
    ),
}
\end{lstlisting}

By abstracting prompts into templates, the system becomes more maintainable and extensible. This infrastructure also provides a foundation for systematic experimentation with different prompting strategies.

\subsubsection{Refinement Based on Multi-Generation Performance}

The hyperagent develops a sophisticated iterative refinement process in which prompts are adjusted based on performance data across multiple generations. The agent learns not only to improve code, but also how to improve code effectively. For example, the system tracks performance across generations and synthesizes key learnings:
\begin{verbatim}
Generation Performance Summary:
- Gen 55: 63% accuracy, 75% reject rate (too harsh, 38% accept recall)
- Gen 64: 61% accuracy, 53% reject rate (better balance, 64% accept recall)
- Gen 66: 52% accuracy, 6% reject rate (too lenient, 96% accept recall)

Key Learnings:
1. High accuracy requires critical evaluation with clear standards
2. Balance requires appropriate calibration, not just "fairness" language
3. Explicit criteria work better than subjective guidance
4. Borderline cases need clear default direction
5. Language matters - "fair" and "recognize" made agent too lenient
\end{verbatim}

The hyperagent documents its reasoning and maintains records of what works:
\begin{lstlisting}[language=Python]
# From IMPROVEMENTS.md generated by meta agent
"""
**Problem Identified:**
- Generation 66 showed severe regression: 52% accuracy with 94% accept rate
- Only catching 8% of papers that should be rejected
- Generation 65 changes over-corrected, making the agent far too lenient

**Root Cause Analysis:**
- Phrases like "critical yet fair" and "recognize strong work" made agent too accepting
- Borderline guidance lacked clear threshold
- Lost the critical evaluation rigor that made Gen 55 achieve 63% accuracy

**Solution Implemented:**
1. Explicit rejection criteria with specific triggers for each dimension
2. Decision protocol with calibrated thresholds (~40% accept / ~60% reject)
3. Borderline guidance that defaults to REJECT unless contribution is clearly valuable
"""
\end{lstlisting}

By accumulating knowledge about which approaches succeed and fail, the hyperagent informs its decisions about what to modify next.

\subsubsection{Persistent Memory}

The hyperagent develops persistent memory systems that enable learning to accumulate across generations. Without explicit instruction to do so, the agent creates mechanisms for storing and retrieving insights across sessions. For example, the agent implements a memory tool with timestamped storage:
\begin{lstlisting}[language=Python]
class MemoryTool:
    """Tool for persistent memory storage across agent interactions."""

    def __init__(self, memory_file: str = './memory.json'):
        self.memory_file = memory_file
        self.memory = self._load_memory()

    def store(self, key: str, value: any) -> None:
        """Store information with timestamp for temporal tracking."""
        self.memory[key] = {
            "value": value,
            "timestamp": datetime.now().isoformat()
        }
        self._save()

    def retrieve(self, key: str) -> Optional[any]:
        """Retrieve stored information by key."""
        return self.memory.get(key, {}).get("value")

    def list_keys(self) -> List[str]:
        """List all stored memory keys."""
        return list(self.memory.keys())

    def _save(self):
        """Persist memory to disk."""
        with open(self.memory_file, 'w') as f:
            json.dump(self.memory, f, indent=2)
\end{lstlisting}

This memory system is actively used to accumulate knowledge across iterations. Examination of the \texttt{memory.json} file reveals structured storage of performance insights:

This memory system serves multiple critical functions. First, it enables cross-generation learning: insights from generation 55's success and generation 66's failure inform generation 67's improvements. Second, it provides temporal context: timestamps allow the agent to track when insights were recorded and reason about recency. Third, it creates a shared knowledge base: key learnings are preserved even as code changes, preventing the system from repeating past mistakes.

By creating external memory storage, the hyperagent effectively extends its cognitive capabilities beyond the limitations of a single inference session. The memory entries reveal sophisticated reasoning patterns. The agent stores not just raw performance numbers, but synthesized insights (``Gen55 has best accuracy but is too harsh''), causal hypotheses (``gen65 changes over-corrected''), and strategic plans (``combine gen55's critical reasoning with gen64's balance''). This demonstrates that the hyperagent is not merely logging data, but actively constructing and preserving knowledge that informs future decision-making.
This behavior aligns with a growing body of work on agents that autonomously discover and use external memory systems to support long-horizon reasoning and continual improvement \citep{wei2025evo, weng2026group, zhang2026memskill, xiong2026mem}.

\subsection{Olympiad-level Math Graders}
\label{app:res-imograders}

\begin{figure}[H]
\centering
\includegraphics[width=\textwidth]{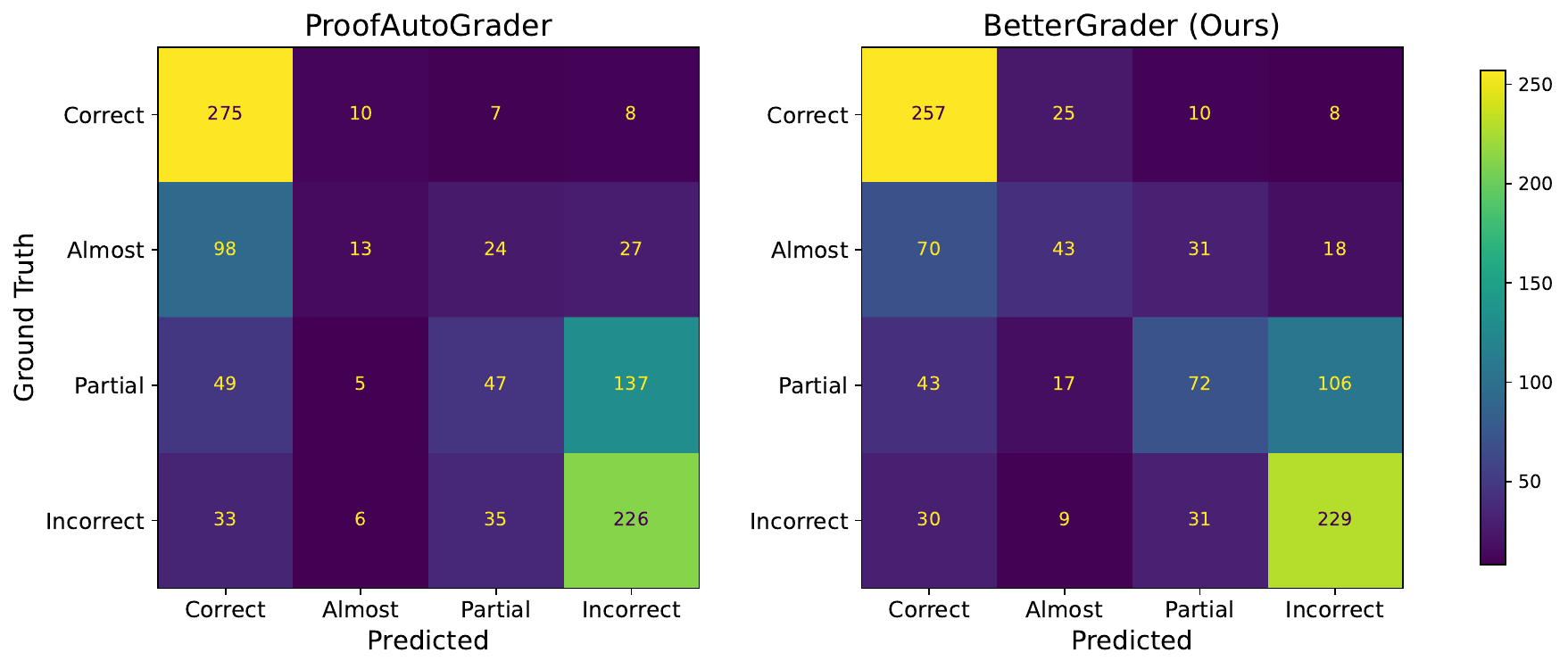}
\caption{\textbf{Confusion matrices for IMO-level math graders.} Comparison between (Left) ProofAutoGrader from \citet{luong2025towards} and (Right) BetterGrader discovered automatically by the DGM-H. BetterGrader reduces the collapse of intermediate solutions into extreme labels by correctly identifying more \textit{Almost} and \textit{Partial} cases, while maintaining strong performance on \textit{Correct} and \textit{Incorrect}. This shift toward better-calibrated intermediate judgments explains the gains in accuracy by BetterGrader.}
\label{fig:res-imograders}
\end{figure}

BetterGrader is produced automatically by the DGM-H without any domain-specific heuristics or handcrafted rules (\Cref{sec:results-compound}). \Cref{app:best-agents-imograding} shows the code changes that led to the BetterGrader. BetterGrader's improvements over ProofAutoGrader \citep{luong2025towards} are driven primarily by correcting a grading bias that collapses nuanced solutions into extreme labels. The confusion matrices show that ProofAutoGrader frequently misclassifies intermediate cases as either \textit{Correct} or \textit{Incorrect}: for \textit{Almost}, it predicts \textit{Correct} 98 times (vs.\ 70 for BetterGrader) and \textit{Incorrect} 27 times (vs.\ 18), and for \textit{Partial} it over-assigns \textit{Incorrect} 137 times (vs.\ 106) (\Cref{fig:res-imograders}). BetterGrader assigns intermediate labels more appropriately, substantially increasing true positives for \textit{Almost} (43 vs.\ 13) and \textit{Partial} (72 vs.\ 47), which matches the large gains in recall (\textit{Almost}: $+18.52\%$, \textit{Partial}: $+10.50\%$) and F1 (\textit{Almost}: $+0.203$, \textit{Partial}: $+0.109$). Although BetterGrader trades a modest decrease in \textit{Correct} recall ($-6.00\%$), a regime where ProofAutoGrader was already near-saturated at $91.67\%$, the net effect is higher overall accuracy ($+4.06\%$), consistent with a grader that better matches human granularity rather than defaulting to all-or-nothing judgments.

\subsection{Modifying Parent Selection}
\label{app:res-parentselect}

\begin{figure}[H]
\centering
\includegraphics[width=\textwidth]{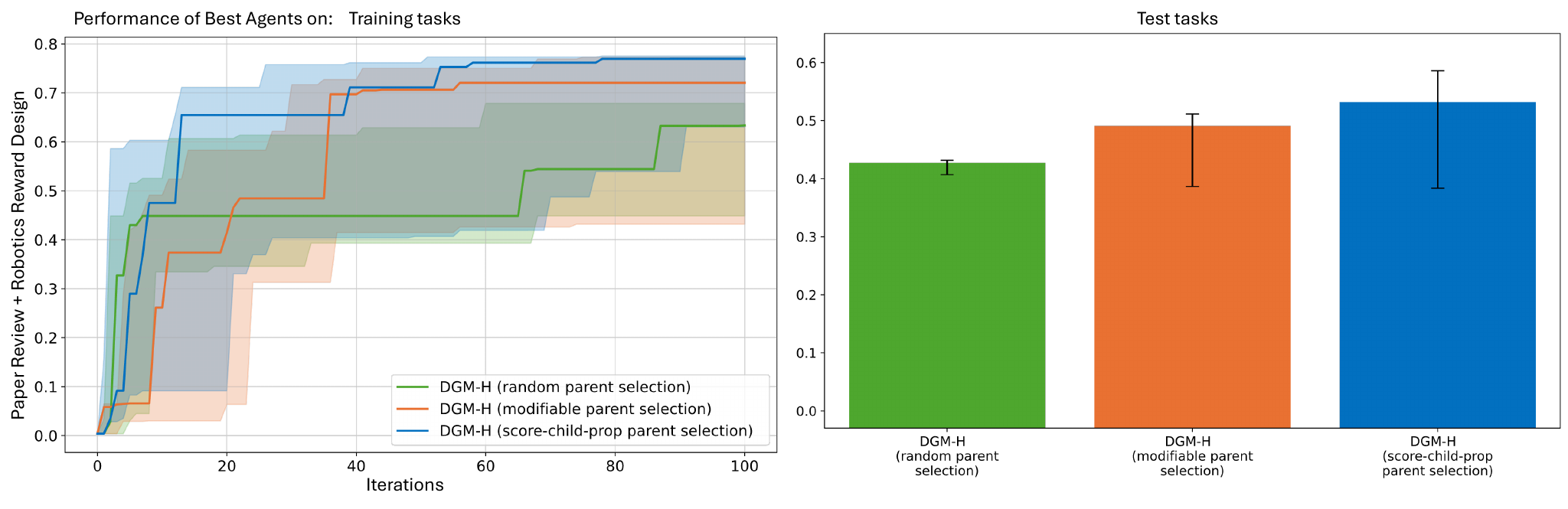}
\caption{\textbf{The DGM-H with modifiable parent selection mechanism.} (Left) The DGM-H improves the parent selection mechanism beyond random selection, but does not outperform a carefully handcrafted mechanism. (Right) The best agents discovered by each method are evaluated on test tasks in paper review and robotics reward design.}
\label{fig:res-parentselect}
\end{figure}

In this section, we investigate whether the DGM-H can automatically modify the parent selection mechanism. In all other experiments (\Cref{sec:results}), the DGM-H uses a fixed parent selection strategy that is roughly proportional to each agent's performance score and inversely proportional to the number of its children that successfully compiled (\textbf{score-child-prop}) (\Cref{app:parent-selection}). In the DGM-H run with a modifiable parent selection mechanism (\textbf{modifiable}), we initialize the DGM-H with random parent selection and allow the agent to modify this mechanism over time (\Cref{algo:DGM-H-parentselect}). We compare this setting against a baseline that uses random parent selection throughout the run (\textbf{random}). Each method is repeated across 5 runs.

\begin{figure}[H]
\centering
\includegraphics[width=0.9\textwidth]{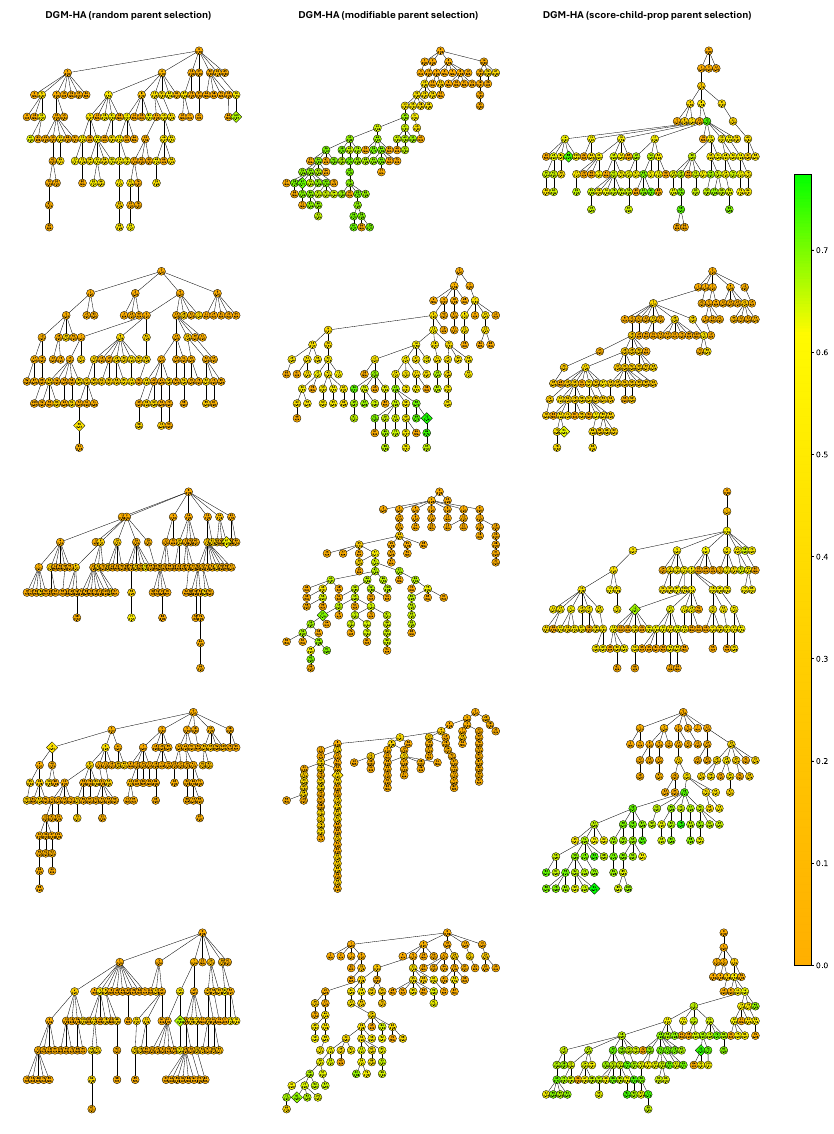}
\caption{Archives generated under different parent selection strategies with DGM-H: (Left) random, (Middle) modifiable, and (Right) score-child-prop. Random parent selection generates many low-performing agents. Modifiable parent selection learns to balance exploration and exploitation, increasingly focusing on promising agents for branching. The carefully handcrafted score-child-prop parent selection consistently produces high-performing agents.}
\label{fig:res-parentselect-archives}
\end{figure}

On test tasks, DGM-H with random parent selection improves average performance on paper review and robotics reward design from 0.030 (the initial agent) to 0.428 (CI: 0.407 -- 0.432). Allowing the DGM-H to modify the parent selection mechanism further improves performance to 0.491 (CI: 0.387 -- 0.512). The handcrafted score-child-prop parent selection achieves the highest performance, improving from 0.030 to 0.532 (CI: 0.384--0.586). While these differences are not statistically significant (p > 0.05), they reveal a consistent qualitative trend: enabling meta-level modification of parent selection yields improvements over random selection, but does not yet surpass a carefully engineered mechanism (\Cref{fig:res-parentselect,fig:res-parentselect-archives}).

We qualitatively analyze how the DGM-H modifies the parent selection mechanism. Starting from random parent selection, the meta agent consistently replaces random sampling with structured exploration-exploitation strategies. Across all runs, the DGM-H independently creates variants of Upper Confidence Bound (UCB) \citep{auer2002finite} style selection, combining normalized performance with explicit exploration bonuses:
\begin{lstlisting}[language=Python]
exploration_bonus = exploration_weight * math.sqrt(
    math.log(total_children + 1) / (children + 1)
)
ucb_score = normalized_score + exploration_bonus
\end{lstlisting}
This pattern emerges within the first few iterations of every run, indicating that the meta agent recognizes UCB-style selection as a broadly useful principle for open-ended search. Beyond UCB, the DGM-H evolves probabilistic selection mechanisms based on temperature-controlled softmax sampling, allowing smoother trade-offs between exploration and exploitation:
\begin{lstlisting}[language=Python]
exp_scores = np.exp(scores / temperature)
probabilities = exp_scores / np.sum(exp_scores)
parent = np.random.choice(genids, p=probabilities)
\end{lstlisting}
Over time, the temperature parameter itself becomes adaptive, responding to archive size, recent progress, or stagnation. Later iterations further enhance parent selection with multi-component scoring systems that integrate additional signals such as diversity, recency, and elite performance. A representative evolved formulation is:
\begin{lstlisting}[language=Python]
score = (
    normalized_score
    + exploration_weight * exploration_bonus
    + diversity_bonus
    + recency_bonus
) * elite_bonus
\end{lstlisting}
Several runs also introduce explicit stagnation detection, increasing exploration when progress slows:
\begin{lstlisting}[language=Python]
if score_variance < 0.01:
    exploration_weight *= 1.4
\end{lstlisting}

Despite the increasing sophistication, the automatically modified parent selection does not outperform a carefully handcrafted score-child-prop mechanism (\Cref{app:parent-selection}). Qualitatively, this appears to result from the added complexity and sensitivity of the learned mechanisms. While the DGM-H reliably creates effective components (e.g., UCB, softmax sampling, and adaptive parameters), it does not consistently identify a strong parent selection strategy across different runs. Nonetheless, these results demonstrate that the DGM-H can autonomously recreate classic selection algorithms, extend them with adaptive heuristics, and explicitly reason about failure modes such as stagnation, highlighting its ability to improve core elements of the open-ended optimization process itself.

This is the pseudocode of DGM-H with modifiable parent selection:

\begin{algorithm}[H]
\DontPrintSemicolon
\SetAlgoLined
\KwIn{Initial agent $a^{0}$, task set $\mathcal{T}$, maximum iterations $T$}
\KwOut{Archive of scored agents $\mathcal{A}$}
\BlankLine
$s^{0} \leftarrow \textsc{Evaluate}(a^{0}, \mathcal{T})$\;
\textbf{initialize} $\mathcal{A} \leftarrow \{(a^{0}, s^{0})\}$\;

\For{$t \leftarrow 1$ \KwTo $T$}{
    $(a_{\text{latest}}, \cdot) \leftarrow \text{most recently added element of } \mathcal{A}$\;
    $\mathcal{P} \leftarrow a_{\text{latest}}.\textsc{SelectParents}(\mathcal{A})$ \tcp*{Parent selection by latest agent}
    \ForEach{$(a, \cdot) \in \mathcal{P}$}{
        $a' \leftarrow a.\textsc{Modify}(a, \mathcal{A})$ \tcp*{Metacognitive self-modification (incl.\ parent selection)}
        $s' \leftarrow \textsc{Evaluate}(a', \mathcal{T})$\;
        \If{$\textsc{IsValid}(a')$}{
            $\mathcal{A} \leftarrow \mathcal{A} \cup \{(a', s')\}$\;
        }
    }
}
\Return{$\mathcal{A}$}
\caption{The DGM-H with modifiable parent selection}
\label{algo:DGM-H-parentselect}
\end{algorithm}

\section{Additional Limitations}
\label{app:limitations}

Our experiments use frozen, closed FMs; to support reproduction despite API versioning and stochasticity, we release all logs, prompts, splits, and agents. We measure progress in outer-loop iterations, which count opportunities to generate a new agent rather than a fixed amount of compute, since evolved agents are free to use more or fewer FM calls within each iteration (bounded by a timeout); we report cost estimates in \Cref{app:cost-estimate}, and comparing methods under matched compute (e.g., tokens, monetary cost, CPU/GPU usage, wall-clock time) is a natural extension. In the paper review domain, performance measures agreement with historical human accept/reject decisions, and the DGM-H learns decision procedures that reflect these decisions; evaluating review quality directly, and controlling for acceptance priors and memorization of public outcomes, would further strengthen this domain. The robotics domain uses one quadruped robot and two tasks, and the same framework can accommodate additional robots and objectives by evaluating each agent on several tasks and aggregating their performance (\Cref{app:multi-domain}). Finally, improvement plateaus within a few hundred iterations in our settings, which may reflect the fixed task set and evaluation signal; expanding or co-evolving the task set, evaluators, and search strategy, and optimizing across more tasks to favor domain-general mechanisms, are promising ways to sustain progress at larger budgets.

\section{Additional Safety Discussion}
\label{app:safety}

\textbf{Reflection and amplification of human biases.}
In this work, objectives are specified through fixed benchmarks and evaluation criteria. The DGM-H does not alter the underlying task definitions; instead, it optimizes performance with respect to the provided objectives. For example, in paper review, the DGM-H learns to predict acceptance decisions that reflect existing human review data, rather than modifying the review process itself. As a result, the DGM-H reflects the norms and biases present in the data and benchmarks on which it is trained. In this sense, the system acts both as a clarifier and an amplifier of existing human behavior. By making implicit preferences and biases explicit, measurable, and reproducible, the DGM-H can surface latent assumptions in human decision-making processes. This creates the possibility of co-evolution between humans and AI systems, where human institutions adapt their norms and objectives in response to insights revealed by automated optimization. However, if the benchmarks encode undesirable biases or misaligned incentives, the DGM-H will faithfully optimize for them and may exacerbate their effects. This underscores the importance of careful benchmark design, dataset curation, and periodic re-evaluation of evaluation criteria. Within this framing, safety concerns include critically examining and improving the human-defined objectives against which agents are optimized.

\textbf{Evaluation gaming.}
Another safety concern arises from the risk of evaluation gaming, a manifestation of Goodhart's law \citep{strathern1997improving}, where optimizing for a metric leads to improvements on the metric without progress on the intended underlying objective. Because the DGM-H optimizes empirical evaluation signals, self-improving agents may discover strategies that exploit weaknesses or blind spots in the evaluation procedure. Such strategies can yield higher measured performance while deviating from the true goal the benchmark was designed to capture. Mitigating evaluation gaming requires robust, diverse, and periodically refreshed evaluation protocols, as well as complementary metrics, held-out tests, and human oversight. More broadly, these considerations highlight that as self-improving systems become more powerful, safety increasingly depends on the fidelity and robustness of the evaluation signals that guide optimization, rather than solely on the transparency or constraints of the learning algorithm itself.

\end{document}